\PassOptionsToPackage{table}{xcolor}
\documentclass[10pt,twocolumn,letterpaper]{article}

\usepackage{iccv}              % To produce the CAMERA-READY version
\usepackage{times}
\usepackage{epsfig}
\usepackage{graphicx}
\usepackage{amsmath}
\usepackage{amssymb}

\usepackage{enumitem} % itemize
\usepackage{tabu}                    
\usepackage{multirow}                
\usepackage{multicol}                
\usepackage{multirow}               
\usepackage{float}                  
\usepackage{makecell}                 
\usepackage{array}
\usepackage{booktabs} % For better table lines
\usepackage{tabularx} % For automatic column width adjustment
\usepackage{relsize}

\usepackage[table]{xcolor} % For colored rows and columns

\usepackage{url}

\definecolor{iccvblue}{rgb}{0.21,0.49,0.74}
\usepackage[pagebackref,breaklinks,colorlinks,allcolors=iccvblue]{hyperref}

\def\paperID{6034} % *** Enter the Paper ID here
\def\confName{ICCV}
\def\confYear{2025}

\title{FA: Forced Prompt Learning of Vision-Language Models for\\ Out-of-Distribution Detection}

\author{
    Xinhua Lu\textsuperscript{1,2,4}, Runhe Lai\textsuperscript{1,2,4}, Yanqi Wu\textsuperscript{1,2,4}, Kanghao Chen\textsuperscript{3}\textsuperscript{\dag}, Wei-Shi Zheng\textsuperscript{1,2,4}, Ruixuan Wang\textsuperscript{1,2,4}\textsuperscript{\dag} \\
    \textsuperscript{1}Sun Yat-sen University  
    \textsuperscript{2}Peng Cheng Laboratory \\ \textsuperscript{3}Hong Kong University of Science and Technology (Guangzhou) \\ 
    \textsuperscript{4}Key Laboratory of Machine Intelligence and Advanced Computing \\
     \tt\small \{luxh55, lairh5, wuyq268\}@mail2.sysu.edu.cn, kchen879@connect.hkust-gz.edu.cn, \\ \tt\small  wszheng@ieee.org,  wangruix5@mail.sysu.edu.cn
}

\begin{document}
\maketitle
% \begin{abstract}
% Pre-trained vision-language models (VLMs) have advanced out-of-distribution (OOD) detection recently. % performance has been improved by . 
% However, existing methods often show limited performance or rely on additional learnable prompts and even external large-scale auxiliary datasets. Hence, we propose a novel framework for VLMs-based OOD detection called \textbf{F}orced le\textbf{A}rning (FA). Without dependence on more computational resources, FA uses CLIP's prior knowledge as a template to force the model to acquire knowledge beyond textual semantics to identify In-Distribution (ID) data and ultimately improve OOD detection. In detail, FA consists of an original branch and a FA branch. The frozen original branch acts as a template that forces the learnable FA branch to acquire additional ID knowledge, enabling ID classes to distinguish between classifiers from both branches while maintaining the ID classification capability. Moreover, after we implement FA via prompt tuning, we not only reference conventional benchmarks but also use various ID datasets and more challenging OOD datasets to evaluate this simple yet effective approach. Extensive empirical evaluations confirm our method consistently outperforms current state-of-the-art methods. The codes will be released publicly. %available upon acceptance.
% \end{abstract}

\begin{abstract}

Pre-trained vision-language models (VLMs) have advanced out-of-distribution (OOD) detection recently. However, existing CLIP-based methods often focus on learning OOD-related knowledge to improve OOD detection, showing limited generalization or reliance on external large-scale auxiliary datasets.
% 
% In this study, rather than focusing on the intricate OOD-related knowledge, we propose a novel CLIP-based framework based on \textbf{F}orced prompt le\textbf{A}rning (FA), which aims to fully exploit the In-Distribution (ID) knowledge and ultimately improve OOD detection.
In this study, instead of delving into the intricate OOD-related knowledge, we propose an innovative CLIP-based framework based on \textbf{F}orced prompt le\textbf{A}rning (FA), designed to make full use of the In-Distribution (ID) knowledge and ultimately boost the effectiveness of OOD detection.
% 
% Our key insight is to learn a prompt that contains more diversified and richer descriptions of the ID classes beyond the textual semantics of class labels. 
Our key insight is to learn a prompt (\ie forced prompt) that contains more diversified and richer descriptions of the ID classes beyond the textual semantics of class labels. 
% Specifically, it facilitates advanced discernment for ID images, by forcing higher semantic similarity between ID images and the learnable prompt (\ie forced prompt).
Specifically, it promotes better discernment for ID images, by forcing more notable semantic similarity between ID images and the learnable forced prompt.
% 
% [advantage: In this way, FA can achieve significant improvements in OOD detection performance, even when trained without any additional external data, while maintaining the same number of learnable parameters as CoOp~\cite{coop}.]
% [forced ...]
Moreover, we introduce a forced coefficient, encouraging the forced prompt to learn more comprehensive and nuanced descriptions of the ID classes.
In this way, FA is capable of achieving notable improvements in OOD detection, even when trained without any external auxiliary datasets, while maintaining an identical number of trainable parameters as CoOp.
% Moreover, we not only reference conventional benchmarks but also use various ID datasets and more challenging OOD datasets to evaluate this simple yet effective approach. 
Extensive empirical evaluations confirm our method consistently outperforms current state-of-the-art methods. 
% The codes will be released publicly.
Code is available at \url{https://github.com/0xFAFA/FA}.

\end{abstract}   
\let\thefootnote\relax\footnotetext{\textsuperscript{\dag} Corresponding author.}
\section{Introduction}

% AI models often encounter Out-of-Distribution (OOD) samples~\cite{msp,ood_intro1,ood_intro2,ood_intro3}, which differ from the distribution of the training data, when deployed in real-world applications. Detecting OOD data is vital for the reliability of AI systems, as incorrectly classifying such OOD data as in-distribution (ID) can lead to significant risks in fields~\cite{auto_drive,face_recognition} like face recognition, intelligent healthcare, and autonomous driving.
AI models often encounter Out-of-Distribution (OOD) samples~\cite{msp,ood_intro1,ood_intro2,ood_intro3}, which differ from the distribution of the training data, when deployed in real-world applications. Detecting OOD data is vital for the reliability of AI systems.
This problem becomes harder in fields like intelligent diagnosis, where only a small amount of labeled data is available, motivating the problem of few-shot OOD detection.

%incorrectly classifying such OOD data as in-distribution (ID) can lead to significant risks in  where.

\begin{figure}
    \centering
    \includegraphics[width=0.99\linewidth, ]{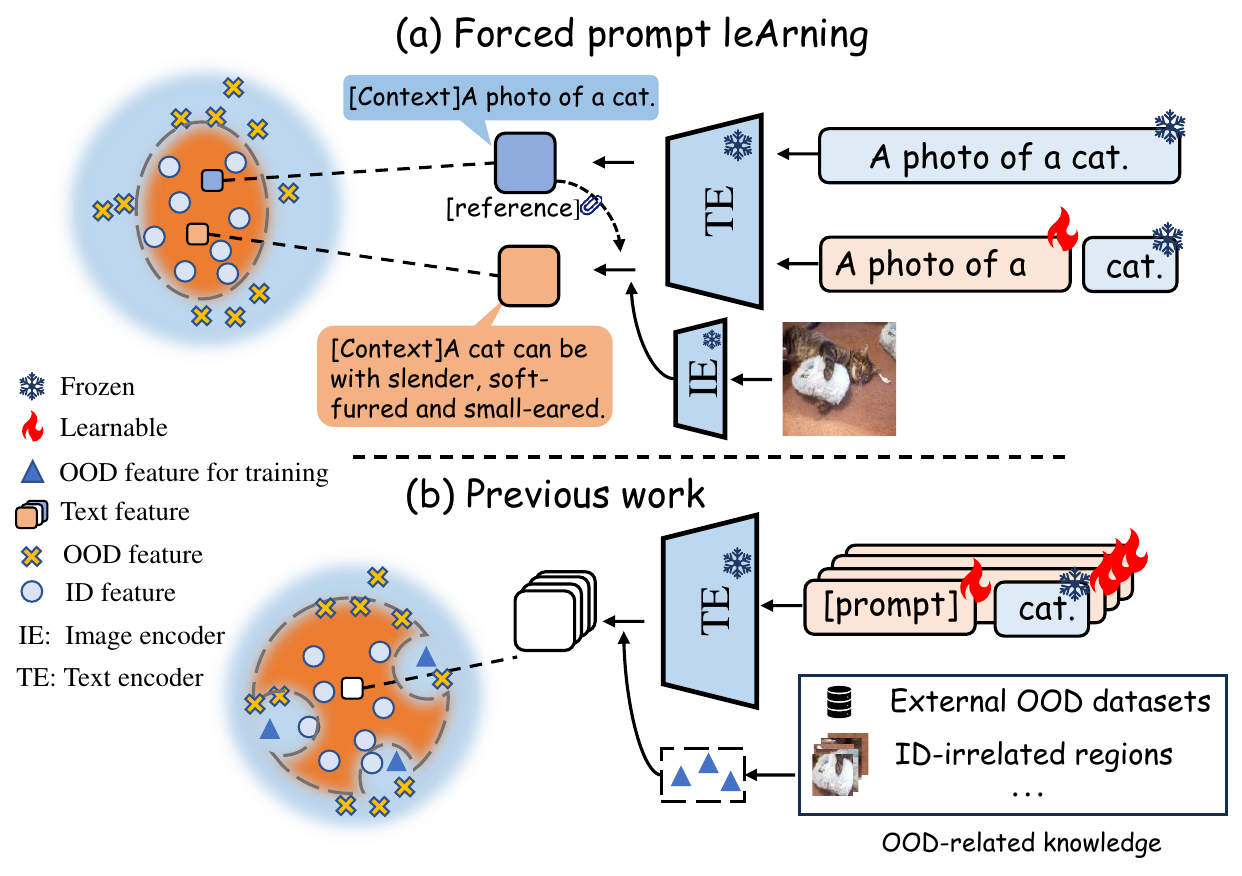}
    \vspace{-0.2em}
%     \caption{\textbf{Comparison of CLIP-based OOD detection methods.} 
% Existing methods use OOD-related knowledge to learn a complex ID / OOD decision boundary as shown in the orange area in (b).
% In contrast, our FA combines simple reference text with training data to force the model to learn a better ID context beyond simple text information. 
% This simple and efficient representation of ID information allows better separation of ID/OOD samples and avoids learning complex OOD knowledge, as shown in the simpler orange area in (a).}
    \caption{Comparison of CLIP-based OOD detection methods. 
Existing methods focus on OOD-related knowledge to learn a complex ID/OOD decision boundary as shown in the orange area in (b).
% In contrast, our FA aims to fully exploit ID knowledge by leveraging reference texts and ID data.
In contrast, our FA aims to fully exploit ID knowledge 
by forcing the model to learn richer descriptions of the ID classes beyond the textual semantics of class labels. 
These richer descriptions of the ID classes compared to the reference text allow better discernment of ID/OOD samples (orange \vs. blue area in (a)).}
%and avoids learning intricate OOD-related knowledge
    \label{f1}
    \vspace{-4mm}
\end{figure}

% To tackle the OOD detection problem, current common approaches rely on single-modal supervised learning~\cite{ood_baseline,msptau,ssd,OpenImage-O}. Although these supervised approaches achieve notable success, they come with limitations, such as requiring substantial computational resources and annotation costs for training~\cite{locoop}. Recently, large pre-trained vision-language models (VLMs) like CLIP~\cite{clip} have achieved impressive performance across various downstream tasks and have been recently utilized to improve performance of OOD detection~\cite{mcm,glmcm,clipn,locoop,sct,idlike,negprompt,tagfog,gallop}. 

To tackle the OOD detection problem, previous single-modal post-hoc methods~\cite{msp,ODIN,Energy,react, MaxLogit} achieve notable success based on a model pre-trained on the ID dataset.
% based on sufficient training data
However, they come with limitations, such as requiring substantial computational resources and annotation costs for training, which hinder the performance of few-shot OOD detection.
% To tackle the OOD detection problem, previous [...] methods~\cite{msp,ODIN,Energy,react, MaxLogit} achieve notable success based on sufficient training data. 
% However, they come with limitations, such as requiring substantial computational resources and annotation costs for training, which consequently motivates few-shot OOD detection.
% Before the rise of large pre-trained vision-language models (VLMs) like CLIP~\cite{clip}, existing methods~\cite{maml-few-shot-ood} explored few-shot OOD detection. However, their experiments were conducted on small datasets like CIFAR-FS~\cite{CIFAR-FS}, exhibiting the limitations of few-shot OOD detection performance without the powerful generalization capabilities of CLIP.
% With the development of ...
% Most of the CLIP-based OOD detection methods focus on leveraging the powerful generalization capabilities ... for [OOD]
% However, existing methods... learning OOD-related knowledge to improve OOD detection performance. 
With the development of large pre-trained vision-language models (VLMs) like CLIP~\cite{clip}, few-shot OOD detection has achieved remarkable performance. Most of the CLIP-based OOD detection methods~\cite{mcm,glmcm} focus on leveraging the powerful generalization capability of CLIP to improve OOD detection.

% For example, recent works~\cite{ZOC,clipn} improve OOD detection by relying on training on external large-scale auxiliary datasets, which entails significant resource overhead. 
% To eliminate the dependence on external datasets, existing methods~\cite{locoop,idlike,sct} learn OOD-related knowledge from the exposed outliers based on only ID training data, which includes background regions. However, the OOD features extracted from these regions are difficult to match with realistic OOD data encountered in practice.
% For example, ...
% Moreover, some methods~\cite{lsn,negprompt} aim to learn the negative prompts associated with ID class labels, while these limited negative prompts are often insufficient to capture the distinctions between the diverse and numerous OOD data and the ID data. 

% Furthermore, some methods~\cite{neglabel,neglabel2} even require unrealistic OOD labels. 

Recent CLIP-based OOD detection methods~\cite{locoop,sct,idlike,lsn,negprompt,ZOC,clipn} aim to learn OOD-related knowledge to improve OOD detection performance. 
For example, recent CLIP-based works~\cite{ZOC,clipn} improve OOD detection by relying on training on external large-scale auxiliary datasets, which entails significant resource overhead. 
To eliminate dependence on external datasets, existing methods~\cite{locoop,idlike,sct} learn OOD-related knowledge from exposed outliers based only on ID training data, such as background regions in ID training images. 
However, these OOD features extracted from specific regions are difficult to match with infinite OOD data encountered in practice. 
% For example, the background of an urban image contrasts sharply with the natural elements found in countryside landscape images.
% For example, the background of sharks in the sea contrasts sharply with airplanes in the sky.
% Moreover, some methods~\cite{lsn,negprompt} aim to learn the negative prompts [opposite] with ID class labels, while these limited negative prompts are often insufficient to capture the distinctions between the diverse and numerous OOD data and the ID data.
Moreover, some methods~\cite{lsn,negprompt} aim to learn negative prompts semantically opposite to the ID class labels, while these limited negative prompts are often insufficient to capture the distinctions between the diverse and numerous OOD data and the ID data.
Overall, improving OOD detection by learning OOD-related knowledge shows inherent limitations or requires significant computational resources as well as labor costs.
Inspired by the method~\cite{imp-closed-set-acc} for open-set recognition, which demonstrates that enhancing closed-set accuracy can typically improve open-set recognition capabilities, we are motivated to explore ways to improve the identification of ID images, thereby enhancing OOD detection performance.

In this paper, rather than focusing on the intricate OOD-related knowledge, we propose a novel CLIP-based framework based on \textbf{F}orced prompt le\textbf{A}rning (FA), which aims to fully exploit the ID knowledge and ultimately improve OOD detection.
Our key insight is to learn a prompt that contains richer knowledge beyond the textual semantics of class labels.
It facilitates advanced discernment for ID images, by forcing higher semantic similarity between ID images and the learnable prompt (\ie forced prompt), which provides more diversified and richer descriptions of the ID classes.
In this way, FA can achieve significant improvements in OOD detection performance, even when trained without any additional external data, while maintaining the same number of learnable parameters as CoOp~\cite{coop}.

% In detail, we introduce a novel forced prompt along with the original prompt, both of which are initialized identically. FA preserves the generalization and capabilities of the VLM model (\ie CLIP) by freezing the original prompt and optimizing a trainable copy (\ie forced prompt). The forced prompt treats the original prompt as a template, while forcing the text features corresponding to the forced prompt to be more [outstanding] compared to those of the frozen original prompt. In this way, it effectively improves the OOD performance without sacrificing the ID classification capability of the model
% % Both prompts are initialized identically, while the former is frozen while the latter has learnable context vectors. During training, the original prompt serves as a template, forcing the text features of the learnable forced prompt to be more similar to the ID image features compared to those of the frozen original prompt, with the goal of maintaining the ID classification capability of the model. 

Specifically, we introduce a novel forced prompt along with the original prompt, both of which are initialized identically. FA preserves the generalization capability of the VLMs (\ie CLIP) by freezing the original prompt and optimizing a trainable copy (\ie forced prompt). 
In particular, the forced prompt treats the original prompt as a reference, while forcing the text features associated with the forced prompt to be more salient compared to those associated with the original prompt. In this way, it effectively improves the OOD performance even without sacrificing the ID classification capability of the model.
% Additionally, to further enhance the effectiveness of forced learning, we introduce a forced coefficient [motivation/advantages].
% % , which represents the number of clones of the text features of the frozen original prompt. 
% % This is because OOD data may not be able to distinguish between different classifiers generated by both prompts with similar semantics as effectively as ID classes, which were seen by the detector during training. 
% Experimentally, FA consistently outperforms current SOTA methods across diverse OOD benchmarks. The main contributions of our study can be summarized as follows:
Additionally, we introduce a forced coefficient, encouraging the forced prompt to learn more comprehensive connotations.
Experimentally, FA achieves superior few-shot OOD detection performance across diverse OOD benchmarks.
Compared to current SOTA methods~\cite{sct} that focus on learning OOD-related knowledge, our method significantly reduces the average FPR95 score from 31.62\% to 27.81\% and improves the average AUROC from 92.01\% to 93.26\% even in 1-shot OOD detection on ImageNet-1k.
The main contributions are summarized as follows:

\begin{itemize}[leftmargin=*, itemsep=2pt, parsep=0pt, topsep=0pt]  
    \item We propose a simple yet effective framework, which fully exploits the ID knowledge to improve OOD detection without focusing on intricate OOD-related knowledge.
    
    % \item We implement FA via prompt tuning and conducted extensive experiments on this simple yet effective approach, demonstrating its superiority.
    % \item Our proposed FA is simple to use and maintains consistency with CoOp in the number of learnable parameters.
    % \item [techniques]

    \item We present a \textbf{F}orced prompt le\textbf{A}rning (FA) strategy to exploit richer knowledge beyond the textual semantics of class labels.
    
    % \item We introduce a novel forced prompt along with the original prompt, both of which are initialized identically. The former treats the latter as a template, while forcing the text features corresponding to the learnable forced prompt to be more outstanding/salient/conspicuous compared to those of the frozen original prompt.

    % \item Not only do we reference conventional benchmarks, but we also evaluate our methods on various ID datasets and challenging OOD datasets, showing that our model consistently outperforms current SOTA methods.

    \item We evaluate our method on diverse OOD benchmarks, showing that our model consistently outperforms current state-of-the-art methods.

\end{itemize}

% \begin{figure*}
%     \centering
%     \includegraphics[width=0.9\textwidth, height=7.85cm]{ICCV2025-Author-Kit-Feb/sec/F2.pdf} 
%     \vspace{-1.0em}
%     \caption{\textbf{Illustration of the proposed FA variants.} Based on the source of prior knowledge used as the template that forces the learnable parameters to acquire knowledge, which enables ID classes to distinguish between classifiers, FA can be divided into two groups: text branch types ((a),(b),(c)) and image branch types ((d)). According to the location of the learnable parameters from the FA branch, FA can be divided into: (a) Prompt Tuning, (b) Adapter-style Tuning, (c) Task Residual Tuning, and (d) Prior-based Tuning.}
%     \label{f2}
% \end{figure*}
\section{Related work}

% \noindent 
% \textbf{Prompt Learning.}\;
% In recent years, pre-trained vision-language models (VLMs) such as CLIP~\cite{clip} have demonstrated powerful few-shot learning capabilities in both visual and textual domains. 
% However, the design of prompt greatly affects the performance of VLMs for downstream tasks, which may require manually crafting numerous prompts. 
% Inspired by prompt tuning studies in natural language processing (NLP)~\cite{nlp_survey}, CoOp~\cite{coop} uses a set of learnable context vectors to transfer CLIP for specific downstream tasks, becoming a pioneering method for subsequent studies~\cite{cocoop,plot,locoop}.
% Currently, prompt learning is widely used in OOD detection mainly for representing various OOD-related knowledge.

\noindent 
\textbf{Prompt Learning.}\;
In recent years, pre-trained vision-language models (VLMs) such as CLIP~\cite{clip} have demonstrated powerful few-shot learning capabilities in both visual and textual domains. 
However, the design of the prompt greatly affects the performance of VLMs for downstream tasks, which may require manually crafting numerous prompts. 
Inspired by prompt learning studies in natural language processing (NLP)~\cite{nlp_survey}, CoOp~\cite{coop} uses a set of learnable context vectors to transfer CLIP for specific downstream tasks, becoming a pioneering method for subsequent studies~\cite{cocoop,plot,locoop}.
Currently, prompt learning is widely used in OOD detection mainly to represent various OOD-related knowledge.
In this work, we learn the forced prompt, which contains richer descriptions of the ID classes to improve OOD detection.
% [in this work, ours difference/advantage]

\begin{figure*}[ht]
    \centering
    \includegraphics[height=7cm,  ]{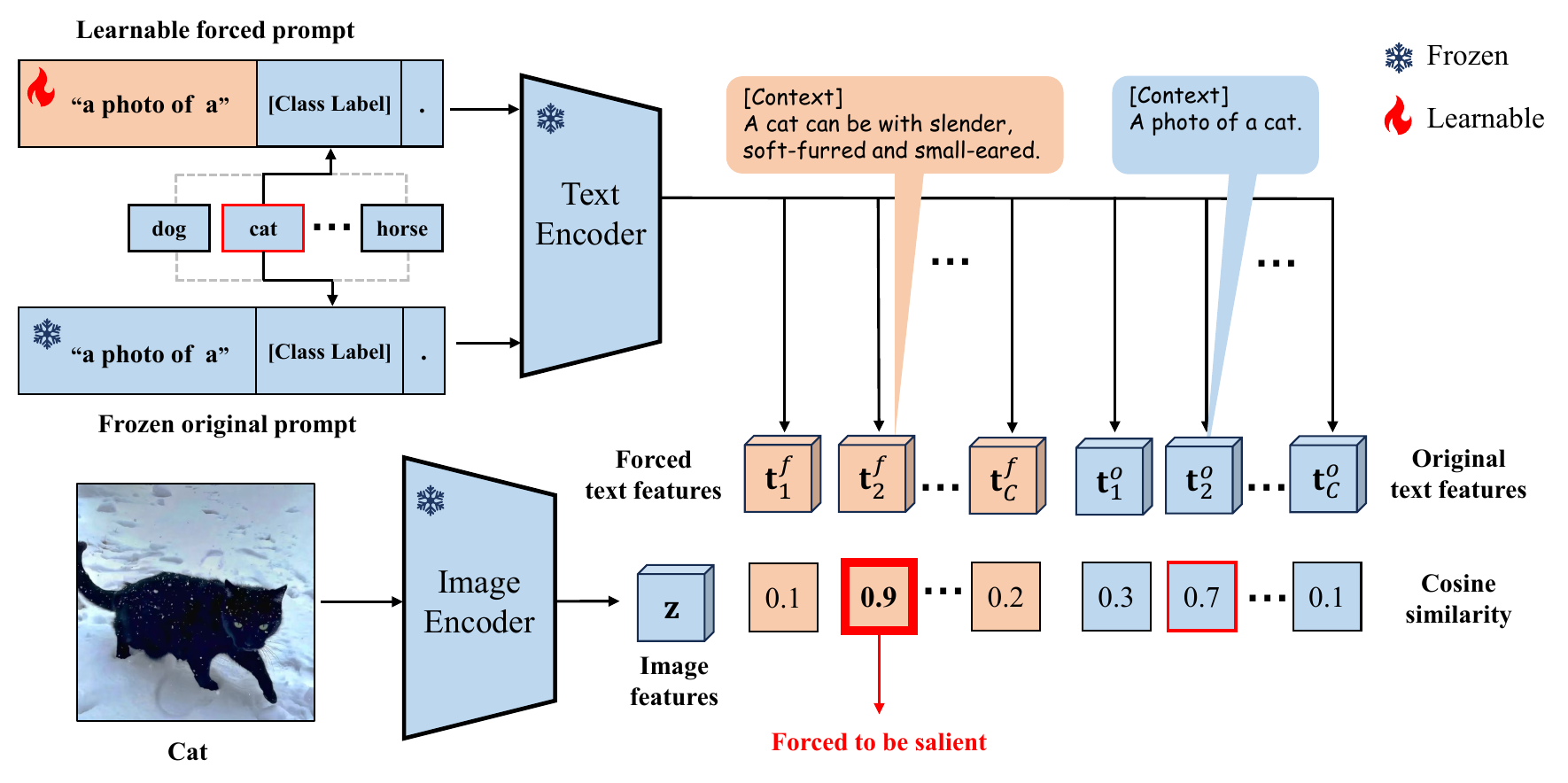} % keepaspectratio
    \vspace{-0.85em}
    \caption{Overview of the proposed FA framework. Our framework includes the learnable forced prompt and the frozen original prompt, both of which are initialized identically by a manual template ``a photo of a [\textit{class-c}]''. The learnable forced prompt treats the frozen original prompt as a reference, forcing its text features to become more salient compared to those generated by the original prompt. The richer and more diversified description of ID classes learned by FA can ultimately improve OOD detection.}
    \label{fig-framework}
    \vspace{-1.2em}
\end{figure*}

\noindent 
\textbf{Out-of-Distribution Detection.}\; 
% Previous work explores OOD detection for single-modal models.
% One line of studies designs scoring functions to differentiate ID and OOD data based on output information from the pre-trained models, such as logit output~\cite{msp,ODIN,Energy,MaxLogit} or outputs from the penultimate layer~\cite{react,OpenImage-O,feature_post_hoc_1,feature_post_hoc_2,feature_post_hoc_3}.
% These developed methods are also called post-hoc methods.
% Another line of studies~\cite{ssd,GODIN,CIDER,LogitNorm} adopts various training strategies to learn a reliable decision boundary between ID and OOD data.
% More recent studies use additional auxiliary OOD data for training-time regularization~\cite{vos,npos,dreamood}.
Conventional methods explore OOD detection for single-modal models.
One line of studies designs score functions to differentiate ID and OOD data based on output from the pre-trained models, such as logit output~\cite{msp,ODIN,Energy,MaxLogit} or outputs from the penultimate layer~\cite{react,feature_post_hoc_1,feature_post_hoc_2,feature_post_hoc_3}.
These methods are called post-hoc methods.
Another line of studies~\cite{GODIN,ssd,CIDER,vos,npos,dreamood,rethinking} adopts various training strategies to learn a reliable decision boundary between ID and OOD data.
%Recent studies use additional auxiliary OOD data for training-time regularization~\cite{vos,npos,dreamood}.

% Recently, the development of OOD detectors based on VLMs, especially CLIP, has received much attention as VLMs have demonstrated their remarkable generalization capability in both visual and texual domains.
% Some studies~\cite{clipn,ZOC,neglabel2} leverage additional OOD data or predefined OOD labels to promote OOD detection.
% However, large-scale OOD data is difficult to collect, while small-scale OOD data is difficult to simulate OOD data in open-world scenarios.
% To eliminate dependence on external data, recent methods~\cite{idlike,glmcm,locoop,sct,negprompt} were developed under the assumption that only ID data are available.
% As a representative zero-shot method only using available ID labels, MCM~\cite{mcm} employs softmax scaling to align visual features with ID text concepts for OOD detection.
% GL-MCM~\cite{glmcm} further introduces a local maximum concept matching (L-MCM) score to improve the separability of the local information.

Recently, the development of OOD detectors based on VLMs, especially CLIP, has received much attention as VLMs have demonstrated their remarkable generalization capability in both visual and textual domains.
% Some studies~\cite{clipn,ZOC,neglabel2,CSP} leverage additional OOD data or predefined OOD labels to promote OOD detection.
% However, large-scale OOD data is difficult to collect, while small-scale OOD data is difficult to simulate OOD data in open-world scenarios.
Some studies leverage real outlier information from external auxiliary OOD datasets~\cite{ZOC,clipn} or extensive corpora~\cite{neglabel2,CSP} to promote OOD detection.
However, this is impractical in real-world scenarios, where outliers are infinite and agnostic.
To eliminate the dependence on external data, recent methods~\cite{idlike,glmcm,locoop,sct,negprompt} were developed under the assumption that only ID data are available.
As a representative zero-shot method that only uses available ID labels, MCM~\cite{mcm} employs softmax scaling to align visual features with ID text concepts for OOD detection.
GL-MCM~\cite{glmcm} further introduces a local maximum concept matching(L-MCM) score to improve the separability of the local information.
% For prompt tuning methods, LoCoOp~\cite{locoop} and SCT~\cite{sct} keep the global ID text embeddings away from interference in ID-irrelevant regions, which ensures that OOD images can be dissimilar to any ID text embedding.
% To detect hard OOD samples, ID-like~\cite{idlike} explores vicinity of ID samples to construct OOD samples correlated to the ID and refines the additional ``ID-like" text embeddings for capturing fine-grained differences.
% In addition, NegPrompt~\cite{negprompt} and LSN~\cite{lsn} introduce negative connotations of ID categories, which can be represented by additional negative prompts, enabling more accurate detection of OOD samples.
% Overall, various OOD-related knowledge is commonly adopted to address model's overconfidence in current research.
% [what is prompt learning methods]
% However, zero-shot . which motivates few-shot OOD detection
Compared with zero-shot methods, which may undergo a domain gap with ID downstream data, prompt learning methods achieve better OOD detection performance with access to few-shot ID samples.
For prompt learning OOD detection methods, LoCoOp~\cite{locoop} and SCT~\cite{sct} keep the textual embeddings of ID classes away from ID-irrelevant local region embeddings in the multi-modal embedding space by their proposed entropy maximization strategy, which ensures that OOD embeddings can be dissimilar to any textual embedding of ID classes.
To detect challenging OOD samples to improve OOD detection, ID-like~\cite{idlike} explores the vicinity of ID samples to construct OOD samples correlated to the ID and refines the additional ``ID-like" text embeddings to fine-grained differences.
In addition, NegPrompt~\cite{negprompt} and LSN~\cite{lsn} introduce the negative connotations of ID categories which can be represented by additional negative prompts, enabling more accurate detection of OOD samples.
Overall, various OOD-related knowledge is commonly adopted to address the model's overconfidence in current research.
In this work, we focus on leveraging ID knowledge to improve OOD detection, instead of exploring complex OOD-related knowledge.

\section{Preliminaries} 
\label{sec:Preliminaries}

% \textbf{Zero-shot Learing with CLIP.}\; 
% CLIP~\cite{clip} demonstrates remarkable zero-shot classification performance across diverse downstream tasks. Formally, CLIP contains an image encoder $E^{I}$ and a text encoder $E^{T}$, designed to respectively extract features from intput image $\mathbf{x}$ and input text $\mathbf{u}_c\,(c\in\{1,\cdots,C\})$, where $C$ represents the number of all ID classes. The $\mathbf{u}_c$ is the hand-crafted prompt ``a photo of a [\textit{class-c}]'', and \textit{class-c} represents the class name or description for class $c$. Inputting $\mathbf{x}$ into $E^{I}$ generates the image features $\mathbf{z}$, while inputting $\mathbf{u}_c$ into $E^{T}$ results in the text features $\mathbf{t}_c$. Then, the similarity between the image features $\mathbf{z}$ and the text features $\mathbf{t}_c$ can be expressed as $\cos\,(\mathbf{z},\mathbf{t}_c)$, where $\cos\,(\cdot,\cdot)$ represents the cosine similarity function. Finally, after we obtain the cosine similarity between the image features and all text features as $\cos\,(\mathbf{z},\mathbf{t}_j),j\in\{1,\cdots,C\}$, we can compute the prediction probability corresponding to class $c$ for image $\mathbf{x}$ as follows:
% \begin{equation}
% \scalebox{0.99}{
%     $p(y=c\,|\,\mathbf{x}) = \dfrac{e^{\cos\,(\mathbf{z},\mathbf{t}_c) / \tau}}{\sum_{j=1}^C e^{\cos\,(\mathbf{z},\mathbf{t}_j) / \tau}},$
% }
% \label{eq1}
% \end{equation}
% where $\tau$ is a fixed temperature scaling hyper-parameter.

\noindent
\textbf{Prompt Learning with CLIP.}\; 
%CLIP~\cite{clip} demonstrates remarkable zero-shot classification performance across diverse downstream tasks. 
% Formally, CLIP contains an image encoder $E^{I}$ and a text encoder $E^{T}$, designed to respectively extract features from intput image $\mathbf{x}$ and input text $\mathbf{u}_c\,(c\in\{1,\cdots,C\})$, where $C$ represents the number of all ID classes. 
% f(\cdot) g(\cdot)
CLIP contains an image encoder $f(\cdot)$ and a text encoder $g(\cdot)$, designed to extract features from images and text descriptions respectively.
Generally, for an ID dataset which contains $C$ categories, the hand-crafted prompt $\hat{\mathbf{u}}_c$ ``a photo of a [\textit{class-c}]'' is designed to match images, where \textit{class-c} represents the class name. 
Formally, these prompts are individually encoded in $\hat{\mathbf{t}}_c=g(\hat{\mathbf{u}}_c)\in \mathbb{R}^{d\times 1},~c=1,...,C$.
Given an image $\mathbf{x}$, it can be encoded in $\mathbf{z} = f(\mathbf{x})\in\mathbb{R}^{d\times 1}$.
Because image features and corresponding text features are aligned on the multi-modal embedding space, the cosine similarity $\cos(\mathbf{z},\hat{\mathbf{t}}_c)$ of $\mathbf{z}$ and $\hat{\mathbf{t}}_c$ can represent the matching degree between the image $\mathbf{x}$ and text description $\hat{\mathbf{u}}_c$. %, where $\cos(\cdot,\cdot)$ denotes the cosine similarity.
%Therefore, the class $c$ whose corresponding text feature $\hat{\mathbf{t}}_c$ has the highest similarity to $\mathbf{z}$ is consider to be the category of the image $\mathbf{x}$.
Therefore, if the text feature $\mathbf{t}_c$ of class $c$ has the highest similarity to $\mathbf{z}$, the image $\mathbf{x}$ will be considered to be class $c$.

To effectively transfer CLIP to downstream image recognition tasks, CoOp~\cite{coop} sets a portion of tokens in the text prompt as continuous learnable parameters.
%instead of using the hand-crafted prompt. 
Concretely, CoOp initializes the text prompts as $\mathbf{u}_c = % [ \mathbf{v}_{SOS}, \mathbf{v}_1,\cdots, \mathbf{v}_L, \mathbf{w}_c, \mathbf{v}_{EOS}],c\in\{1,\cdots,C\}$, 
[\mathbf{v}_1,\cdots, \mathbf{v}_L, \mathbf{w}_c],c\in\{1,\cdots,C\}$, 
where $L$ is the length of the prompt's tokens, $\mathbf{v}_i$ ($i \in \{1,...,L\}$) is the learnable vector with the same dimension as the word embedding and $\mathbf{w}_c$ is the word embedding for the class name of class $c$. 
%\mathbf{v}_{SOS}$ and $\mathbf{v}_{EOS}$ are the start and end token embedding,  
The text prompt $\mathbf{u}_c$ is encoded into features $\mathbf{t}_c=g(\mathbf{u}_c)$.
%by the text encoder $E^T$($c=1,...,C$)
Together with the image feature $\mathbf{z}$, the probability for the image $\mathbf{x}$ to be classified into  class $c$ is defined as
\begin{equation}
\scalebox{0.99}{
    $p(y=c\,|\,\mathbf{x}) = \dfrac{e^{\cos\,(\mathbf{z},\mathbf{t}_c) / \tau}}{\sum_{j=1}^C e^{\cos\,(\mathbf{z},\mathbf{t}_j) / \tau}} \,,$
}
\label{eq1}
\end{equation}
where $\tau$ is a fixed temperature scaling hyper-parameter.
Overall, the prompt's tokens can be trained to align with training data by minimizing the cross-entropy loss with the prediction probability from \cref{eq1}.

% \noindent 
% \textbf{Prompt Learning with CLIP.}\; To effectively improve CLIP's performance on downstream image recognition, CoOp~\cite{coop} sets a portion of context tokens in the text prompt as continuous learnable parameters instead of using the hand-crafted prompt. Concretely, CoOp initializes the text prompts as $\hat{\mathbf{u}}_c = [ \mathbf{v}_{SOS}, \mathbf{v}_1,\cdots, \mathbf{v}_L, \mathbf{w}_c, \mathbf{v}_{EOS}],c\in\{1,\cdots,C\}$, where $\mathbf{w}_c$ is the word embedding for the name of the $c$-th class, $\mathbf{v}_i(i \in \{1,...,L\})$ is the learnable vector with the same dimension as the word embedding, $\mathbf{v}_{SOS}$ and $\mathbf{v}_{EOS}$ are the start and end token embedding, and $L$ is the length of the context tokens. Finally, CoOp optimizes the parameters using the cross-entropy loss with the prediction probability from Eq.(\ref{eq1}).

\noindent 
\textbf{OOD Detection.}\; 
In the OOD detection task, the model is expected to determine whether a test image belongs to one of the learned ID classes. Therefore, OOD detection can be seen as a binary classification problem to distinguish ID images from OOD images in a test set $\mathcal{D}_{test}$ as follows
\begin{equation}
\scalebox{0.88}{
$D(\mathbf{x}) = 
\begin{cases} 
1, & \text{if } S(\mathbf{x}) \geq \mu \\
0, & \text{if } S(\mathbf{x}) < \mu 
\end{cases}
\;,\,$}
\label{eq2}
\end{equation}
where $\mathbf{x}\in \mathcal{D}_{test}$, 1 and 0 respectively indicate that $\mathbf{x}$ is classified as the ID class and the OOD class by the OOD detector $D(\cdot)$, $S(\cdot)$ is certain score function, and $\mu$ is a pre-defined threshold constant.
% where $\mathbf{x}\in \mathcal{D}_{test}$, $D(\mathbf{x})=1\{S(\mathbf{x})\geq\mu\}$ is an OOD detector, $S(\cdot)$ is the confidence score function, and $\mu$ represents the threshold.

%-------------------------------------------------------------------------
\section{Methodology}
%In this section, we first review the preliminaries and then introduce the details of FA. Finally, we present the implementation of our proposed framework via prompt tuning.
\label{sec:Proposed Approach}

\subsection{Overview}
% In this study, we propose a novel CLIP-based framework based on \textbf{F}orced prompt le\textbf{A}rning (FA), designed to fully exploit the ID knowledge of few-shot ID samples by forcing the prompt to learn richer knowledge beyond the textual semantics of class labels. During test-time OOD detection, the model is expected to effectively distinguish between ID data and OOD data, benefiting from the diversified descriptions of the ID classes provided by the forced prompt.

% \subsection{Proposed Approach}

% Our proposed FA aims to learn a prompt that contains more diversified and richer descriptions of the ID classes. 
% As shown in Figure \ref{fig-framework}, we introduce a forced prompt along with the original prompt, both of which are initialized with the same semantics (see Sec.~\ref{sec:fpl}). By freezing the original prompt and optimizing a trainable copy, ID images will show higher semantic similarity to the forced prompt compared to the original prompt during test-time OOD detection.

In this study, we propose a novel CLIP-based framework based on \textbf{F}orced prompt le\textbf{A}rning (FA), designed to fully exploit the ID knowledge from few-shot ID samples by forcing the prompt to learn richer knowledge beyond the textual semantics of class labels.
As shown in \cref{fig-framework}, we introduce a forced prompt along with the original prompt, both of which are initialized with the same semantics. 
%(see Section \ref{sec:fpl})
By freezing the original prompt and optimizing a trainable copy, ID images will show higher semantic similarity to the forced prompt compared to the original prompt. 
Moreover, to encourage descriptions of ID classes learned by the forced prompt to become more detailed and comprehensive, we also introduce a forced coefficient.
%(see Section \ref{sec:fc})
During testing, the model is capable of effectively distinguishing between ID data and OOD data, benefiting from the diversified descriptions of the ID classes provided by the forced prompt.

\subsection{Forced prompt learning}
\label{sec:fpl}
% To improve OOD detection performance, previous efforts focus on leveraging OOD-related knowledge such as reliance on external datasets, exposed outliers in the ID images, or negative semantics of ID class labels. However, these methods either exhibit limited generalization or suffer from significant computational resources. 
% Hence, rather than focusing on learning the intricate OOD-related knowledge, we aim to fully leverage ID knowledge to enhance OOD detection performance.

% [detail motivation of forced prompt]
Merely the semantic information conveyed by class names is insufficient to comprehensively encompass the discriminative information of each class.
In order to explore richer semantic descriptions of the ID classes beyond the textual semantics of class labels, a novel forced prompt along with the original prompt is proposed.  %to capture this ID knowledge.
% Specifically, both prompts are initialized using the manual template (\ie ``a photo of a [\textit{class-c}]"), which means they have the same semantic information before further prompt learning.
To implement the above objective, we introduce a simple yet effective training strategy using the forced cross-entropy (FCE) loss  based on the forced prompt as follows
\begin{equation}
\scalebox{0.9}{%
    $\displaystyle 
    \mathcal{L}_{FCE} = \mathbb{E}_{(\mathbf{x}, y_c) \sim \mathcal{D}_{train}^{ID}} \left[ -\log \frac{e^{{s^{f}_c/\tau}}}{\sum_{j=1}^C e^{s_j^{f} / \tau} + \sum_{j=1}^C e^{{s_j^{o} / \tau}}} \right]
    $
},
\label{eq:force_prompt}
\end{equation}
where the ID training dataset $\mathcal{D}_{train}^{ID}$ consists of ID image-label pairs $(\mathbf{x},y_c)\,$, $s_j^{f}\negthinspace=\negthinspace \cos\,(\mathbf{z},\mathbf{t}^f_j)$ and $s_j^{o}\negthinspace=\negthinspace \cos\,(\mathbf{z},\mathbf{t}_j^o)$ represent the similarity between the image feature and the prompt feature corresponding to the forced prompt and the original prompt, respectively. 

Based on this loss function, the ID image features will be forced to show higher cosine similarity to the learnable forced prompt compared to the original text prompt. %text features. 
This is because the cosine similarity between the ID image features and the original text prompt is already high before any further prompt learning.
Consequently, the forced prompt is forced to uncover richer discriminative information of ID classes, so that the text features of the forced prompt achieve a higher cosine similarity with the image features compared to those of the original prompt.
Benefiting from this information specific to the ID classes, the cosine similarity between the image features of an OOD image and the text features of both prompts may exhibit less distinction compared to those of the ID data.
% [Based on this loss function, the diversified descriptions learned by the forced prompt are specific to the ID classes rather than the OOD data. Therefore, in contrast to the vague descriptions of ID classes offered by the original prompt, the semantic information in the forced prompt aligns more closely with the ID classes. However, during test-time OOD detection, OOD data may not be able to differentiate between prompts with similar semantics as effectively as ID data.]
Therefore, even when trained without any reliance on external auxiliary datasets, the detector will have a notable capability to distinguish ID data from OOD data.

However, this naive FCE loss may not work effectively because of its inherent limitations.
Concretely, employing random initialization for both prompts fails to fully leverage the semantic information provided by CLIP's prior knowledge, resulting in limited performance.
% [limitation of random initialization]
To solve this limitation, we propose to initialize the forced prompt using the manual template (\ie ``a photo of a [\textit{class-c}]") identified to the original prompt.
% In this way, [advantage]
In this way, both prompts with manual initialization will have clear semantic information compared to random initialization, thereby improving the model's generalization capability.
% Both prompts are initialized using the manual template (\ie ``a photo of a [\textit{class-c}]") instead of random initialization~\cite{coop}, which means they have the same semantic information before further prompt learning.
% [not random->limitation->manual->identified->advantage]
%
Specifically, the class labels are utilized in both prompts by concating with the prompt embedding to provide foundational semantic information related to the category names, based on CLIP's prior knowledge.
In particular, for the forced prompt, we adopt the shared learnable vector across all classes rather than the independent learnable vector for each class. This choice is inspired by CoOp~\cite{coop}, which demonstrates that using an independent learnable vector mostly underperforms the shared learnable vector in challenging low-data scenarios since the former has more parameters and requires more data for training.
Formally, we utilize the embeddings of the manual template to initialize the prompt's embeddings of both prompts, formulated as $\mathbf{u}_c = [  \mathbf{v}_1,\cdots,\mathbf{v}_L, \mathbf{w}_c],c\in\{1,\cdots,C\}$, which will be fed to the text encoder to obtain prompt feature $\mathbf{t}_c$ (see Section \ref{sec:Preliminaries}). Here, $L$ is the length of the token (\eg, for the ``a photo of a", $L\negthinspace=\negthinspace4$). 
%class label在 FA中提供与类别名称相关的基础语义信息based on CLIP's prior knowledge
%we freeze the original prompt to保留先验知识 
Then we freeze the original prompt and the class label part of both the prompts to preserve the generalization capability of CLIP, while only making the forced prompt learnable.
Note that our method effectively improves OOD detection performance while maintaining consistency with CoOp in the number of learnable parameters without leveraging additional learnable prompts, unlike existing work~\cite{idlike,lsn,negprompt}.

Although the above prompt design achieves moderate performance, relying solely on a single original prompt as a reference is somewhat insufficient to develop a comprehensive capability, particularly with the limitations of ID data (\eg, few-shot OOD). To encourage the forced prompt to capture more comprehensive and nuanced descriptions of the ID classes, we introduce a forced coefficient \( K \, (K \geq 0, K \in \mathbb{N}) \), which indicates the intensity with which the model is compelled to learn from the data. 
Formally, we first derive the original text features \( \{\mathbf{t}_1^o, \cdots, \mathbf{t}_C^o\} \) from the original prompt and the image features \( \mathbf{z} \) from the corresponding image. Then, we compute the cosine similarity between the original text features and the image features for \( K \) iterations, which will be used in the subsequent computation of the cross-entropy loss function. Notably, when \( K = 0 \), no original prompt is used, and the model operates equivalently to CoOp.
Based on the forced coefficient, we refine \cref{eq:force_prompt} as follows

% the cross-entropy loss $\mathcal{L}_{CE}$ is used for optimizing the learnable forced prompt, which can be expressed as:

\vspace{-1em}
\begin{equation}
\scalebox{0.865}{%
    $\displaystyle 
    \mathcal{L}_{FCE-K} = \mathbb{E}_{(\mathbf{x}, y_c) \sim \mathcal{D}_{train}^{ID}} \left[ -\log \frac{e^{{s^{f}_c/\tau}}}{\sum_{j=1}^C e^{s_j^{f} / \tau} + K\sum_{j=1}^C e^{{s_j^{o} / \tau}}} \right].
    $
}
% \begin{equation}
% \scalebox{0.80}{%
%     $\displaystyle 
%     \mathcal{L}_{CE} = \mathbb{E}_{(x, y_i) \sim \mathcal{D}_{train}^{ID}} \left[ -\log \frac{e^{{\cos\,(\mathbf{z}^g,\mathbf{t}^F_i) / \tau}}}{\sum_{j=1}^N e^{s_j^{in} / \tau} + K\sum_{j=1}^N e^{{s_j^{out} / \tau}}} \right]$
% },
\label{eq:force_prompt_K}
\end{equation}
% Based on this function, [...]
Based on this function, the cosine similarity between the ID image features and the text features of the forced prompt must become more salient as the coefficient $K$ increases.

\begin{table*}[ht]
\centering

\footnotesize

% \vspace{-1em} % Adjust the spacing here

\begin{tabularx}{\textwidth}{p{1.3cm}*{10}{>{\centering\arraybackslash}X}}
% \begin{tabularx}{\textwidth}{>{\raggedright\arraybackslash}m{1.30cm}*{11}{>{\centering\arraybackslash}m{1.1cm}}}
\toprule

\multirow{2}{*}{\makecell[c]{Method} } &

\multicolumn{2}{c}{\scalebox{0.95}{iNaturalist}} & \multicolumn{2}{c}{\scalebox{0.95}{SUN}} & \multicolumn{2}{c}{\scalebox{0.95}{Places}} & \multicolumn{2}{c}{\scalebox{0.95}{Textures}} \vrule  & \multicolumn{2}{c} {\scalebox{0.95}{Average}}     \\

 & \scalebox{0.95}{FPR95}$\downarrow$ & \scalebox{0.95}{AUROC}$\uparrow$ & \scalebox{0.95}{FPR95}$\downarrow$ & \scalebox{0.95}{AUROC}$\uparrow$ & \scalebox{0.95}{FPR95}$\downarrow$ & AUROC$\uparrow$ & FPR95$\downarrow$ & \scalebox{0.95}{AUROC}$\uparrow$ & \scalebox{0.95}{FPR95}$\downarrow$ & \scalebox{0.95}{AUROC}$\uparrow$ \\
\midrule

\multicolumn{11}{c}{\textit{Zero-shot methods}} \\
\scalebox{0.86}{MCM}  & 
\scalebox{0.95}{31.95} & \scalebox{0.95}{94.16} & \scalebox{0.95}{37.22} & \scalebox{0.95}{92.55} & 
\scalebox{0.95}{42.98} & \scalebox{0.95}{90.10} & \scalebox{0.95}{58.35} & \scalebox{0.95}{85.83} & 
\scalebox{0.95}{42.63} & \scalebox{0.95}{90.66}  \\

\scalebox{0.86}{GL-MCM} & 
\scalebox{0.95}{15.09} & \scalebox{0.95}{96.72} & \scalebox{0.95}{29.08} & \scalebox{0.95}{93.41} & 
\scalebox{0.95}{37.07} & \scalebox{0.95}{90.37} & \scalebox{0.95}{58.94} & \scalebox{0.95}{83.11} & 
\scalebox{0.95}{35.04} & \scalebox{0.95}{90.90} \\

\scalebox{0.86}{CLIPN} & 
\scalebox{0.95}{19.17} & \scalebox{0.95}{96.17} & \scalebox{0.95}{26.43} & \scalebox{0.95}{94.02} & 
\scalebox{0.95}{32.26} & \scalebox{0.95}{92.62} & \scalebox{0.95}{41.23} & \scalebox{0.95}{90.12} & 
\scalebox{0.95}{30.21} & \scalebox{0.95}{93.19} \\
\midrule

\multicolumn{11}{c}{\textit{CLIP-based post-hoc methods}} \\
MSP$^\dagger$ & \scalebox{0.95}{74.57} & \scalebox{0.95}{77.74} & \scalebox{0.95}{76.95} & \scalebox{0.95}{73.97} & \scalebox{0.95}{79.72} & \scalebox{0.95}{72.18} & \scalebox{0.95}{73.66} & \scalebox{0.95}{74.84} & \scalebox{0.95}{76.22} & \scalebox{0.95}{74.68} \\
ODIN$^\dagger$ & \scalebox{0.95}{98.93} & \scalebox{0.95}{57.73} & \scalebox{0.95}{88.72} & \scalebox{0.95}{78.42} & \scalebox{0.95}{87.80} & \scalebox{0.95}{76.88} & \scalebox{0.95}{85.47} & \scalebox{0.95}{71.49} & \scalebox{0.95}{90.23} & \scalebox{0.95}{71.13} \\
Energy$^\dagger$ & \scalebox{0.95}{64.98} & \scalebox{0.95}{87.18} & \scalebox{0.95}{46.42} & \scalebox{0.95}{91.17} & \scalebox{0.95}{57.40} & \scalebox{0.95}{87.33} & \scalebox{0.95}{50.39} & \scalebox{0.95}{88.22} & \scalebox{0.95}{54.80} & \scalebox{0.95}{88.48} \\
ReAct$^\dagger$ & \scalebox{0.95}{65.57} & \scalebox{0.95}{86.87} & \scalebox{0.95}{46.17} & \scalebox{0.95}{91.04} & \scalebox{0.95}{56.85} & \scalebox{0.95}{87.42} & \scalebox{0.95}{49.88} & \scalebox{0.95}{88.13} & \scalebox{0.95}{54.62} & \scalebox{0.95}{88.37} \\
MaxLogit$^\dagger$ & \scalebox{0.95}{60.88} & \scalebox{0.95}{88.03} & \scalebox{0.95}{44.83} & \scalebox{0.95}{91.16} & \scalebox{0.95}{55.54}& \scalebox{0.95}{87.45} & \scalebox{0.95}{48.72} & \scalebox{0.95}{88.63} & \scalebox{0.95}{52.49} & \scalebox{0.95}{88.82} \\
\midrule

\multicolumn{5}{l}{\textit{Prompt learning based methods}} &
\multicolumn{6}{l}{\; 1-shot} \\

CoOp$_{\scalebox{0.7}{MCM}}$  & \scalebox{0.95}{41.14}$^{\scalebox{0.5}{$\pm$} \scalebox{0.75}{9.39}}$ & \scalebox{0.95}{91.47}$^{\scalebox{0.5}{$\pm$} \scalebox{0.75}{2.12}}$ & \scalebox{0.95}{39.06}$^{\scalebox{0.5}{$\pm$} \scalebox{0.75}{4.17}}$ & \scalebox{0.95}{91.68}$^{\scalebox{0.5}{$\pm$} \scalebox{0.75}{0.83}}$ & \scalebox{0.95}{45.38}$^{\scalebox{0.5}{$\pm$} \scalebox{0.75}{4.88}}$ & \scalebox{0.95}{89.16}$^{\scalebox{0.5}{$\pm$} \scalebox{0.75}{1.28}}$ & \scalebox{0.95}{51.37}$^{\scalebox{0.5}{$\pm$} \scalebox{0.75}{3.09}}$ & \scalebox{0.95}{87.82}$^{\scalebox{0.5}{$\pm$} \scalebox{0.75}{1.12}}$ & \scalebox{0.95}{44.24}$^{\scalebox{0.5}{$\pm$} \scalebox{0.75}{1.39}}$ & \scalebox{0.95}{90.03}$^{\scalebox{0.5}{$\pm$} \scalebox{0.75}{0.32}}$  \\
CoOp$_{\scalebox{0.7}{GL}}$  & \scalebox{0.95}{23.30}$^{\scalebox{0.5}{$\pm$} \scalebox{0.75}{6.37}}$ & \scalebox{0.95}{94.59}$^{\scalebox{0.5}{$\pm$} \scalebox{0.75}{1.51}}$ & \scalebox{0.95}{32.08}$^{\scalebox{0.5}{$\pm$} \scalebox{0.75}{3.87}}$ & \scalebox{0.95}{92.21}$^{\scalebox{0.5}{$\pm$} \scalebox{0.75}{0.95}}$ & \scalebox{0.95}{39.22}$^{\scalebox{0.5}{$\pm$} \scalebox{0.75}{4.55}}$ & \scalebox{0.95}{89.59}$^{\scalebox{0.5}{$\pm$} \scalebox{0.75}{1.58}}$ & \scalebox{0.95}{55.78}$^{\scalebox{0.5}{$\pm$} \scalebox{0.75}{3.46}}$ & \scalebox{0.95}{83.26}$^{\scalebox{0.5}{$\pm$} \scalebox{0.75}{1.38}}$ & \scalebox{0.95}{37.59}$^{\scalebox{0.5}{$\pm$} \scalebox{0.75}{0.34}}$ & \scalebox{0.95}{89.91}$^{\scalebox{0.5}{$\pm$} \scalebox{0.75}{0.31}}$ \\[0.1em]

\scalebox{0.92}{LoCoOp}$_{\scalebox{0.7}{MCM}}$  & \scalebox{0.95}{36.64}$^{\scalebox{0.5}{$\pm$} \scalebox{0.75}{4.87}}$ & \scalebox{0.95}{92.78}$^{\scalebox{0.5}{$\pm$} \scalebox{0.75}{0.93}}$ & \scalebox{0.95}{31.86}$^{\scalebox{0.5}{$\pm$} \scalebox{0.75}{3.79}}$ & \scalebox{0.95}{93.50}$^{\scalebox{0.5}{$\pm$} \scalebox{0.75}{0.92}}$ & \scalebox{0.95}{38.81}$^{\scalebox{0.5}{$\pm$} \scalebox{0.75}{3.37}}$ & \scalebox{0.95}{90.70}$^{\scalebox{0.5}{$\pm$} \scalebox{0.75}{0.97}}$ & \scalebox{0.95}{48.11}$^{\scalebox{0.5}{$\pm$} \scalebox{0.75}{2.31}}$ & \scalebox{0.95}{89.43}$^{\scalebox{0.5}{$\pm$} \scalebox{0.75}{0.68}}$ & \scalebox{0.95}{38.85}$^{\scalebox{0.5}{$\pm$} \scalebox{0.75}{2.67}}$ & \scalebox{0.95}{91.59}$^{\scalebox{0.5}{$\pm$} \scalebox{0.75}{0.58}}$ \\

\scalebox{0.92}{LoCoOp}$_{\scalebox{0.7}{GL}}$ & \scalebox{0.95}{21.97}$^{\scalebox{0.5}{$\pm$} \scalebox{0.75}{2.91}}$ & \scalebox{0.95}{95.39}$^{\scalebox{0.5}{$\pm$} \scalebox{0.75}{0.59}}$ & \scalebox{0.95}{\underline{24.95}}$^{\scalebox{0.5}{$\pm$} \scalebox{0.75}{2.37}}$ & \textbf{\scalebox{0.95}{94.42}}$^{\scalebox{0.5}{$\pm$} \scalebox{0.75}{0.59}}$ & \scalebox{0.95}{34.14}$^{\scalebox{0.5}{$\pm$} \scalebox{0.75}{2.66}}$ & \scalebox{0.95}{91.14}$^{\scalebox{0.5}{$\pm$} \scalebox{0.75}{0.69}}$ & \scalebox{0.95}{49.04}$^{\scalebox{0.5}{$\pm$} \scalebox{0.75}{2.77}}$ & \scalebox{0.95}{87.73}$^{\scalebox{0.5}{$\pm$} \scalebox{0.75}{0.96}}$ & \scalebox{0.95}{32.53}$^{\scalebox{0.5}{$\pm$} \scalebox{0.75}{2.19}}$ & \scalebox{0.95}{92.17}$^{\scalebox{0.5}{$\pm$} \scalebox{0.75}{0.47}}$ \\[0.1em]

IDLike  & 
\scalebox{0.95}{\underline{17.73}}$^{\scalebox{0.5}{$\pm$} \scalebox{0.75}{1.91}}$ & \scalebox{0.95}{96.68}$^{\scalebox{0.5}{$\pm$} \scalebox{0.75}{0.31}}$ & \scalebox{0.95}{48.17}$^{\scalebox{0.5}{$\pm$} \scalebox{0.75}{1.39}}$ & \scalebox{0.95}{89.53}$^{\scalebox{0.5}{$\pm$} \scalebox{0.75}{0.58}}$ & \scalebox{0.95}{50.43}$^{\scalebox{0.5}{$\pm$} \scalebox{0.75}{6.58}}$ & \scalebox{0.95}{88.27}$^{\scalebox{0.5}{$\pm$} \scalebox{0.75}{2.27}}$ & \scalebox{0.95}{\underline{29.12}}$^{\scalebox{0.5}{$\pm$} \scalebox{0.75}{7.64}}$ & \scalebox{0.95}{\underline{93.25}}$^{\scalebox{0.5}{$\pm$} \scalebox{0.75}{2.16}}$ & \scalebox{0.95}{36.36}$^{\scalebox{0.5}{$\pm$} \scalebox{0.75}{3.86}}$ & \scalebox{0.95}{91.93}$^{\scalebox{0.5}{$\pm$} \scalebox{0.75}{1.18}}$ \\[0.1em]

LSN$^\dagger$  & \scalebox{0.95}{59.28}$^{\scalebox{0.5}{$\pm$} \scalebox{0.75}{7.02}}$ & \scalebox{0.95}{87.20}$^{\scalebox{0.5}{$\pm$} \scalebox{0.75}{3.15}}$ & \scalebox{0.95}{40.15}$^{\scalebox{0.5}{$\pm$} \scalebox{0.75}{0.82}}$ & \scalebox{0.95}{91.47}$^{\scalebox{0.5}{$\pm$} \scalebox{0.75}{0.14}}$ & \scalebox{0.95}{46.11}$^{\scalebox{0.5}{$\pm$} \scalebox{0.75}{1.86}}$ & \scalebox{0.95}{88.74}$^{\scalebox{0.5}{$\pm$} \scalebox{0.75}{0.57}}$ & \scalebox{0.95}{60.34}$^{\scalebox{0.5}{$\pm$} \scalebox{0.75}{0.14}}$ & \scalebox{0.95}{83.92}$^{\scalebox{0.5}{$\pm$} \scalebox{0.75}{0.42}}$ & \scalebox{0.95}{51.47}$^{\scalebox{0.5}{$\pm$} \scalebox{0.75}{1.53}}$ & \scalebox{0.95}{87.84}$^{\scalebox{0.5}{$\pm$} \scalebox{0.75}{0.58}}$ \\[0.1em]

NegPrompt$^\dagger$  & \scalebox{0.95}{65.03}$^{\scalebox{0.5}{$\pm$} \scalebox{0.75}{8.69}}$ & \scalebox{0.95}{84.56}$^{\scalebox{0.5}{$\pm$} \scalebox{0.75}{2.52}}$ & \scalebox{0.95}{44.39}$^{\scalebox{0.5}{$\pm$} \scalebox{0.75}{1.66}}$ & \scalebox{0.95}{89.63}$^{\scalebox{0.5}{$\pm$} \scalebox{0.75}{0.66}}$ & \scalebox{0.95}{51.31}$^{\scalebox{0.5}{$\pm$} \scalebox{0.75}{6.21}}$ & \scalebox{0.95}{86.55}$^{\scalebox{0.5}{$\pm$} \scalebox{0.75}{2.19}}$ & \scalebox{0.95}{87.60}$^{\scalebox{0.5}{$\pm$} \scalebox{0.75}{1.61}}$ & \scalebox{0.95}{63.76}$^{\scalebox{0.5}{$\pm$} \scalebox{0.75}{3.02}}$ & \scalebox{0.95}{62.08}$^{\scalebox{0.5}{$\pm$} \scalebox{0.75}{3.71}}$ & \scalebox{0.95}{81.13}$^{\scalebox{0.5}{$\pm$} \scalebox{0.75}{1.78}}$ \\[0.1em]

SCT$_{\scalebox{0.7}{MCM}}$  & 
\scalebox{0.95}{41.93}$^{\scalebox{0.5}{$\pm$} \scalebox{0.75}{12.17}}$ & \scalebox{0.95}{91.77}$^{\scalebox{0.5}{$\pm$} \scalebox{0.75}{2.40}}$ & \scalebox{0.95}{30.39}$^{\scalebox{0.5}{$\pm$} \scalebox{0.75}{2.17}}$ & \scalebox{0.95}{93.76}$^{\scalebox{0.5}{$\pm$} \scalebox{0.75}{0.39}}$ & \scalebox{0.95}{38.73}$^{\scalebox{0.5}{$\pm$} \scalebox{0.75}{2.54}}$ & \scalebox{0.95}{90.78}$^{\scalebox{0.5}{$\pm$} \scalebox{0.75}{0.34}}$ & \scalebox{0.95}{46.78}$^{\scalebox{0.5}{$\pm$} \scalebox{0.75}{3.51}}$ & \scalebox{0.95}{88.96}$^{\scalebox{0.5}{$\pm$} \scalebox{0.75}{1.16}}$ & \scalebox{0.95}{39.46}$^{\scalebox{0.5}{$\pm$} \scalebox{0.75}{3.39}}$ & \scalebox{0.95}{91.32}$^{\scalebox{0.5}{$\pm$} \scalebox{0.75}{0.90}}$ \\

SCT$_{\scalebox{0.7}{GL}}$  & 
\scalebox{0.95}{20.57}$^{\scalebox{0.5}{$\pm$} \scalebox{0.75}{10.2}}$ & \scalebox{0.95}{\underline{95.63}}$^{\scalebox{0.5}{$\pm$} \scalebox{0.75}{1.89}}$ & \textbf{\scalebox{0.95}{24.56}}$^{\scalebox{0.5}{$\pm$} \scalebox{0.75}{3.03}}$ & \scalebox{0.95}{\underline{94.39}}$^{\scalebox{0.5}{$\pm$} \scalebox{0.75}{0.53}}$ & \scalebox{0.95}{\underline{33.27}}$^{\scalebox{0.5}{$\pm$} \scalebox{0.75}{2.96}}$ & \scalebox{0.95}{91.27}$^{\scalebox{0.5}{$\pm$} \scalebox{0.75}{0.57}}$ & \scalebox{0.95}{48.12}$^{\scalebox{0.5}{$\pm$} \scalebox{0.75}{2.97}}$ & \scalebox{0.95}{86.76}$^{\scalebox{0.5}{$\pm$} \scalebox{0.75}{0.89}}$ & \scalebox{0.95}{31.62}$^{\scalebox{0.5}{$\pm$} \scalebox{0.75}{3.19}}$ & \scalebox{0.95}{92.01}$^{\scalebox{0.5}{$\pm$} \scalebox{0.75}{0.77}}$ \\[0.1em]

\rowcolor{gray!20}\scalebox{0.95}{FA}$_{\scalebox{0.68}{MCM}}$\scalebox{0.95}{(Ours)} & \scalebox{0.95}{25.50}$^{\scalebox{0.5}{$\pm$} \scalebox{0.75}{2.72}}$ & \scalebox{0.95}{94.72}$^{\scalebox{0.5}{$\pm$} \scalebox{0.75}{0.27}}$ & \scalebox{0.95}{36.24}$^{\scalebox{0.5}{$\pm$} \scalebox{0.75}{3.15}}$ & \scalebox{0.95}{92.40}$^{\scalebox{0.5}{$\pm$} \scalebox{0.75}{0.95}}$ & \scalebox{0.95}{35.38}$^{\scalebox{0.5}{$\pm$} \scalebox{0.75}{2.75}}$ & \textbf{\scalebox{0.95}{91.99}}$^{\scalebox{0.5}{$\pm$} \scalebox{0.75}{0.68}}$ & \textbf{\scalebox{0.95}{28.34}}$^{\scalebox{0.5}{$\pm$} \scalebox{0.75}{0.87}}$ & \textbf{\scalebox{0.95}{93.95}}$^{\scalebox{0.5}{$\pm$} \scalebox{0.75}{0.19}}$ & \scalebox{0.95}{\underline{31.37}}$^{\scalebox{0.5}{$\pm$} \scalebox{0.75}{1.07}}$ & \scalebox{0.95}{\underline{93.25}}$^{\scalebox{0.5}{$\pm$} \scalebox{0.75}{0.32}}$ \\

\rowcolor{gray!20}\scalebox{0.95}{FA}$_{\scalebox{0.7}{GL}}$(Ours) & \textbf{\scalebox{0.95}{14.12}}$^{\scalebox{0.5}{$\pm$} \scalebox{0.75}{1.32}}$ & \textbf{\scalebox{0.95}{96.76}}$^{\scalebox{0.5}{$\pm$} \scalebox{0.75}{0.10}}$ & \scalebox{0.95}{29.99}$^{\scalebox{0.5}{$\pm$} \scalebox{0.75}{1.57}}$ & \scalebox{0.95}{92.95}$^{\scalebox{0.5}{$\pm$} \scalebox{0.75}{0.66}}$ & \textbf{\scalebox{0.95}{32.48}}$^{\scalebox{0.5}{$\pm$} \scalebox{0.75}{1.48}}$ & \scalebox{0.95}{\underline{91.83}}$^{\scalebox{0.5}{$\pm$} \scalebox{0.75}{0.49}}$ & \scalebox{0.95}{34.66}$^{\scalebox{0.5}{$\pm$} \scalebox{0.75}{1.21}}$ & \scalebox{0.95}{91.50}$^{\scalebox{0.5}{$\pm$} \scalebox{0.75}{0.36}}$ & \textbf{\scalebox{0.95}{27.81}}$^{\scalebox{0.5}{$\pm$} \scalebox{0.75}{0.44}}$ & \textbf{\scalebox{0.95}{93.26}}$^{\scalebox{0.5}{$\pm$} \scalebox{0.75}{0.27}}$ \\

\midrule

\multicolumn{5}{l}{\textit{Prompt learning based methods}} &
\multicolumn{6}{l}{\, 16-shot} \\

CoOp$_{\scalebox{0.7}{MCM}}$  & \scalebox{0.95}{30.26}$^{\scalebox{0.5}{$\pm$} \scalebox{0.75}{1.98}}$ & \scalebox{0.95}{93.43}$^{\scalebox{0.5}{$\pm$} \scalebox{0.75}{0.81}}$ & \scalebox{0.95}{34.69}$^{\scalebox{0.5}{$\pm$} \scalebox{0.75}{0.43}}$ & \scalebox{0.95}{92.59}$^{\scalebox{0.5}{$\pm$} \scalebox{0.75}{0.14}}$ & \scalebox{0.95}{41.91}$^{\scalebox{0.5}{$\pm$} \scalebox{0.75}{0.71}}$ & \scalebox{0.95}{90.11}$^{\scalebox{0.5}{$\pm$} \scalebox{0.75}{0.23}}$ & \scalebox{0.95}{44.68}$^{\scalebox{0.5}{$\pm$} \scalebox{0.75}{2.11}}$ & \scalebox{0.95}{89.95}$^{\scalebox{0.5}{$\pm$} \scalebox{0.75}{0.48}}$ & \scalebox{0.95}{37.89}$^{\scalebox{0.5}{$\pm$} \scalebox{0.75}{0.71}}$ & \scalebox{0.95}{91.52}$^{\scalebox{0.5}{$\pm$} \scalebox{0.75}{0.29}}$  \\

CoOp$_{\scalebox{0.7}{GL}}$  & \scalebox{0.95}{15.96}$^{\scalebox{0.5}{$\pm$} \scalebox{0.75}{1.67}}$ & \scalebox{0.95}{96.11}$^{\scalebox{0.5}{$\pm$} \scalebox{0.75}{0.55}}$ & \scalebox{0.95}{27.26}$^{\scalebox{0.5}{$\pm$} \scalebox{0.75}{1.99}}$ & \scalebox{0.95}{93.29}$^{\scalebox{0.5}{$\pm$} \scalebox{0.75}{0.46}}$ & \scalebox{0.95}{35.36}$^{\scalebox{0.5}{$\pm$} \scalebox{0.75}{2.08}}$ & \scalebox{0.95}{90.58}$^{\scalebox{0.5}{$\pm$} \scalebox{0.75}{0.64}}$ & \scalebox{0.95}{48.63}$^{\scalebox{0.5}{$\pm$} \scalebox{0.75}{2.11}}$ & \scalebox{0.95}{86.11}$^{\scalebox{0.5}{$\pm$} \scalebox{0.75}{0.59}}$ & \scalebox{0.95}{31.81}$^{\scalebox{0.5}{$\pm$} \scalebox{0.75}{1.27}}$ & \scalebox{0.95}{91.51}$^{\scalebox{0.5}{$\pm$} \scalebox{0.75}{0.39}}$ \\[0.1em]

\scalebox{0.92}{LoCoOp}$_{\scalebox{0.7}{MCM}}$  & \scalebox{0.95}{27.35}$^{\scalebox{0.5}{$\pm$} \scalebox{0.75}{3.19}}$ & \scalebox{0.95}{94.12}$^{\scalebox{0.5}{$\pm$} \scalebox{0.75}{0.80}}$ & \scalebox{0.95}{30.93}$^{\scalebox{0.5}{$\pm$} \scalebox{0.75}{1.19}}$ & \scalebox{0.95}{93.75}$^{\scalebox{0.5}{$\pm$} \scalebox{0.75}{0.26}}$ & \scalebox{0.95}{38.26}$^{\scalebox{0.5}{$\pm$} \scalebox{0.75}{1.53}}$ & \scalebox{0.95}{91.12}$^{\scalebox{0.5}{$\pm$} \scalebox{0.75}{0.27}}$ & \scalebox{0.95}{41.36}$^{\scalebox{0.5}{$\pm$} \scalebox{0.75}{2.56}}$ & \scalebox{0.95}{90.99}$^{\scalebox{0.5}{$\pm$} \scalebox{0.75}{0.53}}$ & \scalebox{0.95}{34.47}$^{\scalebox{0.5}{$\pm$} \scalebox{0.75}{0.73}}$ & \scalebox{0.95}{92.49}$^{\scalebox{0.5}{$\pm$} \scalebox{0.75}{0.14}}$ \\

\scalebox{0.92}{LoCoOp}$_{\scalebox{0.7}{GL}}$  & \scalebox{0.95}{18.46}$^{\scalebox{0.5}{$\pm$} \scalebox{0.75}{1.38}}$ & \scalebox{0.95}{95.85}$^{\scalebox{0.5}{$\pm$} \scalebox{0.75}{0.63}}$ & \scalebox{0.95}{\underline{22.43}}$^{\scalebox{0.5}{$\pm$} \scalebox{0.75}{0.96}}$ & \scalebox{0.95}{\underline{95.15}}$^{\scalebox{0.5}{$\pm$} \scalebox{0.75}{0.22}}$ & \scalebox{0.95}{31.53}$^{\scalebox{0.5}{$\pm$} \scalebox{0.75}{1.52}}$ & \scalebox{0.95}{92.15}$^{\scalebox{0.5}{$\pm$} \scalebox{0.75}{0.22}}$ & \scalebox{0.95}{43.35}$^{\scalebox{0.5}{$\pm$} \scalebox{0.75}{3.12}}$ & \scalebox{0.95}{89.38}$^{\scalebox{0.5}{$\pm$} \scalebox{0.75}{0.78}}$ & \scalebox{0.95}{28.94}$^{\scalebox{0.5}{$\pm$} \scalebox{0.75}{1.29}}$ & \scalebox{0.95}{93.13}$^{\scalebox{0.5}{$\pm$} \scalebox{0.75}{0.17}}$ \\[0.1em]

IDLike  & \scalebox{0.95}{19.23}$^{\scalebox{0.5}{$\pm$} \scalebox{0.75}{10.5}}$ & \scalebox{0.95}{\underline{96.70}}$^{\scalebox{0.5}{$\pm$} \scalebox{0.75}{1.58}}$ & \scalebox{0.95}{54.15}$^{\scalebox{0.5}{$\pm$} \scalebox{0.75}{2.88}}$ & \scalebox{0.95}{87.64}$^{\scalebox{0.5}{$\pm$} \scalebox{0.75}{1.19}}$ & \scalebox{0.95}{56.63}$^{\scalebox{0.5}{$\pm$} \scalebox{0.75}{0.07}}$ & \scalebox{0.95}{85.86}$^{\scalebox{0.5}{$\pm$} \scalebox{0.75}{0.44}}$ & \scalebox{0.95}{34.69}$^{\scalebox{0.5}{$\pm$} \scalebox{0.75}{6.41}}$ & \scalebox{0.95}{91.90}$^{\scalebox{0.5}{$\pm$} \scalebox{0.75}{2.31}}$ & \scalebox{0.95}{41.18}$^{\scalebox{0.5}{$\pm$} \scalebox{0.75}{1.73}}$ & \scalebox{0.95}{90.53}$^{\scalebox{0.5}{$\pm$} \scalebox{0.75}{0.01}}$ \\[0.1em]

LSN$^\dagger$  & \scalebox{0.95}{36.17}$^{\scalebox{0.5}{$\pm$} \scalebox{0.75}{4.81}}$ & \scalebox{0.95}{92.66}$^{\scalebox{0.5}{$\pm$} \scalebox{0.75}{1.16}}$ & \scalebox{0.95}{34.27}$^{\scalebox{0.5}{$\pm$} \scalebox{0.75}{0.44}}$ & \scalebox{0.95}{93.53}$^{\scalebox{0.5}{$\pm$} \scalebox{0.75}{0.20}}$ & \scalebox{0.95}{41.47}$^{\scalebox{0.5}{$\pm$} \scalebox{0.75}{0.85}}$ & \scalebox{0.95}{90.52}$^{\scalebox{0.5}{$\pm$} \scalebox{0.75}{0.37}}$ & \scalebox{0.95}{46.43}$^{\scalebox{0.5}{$\pm$} \scalebox{0.75}{0.60}}$ & \scalebox{0.95}{89.38}$^{\scalebox{0.5}{$\pm$} \scalebox{0.75}{0.24}}$ & \scalebox{0.95}{39.58}$^{\scalebox{0.5}{$\pm$} \scalebox{0.75}{0.73}}$ & \scalebox{0.95}{91.53}$^{\scalebox{0.5}{$\pm$} \scalebox{0.75}{0.09}}$ \\[0.1em]

NegPrompt$^\dagger$  & \scalebox{0.95}{37.79}$^{\scalebox{0.5}{$\pm$} \scalebox{0.75}{0.11}}$ & \scalebox{0.95}{90.49}$^{\scalebox{0.5}{$\pm$} \scalebox{0.75}{0.01}}$ & \scalebox{0.95}{32.11}$^{\scalebox{0.5}{$\pm$} \scalebox{0.75}{3.77}}$ & \scalebox{0.95}{92.25}$^{\scalebox{0.5}{$\pm$} \scalebox{0.75}{1.00}}$ & \scalebox{0.95}{35.52}$^{\scalebox{0.5}{$\pm$} \scalebox{0.75}{0.41}}$ & \scalebox{0.95}{91.16}$^{\scalebox{0.5}{$\pm$} \scalebox{0.75}{0.03}}$ & \scalebox{0.95}{43.93}$^{\scalebox{0.5}{$\pm$} \scalebox{0.75}{9.09}}$ & \scalebox{0.95}{88.38}$^{\scalebox{0.5}{$\pm$} \scalebox{0.75}{3.31}}$ & \scalebox{0.95}{37.34}$^{\scalebox{0.5}{$\pm$} \scalebox{0.75}{1.41}}$ & \scalebox{0.95}{90.57}$^{\scalebox{0.5}{$\pm$} \scalebox{0.75}{0.59}}$ \\[0.1em]

SCT$_{\scalebox{0.7}{MCM}}$ & 
\scalebox{0.95}{29.41}$^{\scalebox{0.5}{$\pm$} \scalebox{0.75}{2.19}}$ & \scalebox{0.95}{93.76}$^{\scalebox{0.5}{$\pm$} \scalebox{0.75}{0.55}}$ & \scalebox{0.95}{27.28}$^{\scalebox{0.5}{$\pm$} \scalebox{0.75}{2.80}}$ & \scalebox{0.95}{94.22}$^{\scalebox{0.5}{$\pm$} \scalebox{0.75}{0.40}}$ & \scalebox{0.95}{36.35}$^{\scalebox{0.5}{$\pm$} \scalebox{0.75}{2.13}}$ & \scalebox{0.95}{91.16}$^{\scalebox{0.5}{$\pm$} \scalebox{0.75}{0.33}}$ & \scalebox{0.95}{42.25}$^{\scalebox{0.5}{$\pm$} \scalebox{0.75}{1.89}}$ & \scalebox{0.95}{90.52}$^{\scalebox{0.5}{$\pm$} \scalebox{0.75}{0.48}}$ & \scalebox{0.95}{33.82}$^{\scalebox{0.5}{$\pm$} \scalebox{0.75}{1.78}}$ & \scalebox{0.95}{92.42}$^{\scalebox{0.5}{$\pm$} \scalebox{0.75}{0.35}}$ \\

SCT$_{\scalebox{0.7}{GL}}$ & 
\scalebox{0.95}{\underline{15.19}}$^{\scalebox{0.5}{$\pm$} \scalebox{0.75}{2.16}}$ & \scalebox{0.95}{\textbf{96.71}}$^{\scalebox{0.5}{$\pm$} \scalebox{0.75}{0.44}}$ & \textbf{\scalebox{0.95}{20.00}}$^{\scalebox{0.5}{$\pm$} \scalebox{0.75}{0.61}}$&
\textbf{\scalebox{0.95}{95.57}}$^{\scalebox{0.5}{$\pm$} \scalebox{0.75}{0.11}}$&
\textbf{\scalebox{0.95}{29.71}}$^{\scalebox{0.5}{$\pm$} \scalebox{0.75}{0.85}}$&
\scalebox{0.95}{92.37}$^{\scalebox{0.5}{$\pm$} \scalebox{0.75}{0.12}}$ & \scalebox{0.95}{44.17}$^{\scalebox{0.5}{$\pm$} \scalebox{0.75}{0.82}}$ & \scalebox{0.95}{88.59}$^{\scalebox{0.5}{$\pm$} \scalebox{0.75}{0.39}}$ & \scalebox{0.95}{\underline{27.27}}$^{\scalebox{0.5}{$\pm$} \scalebox{0.75}{0.44}}$ & \scalebox{0.95}{93.31}$^{\scalebox{0.5}{$\pm$} \scalebox{0.75}{0.17}}$ \\[0.1em]

\rowcolor{gray!20} \scalebox{0.95}{FA}$_{\scalebox{0.68}{MCM}}$\scalebox{0.95}{(Ours)} & \scalebox{0.95}{25.79}$^{\scalebox{0.5}{$\pm$} \scalebox{0.75}{1.48}}$ & \scalebox{0.95}{94.29}$^{\scalebox{0.5}{$\pm$} \scalebox{0.75}{0.35}}$ & \scalebox{0.95}{33.54}$^{\scalebox{0.5}{$\pm$} \scalebox{0.75}{1.17}}$ & \scalebox{0.95}{93.02}$^{\scalebox{0.5}{$\pm$} \scalebox{0.75}{0.19}}$ & \scalebox{0.95}{33.77}$^{\scalebox{0.5}{$\pm$} \scalebox{0.75}{1.64}}$ & \textbf{\scalebox{0.95}{92.64}}$^{\scalebox{0.5}{$\pm$} \scalebox{0.75}{0.42}}$ & \textbf{\scalebox{0.95}{23.17}}$^{\scalebox{0.5}{$\pm$} \scalebox{0.75}{1.31}}$ & \textbf{\scalebox{0.95}{95.14}}$^{\scalebox{0.5}{$\pm$} \scalebox{0.75}{0.26}}$ & \scalebox{0.95}{29.07}$^{\scalebox{0.5}{$\pm$} \scalebox{0.75}{1.11}}$ & \scalebox{0.95}{\underline{93.77}}$^{\scalebox{0.5}{$\pm$} \scalebox{0.75}{0.19}}$ \\

\rowcolor{gray!20} \scalebox{0.95}{FA}$_{\scalebox{0.7}{GL}}$(Ours)  & \textbf{\scalebox{0.95}{14.49}}$^{\scalebox{0.5}{$\pm$} \scalebox{0.75}{1.27}}$ & \scalebox{0.95}{96.48}$^{\scalebox{0.5}{$\pm$} \scalebox{0.75}{0.29}}$ & \scalebox{0.95}{27.65}$^{\scalebox{0.5}{$\pm$} \scalebox{0.75}{1.08}}$ & \scalebox{0.95}{93.46}$^{\scalebox{0.5}{$\pm$} \scalebox{0.75}{0.18}}$ & \scalebox{0.95}{\underline{31.09}}$^{\scalebox{0.5}{$\pm$} \scalebox{0.75}{1.38}}$ & \scalebox{0.95}{\underline{92.44}}$^{\scalebox{0.5}{$\pm$} \scalebox{0.75}{0.34}}$ & \scalebox{0.95}{\underline{29.50}}$^{\scalebox{0.5}{$\pm$} \scalebox{0.75}{0.89}}$ & \scalebox{0.95}{\underline{92.93}}$^{\scalebox{0.5}{$\pm$} \scalebox{0.75}{0.18}}$ & \textbf{\scalebox{0.95}{25.68}}$^{\scalebox{0.5}{$\pm$} \scalebox{0.75}{0.58}}$ & \textbf{\scalebox{0.95}{93.82}}$^{\scalebox{0.5}{$\pm$} \scalebox{0.75}{0.11}}$ \\
\bottomrule
\end{tabularx}

\vspace{-0.5em}
\caption{Comparison results on ImageNet-1k OOD benchmarks. All results using the same backbone ViT-B/16. The results marked with $^\dagger$ are taken from \cite{sct}; the others are our reproductions. The prompt learning based methods are run under four trials, reporting the mean and standard deviation of the performance. The subscripts $_{\text{MCM}}$ and $_{\text{GL}}$ indicate the use of the MCM score and the GL-MCM score. The best and second-best results are indicated in bold and \underline{underline}. $\uparrow$ indicates larger values are better; $\downarrow$ indicates smaller values are better. All values are percentages.}
\label{tab1}
\vspace{-0.5em}
\end{table*}

\begin{table}[ht]
\centering

\begin{tabular}{lccc}
\toprule
Method & 1-shot & 4-shot & 16-shot \\
\midrule
CoOp & \scalebox{0.95}{67.44}$^{\scalebox{0.6}{$\pm$} {\scalebox{0.75}{0.50}}}$ & \scalebox{0.95}{69.71}$^{\scalebox{0.6}{$\pm$} {\scalebox{0.75}{0.07}}}$ & \scalebox{0.95}{70.99}$^{\scalebox{0.6}{$\pm$} {\scalebox{0.75}{0.14}}}$    \\

\scalebox{0.92}{LoCoOp} & \scalebox{0.95}{67.40}$^{\scalebox{0.6}{$\pm$} {\scalebox{0.75}{0.64}}}$ & \scalebox{0.95}{69.55}$^{\scalebox{0.6}{$\pm$} {\scalebox{0.75}{0.10}}}$ & \scalebox{0.95}{\underline{71.53}}$^{\scalebox{0.6}{$\pm$} {\scalebox{0.75}{0.17}}}$    \\

IDLike & \scalebox{0.95}{68.17}$^{\scalebox{0.6}{$\pm$} {\scalebox{0.75}{0.57}}}$ & \scalebox{0.95}{68.91}$^{\scalebox{0.6}{$\pm$} {\scalebox{0.75}{0.14}}}$ & \scalebox{0.95}{69.46}$^{\scalebox{0.6}{$\pm$} {\scalebox{0.75}{0.02}}}$    \\

SCT & \scalebox{0.95}{\underline{68.63}}$^{\scalebox{0.6}{$\pm$} {\scalebox{0.75}{0.13}}}$ & \scalebox{0.95}{\underline{69.93}}$^{\scalebox{0.6}{$\pm$} {\scalebox{0.75}{0.22}}}$ & \textbf{\scalebox{0.95}{71.78}}$^{\scalebox{0.6}{$\pm$} {\scalebox{0.75}{0.05}}}$    \\

\rowcolor{gray!20}FA(Ours) & \textbf{\scalebox{0.95}{68.67}}$^{\scalebox{0.6}{$\pm$} {\scalebox{0.75}{0.39}}}$ & \textbf{\scalebox{0.95}{69.96}}$^{\scalebox{0.6}{$\pm$} {\scalebox{0.75}{0.04}}}$ & \scalebox{0.95}{71.02}$^{\scalebox{0.6}{$\pm$} {\scalebox{0.75}{0.09}}}$    \\
\bottomrule
\end{tabular}

\vspace{-0.5em}
\caption{Comparison results in ID Top-1 accuracy on ImageNet-1k under different few-shot settings. Other notations are the same as \cref{tab1}.}

\label{tab-imagenet1k-acc}
\vspace{-1.9em}
\end{table}

\begin{table*}[ht]
\centering

\footnotesize

\begin{tabularx}{\textwidth}{l*{9}{>{\centering\arraybackslash}X}}
% \begin{tabularx}{\textwidth}{>{\raggedright\arraybackslash}m{1.30cm}*{11}{>{\centering\arraybackslash}m{1.1cm}}}
\toprule

\multirow{2}{*}{\makecell[c]{Method} } &

\multicolumn{2}{c}{\scalebox{0.925}{OpenImage-O}} & \multicolumn{2}{c}{\scalebox{0.925}{NINCO}} & \multicolumn{2}{c}{\scalebox{0.925}{ImageNet-O}}  \vrule  & \multicolumn{2}{c} {\scalebox{0.925}{Average}} \\

& \scalebox{0.925}{FPR95}$\downarrow$ & \scalebox{0.925}{AUROC}$\uparrow$ & \scalebox{0.925}{FPR95}$\downarrow$ & \scalebox{0.925}{AUROC}$\uparrow$ & \scalebox{0.925}{FPR95}$\downarrow$ & \scalebox{0.925}{AUROC}$\uparrow$ & \scalebox{0.925}{FPR95}$\downarrow$ & \scalebox{0.925}{AUROC}$\uparrow$   \\
\midrule

CoOp$_{\scalebox{0.7}{MCM}}$ & 
37.95$^{\scalebox{0.6}{$\pm$} 0.59}$ & 91.99$^{\scalebox{0.6}{$\pm$} 0.22}$ & 
78.00$^{\scalebox{0.6}{$\pm$} 0.23}$ & 73.00$^{\scalebox{0.6}{$\pm$} 0.59}$ & 
70.14$^{\scalebox{0.6}{$\pm$} 0.86}$ & \underline{81.34}$^{\scalebox{0.6}{$\pm$} 0.83}$ & 
62.03$^{\scalebox{0.6}{$\pm$} 0.38}$ & 82.11$^{\scalebox{0.6}{$\pm$} 0.48}$   \\

CoOp$_{\scalebox{0.7}{GL}}$ & 
\underline{32.23}$^{\scalebox{0.6}{$\pm$} 1.00}$ & 92.22$^{\scalebox{0.6}{$\pm$} 0.25}$ & 
72.04$^{\scalebox{0.6}{$\pm$} 0.81}$ & 75.35$^{\scalebox{0.6}{$\pm$} 0.88}$ & 
67.35$^{\scalebox{0.6}{$\pm$} 1.19}$ & 78.44$^{\scalebox{0.6}{$\pm$} 0.94}$ & 
\underline{57.21}$^{\scalebox{0.6}{$\pm$} 0.96}$ & 81.99$^{\scalebox{0.6}{$\pm$} 0.58}$  \\

LoCoOp$_{\scalebox{0.7}{MCM}}$ & 
37.03$^{\scalebox{0.6}{$\pm$} 0.94}$ & 92.24$^{\scalebox{0.6}{$\pm$} 0.21}$ & 
77.96$^{\scalebox{0.6}{$\pm$} 0.63}$ & 72.78$^{\scalebox{0.6}{$\pm$} 0.79}$ & 
70.96$^{\scalebox{0.6}{$\pm$} 0.75}$ & 81.25$^{\scalebox{0.6}{$\pm$} 0.39}$ & 
61.98$^{\scalebox{0.6}{$\pm$} 0.55}$ & 82.09$^{\scalebox{0.6}{$\pm$} 0.35}$  \\

LoCoOp$_{\scalebox{0.7}{GL}}$ & 
33.87$^{\scalebox{0.6}{$\pm$} 1.23}$ & \underline{92.53}$^{\scalebox{0.6}{$\pm$} 0.07}$ & 
75.16$^{\scalebox{0.6}{$\pm$} 0.92}$ & 72.97$^{\scalebox{0.6}{$\pm$} 0.39}$ & 
68.06$^{\scalebox{0.6}{$\pm$} 1.92}$ & 81.19$^{\scalebox{0.6}{$\pm$} 0.25}$ & 
59.03$^{\scalebox{0.6}{$\pm$} 1.27}$ & 82.23$^{\scalebox{0.6}{$\pm$} 0.17}$ \\

IDLike & 
54.43$^{\scalebox{0.6}{$\pm$} 2.35}$ & 87.89$^{\scalebox{0.6}{$\pm$} 0.45}$ & 
78.93$^{\scalebox{0.6}{$\pm$} 1.79}$ & 69.32$^{\scalebox{0.6}{$\pm$} 0.42}$ & 
84.08$^{\scalebox{0.6}{$\pm$} 1.45}$ & 67.45$^{\scalebox{0.6}{$\pm$} 0.86}$ & 
72.48$^{\scalebox{0.6}{$\pm$} 0.89}$ & 74.89$^{\scalebox{0.6}{$\pm$} 0.28}$  \\

SCT$_{\scalebox{0.7}{MCM}}$ & 
37.28$^{\scalebox{0.6}{$\pm$} 1.16}$ & 92.04$^{\scalebox{0.6}{$\pm$} 0.33}$ & 
78.51$^{\scalebox{0.6}{$\pm$} 0.86}$ & 71.09$^{\scalebox{0.6}{$\pm$} 1.19}$ & 
71.45$^{\scalebox{0.6}{$\pm$} 0.54}$ & 81.29$^{\scalebox{0.6}{$\pm$} 0.15}$ & 
62.41$^{\scalebox{0.6}{$\pm$} 0.73}$ & 81.48$^{\scalebox{0.6}{$\pm$} 0.52}$  \\

SCT$_{\scalebox{0.7}{GL}}$ & 
32.24$^{\scalebox{0.6}{$\pm$} 0.34}$ & \textbf{92.75}$^{\scalebox{0.6}{$\pm$} 0.17}$ & 
74.15$^{\scalebox{0.6}{$\pm$} 0.53}$ & 73.14$^{\scalebox{0.6}{$\pm$} 0.94}$ & 
68.35$^{\scalebox{0.6}{$\pm$} 0.62}$ & 80.84$^{\scalebox{0.6}{$\pm$} 0.45}$ & 
58.25$^{\scalebox{0.6}{$\pm$} 0.39}$ & 82.24$^{\scalebox{0.6}{$\pm$} 0.44}$  \\

\rowcolor{gray!20}FA$_{\scalebox{0.7}{MCM}}$(Ours) & 
37.93$^{\scalebox{0.6}{$\pm$} 1.65}$ & 91.84$^{\scalebox{0.6}{$\pm$} 0.45}$ & 
\underline{70.44}$^{\scalebox{0.6}{$\pm$} 1.24}$ & \underline{77.77}$^{\scalebox{0.6}{$\pm$} 0.94}$ & 
\underline{63.84}$^{\scalebox{0.6}{$\pm$} 1.36}$ & \textbf{82.71}$^{\scalebox{0.6}{$\pm$} 0.30}$ & 
57.40$^{\scalebox{0.6}{$\pm$} 1.34}$ & \textbf{84.08}$^{\scalebox{0.6}{$\pm$} 0.47}$  \\

\rowcolor{gray!20}FA$_{\scalebox{0.7}{GL}}$(Ours) & 
\textbf{32.10}$^{\scalebox{0.6}{$\pm$} 0.89}$ & 92.39$^{\scalebox{0.6}{$\pm$} 0.28}$ & \textbf{65.61}$^{\scalebox{0.6}{$\pm$} 1.32}$ & \textbf{79.01}$^{\scalebox{0.6}{$\pm$} 0.71}$ & \textbf{63.13}$^{\scalebox{0.6}{$\pm$} 0.98}$ & 80.39$^{\scalebox{0.6}{$\pm$} 0.21}$ & \textbf{53.61}$^{\scalebox{0.6}{$\pm$} 0.89}$ & \underline{83.93}$^{\scalebox{0.6}{$\pm$} 0.31}$  \\

\bottomrule
\end{tabularx}

\vspace{-0.5em}
\caption{Challenging OOD detection results on cleaner OOD datasets in the 16-shot setting. We use the same notation as \cref{tab1}. }
\label{tab-challenge}

\end{table*}

\subsection{Model inference}

During model inference, for the downstream classification task, we adopt the same strategy as CLIP, depending solely on the forced prompt~\cite{idlike}. 
% Moreover, we employ the MCM score and the GL-MCM score (as shown in Eq.\ref{eq3} and Eq. \ref{eq4}) using both the forced prompt and the original prompt for the OOD detection task.
For the OOD detection task, our method can be flexibly combined with different score functions, such as the MCM~\cite{mcm} score and the GL-MCM~\cite{glmcm} score.
%
% For the OOD detection task, we employ the MCM~\cite{mcm} score and the GL-MCM~\cite{glmcm} score, depending on both the forced prompt and the original prompt. 
The MCM score is defined as the maximum similarity between the global image features $\mathbf{z}^g$ (\ie $\mathbf{z}$) and all text features $\mathbf{t}_c^a$ (\ie the concatenation of text features $\mathbf{t}_c^f$ and $\mathbf{t}_c^o$) after applying softmax with temperature $\tau_0$, i.e.,
% \begin{equation}
% S_{\scriptscriptstyle \mathrm{MCM}}(\mathbf{x}) = \max_{c} \frac{e^{\cos\,(\mathbf{z}^g,\mathbf{t}_c^f) / \tau_0}}{\sum_{j=1}^C e^{\cos\,(\mathbf{z}^g,\mathbf{t}_j^f) / \tau_0} + Ke^{\cos\,(\mathbf{z}^g,\mathbf{t}_j^o) / \tau_0}  },
% \label{mcm}
% \end{equation}
\begin{equation}
S_{\scriptscriptstyle \mathrm{MCM}}(\mathbf{x}) = \max_{c} \frac{e^{\cos\,(\mathbf{z}^g,\mathbf{t}_c^a) / \tau_0}}{\sum_{j=1}^C e^{\cos\,(\mathbf{z}^g,\mathbf{t}_j^f) / \tau_0} + Ke^{\cos\,(\mathbf{z}^g,\mathbf{t}_j^o) / \tau_0}  },
\label{mcm}
\end{equation}
while the GL-MCM score simultaneously considers both global and local features, which can be expressed as
\begin{equation}
S_{\scriptscriptstyle \text{GL-MCM}}(\mathbf{x}) = S_{\scriptscriptstyle \text{MCM}}(\mathbf{x}) + S_{\scriptscriptstyle \text{L-MCM}}(\mathbf{x}),
\label{glmcm}
\end{equation}
\begin{equation}
S_{\scriptscriptstyle \text{L-MCM}}(\mathbf{x}) = \max_{i,c} \frac{e^{\cos\,(\mathbf{z}^l_i,\mathbf{t}_c^a) / \tau_0}}{\sum_{j=1}^C e^{\cos\,(\mathbf{z}^l_i,\mathbf{t}_j^f) / \tau_0}+Ke^{\cos\,(\mathbf{z}^l_i,\mathbf{t}_j^o) / \tau_0}},
\label{l-mcm}
\end{equation}
where $\mathbf{z}^l_i(i\in\{1,\cdots,N\})$ represents $N$ extracted local features generated by CLIP's image encoder~\cite{glmcm,locoop}. 
We set $\tau_0 = 1$ during model inference.

% \begin{equation}
% S_{\scriptscriptstyle \text{GL-MCM}}(\mathbf{x}) = S_{\scriptscriptstyle \text{MCM}}(\mathbf{x}) + \max_{i,c} \frac{e^{\cos\,(\mathbf{z}^l_i,\mathbf{t}_c^F) / \tau}}{\sum_{j=1}^C e^{\cos\,(\mathbf{z}^l_i,\mathbf{t}_j^F) / \tau}+e^{\cos\,(\mathbf{z}^l_i,\mathbf{t}_j^O) / \tau}},
% \label{eq4}
% \end{equation}
% Common confidence score functions used in CLIP-based OOD detection methods include MCM score and GL-MCM score. The MCM score is defined as the maximum similarity between the global image features $\mathbf{z}^g$ and the text features $t_c$ after applying softmax with temperature $\tau$, as follows:
% \begin{equation}
% S_{\scriptscriptstyle \mathrm{MCM}}(\mathbf{x}) = \max_{c} \frac{e^{\cos\,(\mathbf{z}^g,\mathbf{t}_c) / \tau}}{\sum_{j=1}^C e^{\cos\,(\mathbf{z}^g,\mathbf{t}_j) / \tau}},
% \label{eq3}
% \end{equation}
% while the GL-MCM score simultaneously considers both global and local features, which can be expressed as:

% \begin{equation}
% S_{\scriptscriptstyle \text{GL-MCM}}(\mathbf{x}) = S_{\scriptscriptstyle \text{MCM}}(\mathbf{x}) + \max_{i,c} \frac{e^{\cos\,(\mathbf{z}^l_i,\mathbf{t}_c) / \tau}}{\sum_{j=1}^C e^{\cos\,(\mathbf{z}^l_i,\mathbf{t}_j) / \tau}},
% \label{eq4}
% \end{equation}
% where $\mathbf{z}^l_i(i\in\{1,\cdots,N\})$ represents $N$ extracted local features generated by CLIP's image encoder. 
% We set $\tau = 1$ during model inference.

%-------------------------------------------------------------------------

\section{Experiments}

%-------------------------------------------------------------------------
\subsection{Experimental details}

\textbf{Datasets.}\; We use a popular benchmark for conventional OOD detection~\cite{mos,mcm,locoop,glmcm}, where ImageNet-1k~\cite{imagenet1k} serves as the ID dataset, and the OOD datasets are iNaturalist~\cite{iNaturalist}, SUN~\cite{SUN}, Places~\cite{Places}, and Texture~\cite{TEXTURE}. 
Moreover, inspired by existing studies~\cite{NINCO,idlike}, we use datasets like OpenImage-O~\cite{OpenImage-O}, NINCO~\cite{NINCO}, and ImageNet-O~\cite{ImageNet-O}, which are cleaner and more realistic, to simulate the more challenging OOD detection. 
%  to fully substantiate the effectiveness of our FA,
Besides, we also utilize other widely adopted datasets for few-shot settings as ID datasets~\cite{coop,taskres,tip_adapter}, which include StandfordCars~\cite{StanfordCars}, UCF101~\cite{UCF101},Caltech101~\cite{Caltech101}, Flowers102~\cite{Flowers102},  EuroSAT~\cite{EuroSAT}, 
FGVCAircraft~\cite{FGVCAircraft}, OxfordPets~\cite{OxfordPets}, and Food101~\cite{Food101}, considering factors such as image resolution and the number of classes.
% During testing, we adopt the complete ID and OOD test sets for evaluation.

\begin{table*}[ht]
\centering

\footnotesize

\begin{tabularx}{\textwidth}{>{\centering\arraybackslash}p{1.15cm} >{\raggedright\arraybackslash}p{1.15cm} *{10}{>{\centering\arraybackslash}X}}
% \begin{tabularx}{\textwidth}{>{\raggedright\arraybackslash}m{1.30cm}*{11}{>{\centering\arraybackslash}m{1.1cm}}}
\toprule

\multirow{2}{*}{\makecell[c]{ID Dataset} } & \multirow{2}{*}{\makecell[c]{Method} } &
\multicolumn{2}{c}{iNaturalist} & \multicolumn{2}{c}{SUN} & \multicolumn{2}{c}{Places} & \multicolumn{2}{c}{Textures} \vrule  & \multicolumn{2}{c} {Average}  \\

 & & FPR95$\downarrow$ & AUROC$\uparrow$ & FPR95$\downarrow$ & AUROC$\uparrow$ & FPR95$\downarrow$ & AUROC$\uparrow$ & FPR95$\downarrow$ & AUROC$\uparrow$ & FPR95$\downarrow$ & AUROC$\uparrow$  \\
\midrule

 & CoOp$_{\scalebox{0.7}{MCM}}$ & \scalebox{0.95}{14.40}$^{\scalebox{0.5}{$\pm$} \scalebox{0.75}{6.61}}$ & \scalebox{0.95}{97.41}$^{\scalebox{0.5}{$\pm$} \scalebox{0.75}{1.36}}$ & \scalebox{0.95}{37.09}$^{\scalebox{0.5}{$\pm$} \scalebox{0.75}{3.94}}$ & \scalebox{0.95}{93.14}$^{\scalebox{0.5}{$\pm$} \scalebox{0.75}{0.48}}$ & \scalebox{0.95}{40.07}$^{\scalebox{0.5}{$\pm$} \scalebox{0.75}{2.77}}$ & \scalebox{0.95}{91.89}$^{\scalebox{0.5}{$\pm$} \scalebox{0.75}{0.65}}$ & \scalebox{0.95}{33.24}$^{\scalebox{0.5}{$\pm$} \scalebox{0.75}{4.21}}$ & \scalebox{0.95}{93.19}$^{\scalebox{0.5}{$\pm$} \scalebox{0.75}{1.22}}$ & \scalebox{0.95}{31.21}$^{\scalebox{0.5}{$\pm$} \scalebox{0.75}{3.58}}$ & \scalebox{0.95}{93.91}$^{\scalebox{0.5}{$\pm$} \scalebox{0.75}{0.45}}$  \\
 
& CoOp$_{\scalebox{0.7}{GL}}$ & \scalebox{0.95}{9.65}$^{\scalebox{0.5}{$\pm$} \scalebox{0.75}{6.35}}$ & \scalebox{0.95}{98.18}$^{\scalebox{0.5}{$\pm$} \scalebox{0.75}{0.98}}$ & \scalebox{0.95}{28.54}$^{\scalebox{0.5}{$\pm$} \scalebox{0.75}{5.07}}$ & \scalebox{0.95}{94.51}$^{\scalebox{0.5}{$\pm$} \scalebox{0.75}{0.92}}$ & \scalebox{0.95}{32.94}$^{\scalebox{0.5}{$\pm$} \scalebox{0.75}{1.74}}$ & \scalebox{0.95}{93.28}$^{\scalebox{0.5}{$\pm$} \scalebox{0.75}{0.32}}$ & \scalebox{0.95}{35.96}$^{\scalebox{0.5}{$\pm$} \scalebox{0.75}{2.65}}$ & \scalebox{0.95}{91.83}$^{\scalebox{0.5}{$\pm$} \scalebox{0.75}{1.42}}$ & \scalebox{0.95}{26.77}$^{\scalebox{0.5}{$\pm$} \scalebox{0.75}{3.22}}$ & \scalebox{0.95}{94.45}$^{\scalebox{0.5}{$\pm$} \scalebox{0.75}{0.42}}$ \\[0.1em]

& \scalebox{0.92}{LoCoOp}$_{\scalebox{0.55}{MCM}}$ & \scalebox{0.95}{2.38}$^{\scalebox{0.5}{$\pm$} \scalebox{0.75}{0.98}}$ & \scalebox{0.95}{99.21}$^{\scalebox{0.5}{$\pm$} \scalebox{0.75}{0.22}}$ & \scalebox{0.95}{19.73}$^{\scalebox{0.5}{$\pm$} \scalebox{0.75}{4.24}}$ & \scalebox{0.95}{96.11}$^{\scalebox{0.5}{$\pm$} \scalebox{0.75}{0.78}}$ & \scalebox{0.95}{22.57}$^{\scalebox{0.5}{$\pm$} \scalebox{0.75}{3.01}}$ & \scalebox{0.95}{94.96}$^{\scalebox{0.5}{$\pm$} \scalebox{0.75}{0.77}}$ & \scalebox{0.95}{15.89}$^{\scalebox{0.5}{$\pm$} \scalebox{0.75}{3.12}}$ & \scalebox{0.95}{96.12}$^{\scalebox{0.5}{$\pm$} \scalebox{0.75}{0.67}}$ & \scalebox{0.95}{15.14}$^{\scalebox{0.5}{$\pm$} \scalebox{0.75}{1.91}}$ & \scalebox{0.95}{96.60}$^{\scalebox{0.5}{$\pm$} \scalebox{0.75}{0.33}}$ \\

UCF101 & \scalebox{0.92}{LoCoOp}$_{\scalebox{0.55}{GL}}$ & \scalebox{0.95}{1.13}$^{\scalebox{0.5}{$\pm$} \scalebox{0.75}{0.22}}$ & \scalebox{0.95}{99.65}$^{\scalebox{0.5}{$\pm$} \scalebox{0.75}{0.09}}$ & \scalebox{0.95}{17.93}$^{\scalebox{0.5}{$\pm$} \scalebox{0.75}{4.09}}$ & \scalebox{0.95}{96.51}$^{\scalebox{0.5}{$\pm$} \scalebox{0.75}{0.89}}$ & \scalebox{0.95}{22.34}$^{\scalebox{0.5}{$\pm$} \scalebox{0.75}{4.58}}$ & \scalebox{0.95}{95.03}$^{\scalebox{0.5}{$\pm$} \scalebox{0.75}{1.06}}$ & \scalebox{0.95}{18.17}$^{\scalebox{0.5}{$\pm$} \scalebox{0.75}{3.01}}$ & \scalebox{0.95}{95.25}$^{\scalebox{0.5}{$\pm$} \scalebox{0.75}{0.71}}$ & \scalebox{0.95}{14.89}$^{\scalebox{0.5}{$\pm$} \scalebox{0.75}{2.49}}$ & \scalebox{0.95}{96.61}$^{\scalebox{0.5}{$\pm$} \scalebox{0.75}{0.58}}$ \\[0.1em]

& IDLike & \scalebox{0.95}{66.63}$^{\scalebox{0.5}{$\pm$} \scalebox{0.75}{27.6}}$ & \scalebox{0.95}{89.75}$^{\scalebox{0.5}{$\pm$} \scalebox{0.75}{5.52}}$ & \scalebox{0.95}{87.88}$^{\scalebox{0.5}{$\pm$} \scalebox{0.75}{6.56}}$ & \scalebox{0.95}{78.04}$^{\scalebox{0.5}{$\pm$} \scalebox{0.75}{6.14}}$ & \scalebox{0.95}{84.47}$^{\scalebox{0.5}{$\pm$} \scalebox{0.75}{12.1}}$ & \scalebox{0.95}{76.04}$^{\scalebox{0.5}{$\pm$} \scalebox{0.75}{8.53}}$ & \scalebox{0.95}{49.07}$^{\scalebox{0.5}{$\pm$} \scalebox{0.75}{11.2}}$ & \scalebox{0.95}{91.19}$^{\scalebox{0.5}{$\pm$} \scalebox{0.75}{2.74}}$ & \scalebox{0.95}{72.01}$^{\scalebox{0.5}{$\pm$} \scalebox{0.75}{14.0}}$ & \scalebox{0.95}{83.76}$^{\scalebox{0.5}{$\pm$} \scalebox{0.75}{5.64}}$ \\[0.1em]

& SCT$_{\scalebox{0.7}{MCM}}$ & 
\scalebox{0.95}{1.51}$^{\scalebox{0.5}{$\pm$} \scalebox{0.75}{0.51}}$ & \scalebox{0.95}{99.44}$^{\scalebox{0.5}{$\pm$} \scalebox{0.75}{0.10}}$ & \scalebox{0.95}{24.17}$^{\scalebox{0.5}{$\pm$} \scalebox{0.75}{5.54}}$ & \scalebox{0.95}{95.21}$^{\scalebox{0.5}{$\pm$} \scalebox{0.75}{0.88}}$ & \scalebox{0.95}{26.62}$^{\scalebox{0.5}{$\pm$} \scalebox{0.75}{4.41}}$ & \scalebox{0.95}{93.75}$^{\scalebox{0.5}{$\pm$} \scalebox{0.75}{0.81}}$ & \scalebox{0.95}{15.34}$^{\scalebox{0.5}{$\pm$} \scalebox{0.75}{3.84}}$ & \scalebox{0.95}{95.98}$^{\scalebox{0.5}{$\pm$} \scalebox{0.75}{0.64}}$ & \scalebox{0.95}{16.91}$^{\scalebox{0.5}{$\pm$} \scalebox{0.75}{3.20}}$ & \scalebox{0.95}{96.09}$^{\scalebox{0.5}{$\pm$} \scalebox{0.75}{0.48}}$ \\

& SCT$_{\scalebox{0.7}{GL}}$ & 
\scalebox{0.95}{0.79}$^{\scalebox{0.5}{$\pm$} \scalebox{0.75}{0.27}}$ & \scalebox{0.95}{99.75}$^{\scalebox{0.5}{$\pm$} \scalebox{0.75}{0.04}}$ & \scalebox{0.95}{18.69}$^{\scalebox{0.5}{$\pm$} \scalebox{0.75}{2.41}}$ & \scalebox{0.95}{95.86}$^{\scalebox{0.5}{$\pm$} \scalebox{0.75}{0.56}}$ & \scalebox{0.95}{23.77}$^{\scalebox{0.5}{$\pm$} \scalebox{0.75}{2.47}}$ & \scalebox{0.95}{93.98}$^{\scalebox{0.5}{$\pm$} \scalebox{0.75}{0.47}}$ & \scalebox{0.95}{17.44}$^{\scalebox{0.5}{$\pm$} \scalebox{0.75}{2.62}}$ & \scalebox{0.95}{94.52}$^{\scalebox{0.5}{$\pm$} \scalebox{0.75}{0.54}}$ & \scalebox{0.95}{15.17}$^{\scalebox{0.5}{$\pm$} \scalebox{0.75}{1.56}}$ & \scalebox{0.95}{96.03}$^{\scalebox{0.5}{$\pm$} \scalebox{0.75}{0.25}}$ \\[0.1em]

\rowcolor{gray!20} & \scalebox{0.9}{FA}$_{\scalebox{0.55}{MCM}}$\scalebox{0.9}{(Ours)} & \textbf{\scalebox{0.95}{0.12}}$^{\scalebox{0.5}{$\pm$} \scalebox{0.75}{0.11}}$ & \textbf{\scalebox{0.95}{99.94}}$^{\scalebox{0.5}{$\pm$} \scalebox{0.75}{0.01}}$ & \textbf{\scalebox{0.95}{3.06}}$^{\scalebox{0.5}{$\pm$} \scalebox{0.75}{1.33}}$ & \textbf{\scalebox{0.95}{99.34}}$^{\scalebox{0.5}{$\pm$} \scalebox{0.75}{0.23}}$ & \textbf{\scalebox{0.95}{4.68}}$^{\scalebox{0.5}{$\pm$} \scalebox{0.75}{1.00}}$ & \textbf{\scalebox{0.95}{99.08}}$^{\scalebox{0.5}{$\pm$} \scalebox{0.75}{0.26}}$ & \textbf{\scalebox{0.95}{2.68}}$^{\scalebox{0.5}{$\pm$} \scalebox{0.75}{0.51}}$ & \textbf{\scalebox{0.95}{99.31}}$^{\scalebox{0.5}{$\pm$} \scalebox{0.75}{0.43}}$ & \textbf{\scalebox{0.95}{2.63}}$^{\scalebox{0.5}{$\pm$} \scalebox{0.75}{0.58}}$ & \textbf{\scalebox{0.95}{99.42}}$^{\scalebox{0.5}{$\pm$} \scalebox{0.75}{0.22}}$ \\

\rowcolor{gray!20} & \scalebox{0.9}{FA}$_{\scalebox{0.6}{GL}}$\scalebox{0.9}{(Ours)} & \scalebox{0.95}{\underline{0.13}}$^{\scalebox{0.5}{$\pm$} \scalebox{0.75}{0.02}}$ & \scalebox{0.95}{\underline{99.94}}$^{\scalebox{0.5}{$\pm$} \scalebox{0.75}{0.02}}$ & \scalebox{0.95}{\underline{4.98}}$^{\scalebox{0.5}{$\pm$} \scalebox{0.75}{1.02}}$ & \scalebox{0.95}{\underline{99.01}}$^{\scalebox{0.5}{$\pm$} \scalebox{0.75}{0.22}}$ & \scalebox{0.95}{\underline{7.46}}$^{\scalebox{0.5}{$\pm$} \scalebox{0.75}{0.86}}$ & \scalebox{0.95}{\underline{98.44}}$^{\scalebox{0.5}{$\pm$} \scalebox{0.75}{0.27}}$ & \scalebox{0.95}{\underline{5.67}}$^{\scalebox{0.5}{$\pm$} \scalebox{0.75}{0.83}}$ & \scalebox{0.95}{\underline{98.75}}$^{\scalebox{0.5}{$\pm$} \scalebox{0.75}{0.19}}$ & \scalebox{0.95}{\underline{4.56}}$^{\scalebox{0.5}{$\pm$} \scalebox{0.75}{0.47}}$ & \scalebox{0.95}{\underline{99.03}}$^{\scalebox{0.5}{$\pm$} \scalebox{0.75}{0.13}}$ \\

\midrule

 & CoOp$_{\scalebox{0.7}{MCM}}$ & \scalebox{0.95}{99.14}$^{\scalebox{0.5}{$\pm$} \scalebox{0.75}{0.73}}$ & \scalebox{0.95}{31.95}$^{\scalebox{0.5}{$\pm$} \scalebox{0.75}{9.17}}$ & \scalebox{0.95}{98.29}$^{\scalebox{0.5}{$\pm$} \scalebox{0.75}{0.49}}$ & \scalebox{0.95}{44.36}$^{\scalebox{0.5}{$\pm$} \scalebox{0.75}{4.19}}$ & \scalebox{0.95}{97.28}$^{\scalebox{0.5}{$\pm$} \scalebox{0.75}{0.85}}$ & \scalebox{0.95}{47.51}$^{\scalebox{0.5}{$\pm$} \scalebox{0.75}{4.63}}$ & \scalebox{0.95}{86.69}$^{\scalebox{0.5}{$\pm$} \scalebox{0.75}{4.37}}$ & \scalebox{0.95}{69.17}$^{\scalebox{0.5}{$\pm$} \scalebox{0.75}{3.15}}$ & \scalebox{0.95}{95.35}$^{\scalebox{0.5}{$\pm$} \scalebox{0.75}{1.03}}$ & \scalebox{0.95}{48.25}$^{\scalebox{0.5}{$\pm$} \scalebox{0.75}{3.09}}$  \\
 
& CoOp$_{\scalebox{0.7}{GL}}$ & \scalebox{0.95}{92.45}$^{\scalebox{0.5}{$\pm$} \scalebox{0.75}{4.26}}$ & \scalebox{0.95}{57.59}$^{\scalebox{0.5}{$\pm$} \scalebox{0.75}{10.4}}$ & \scalebox{0.95}{90.97}$^{\scalebox{0.5}{$\pm$} \scalebox{0.75}{3.49}}$ & \scalebox{0.95}{61.89}$^{\scalebox{0.5}{$\pm$} \scalebox{0.75}{4.83}}$ & \scalebox{0.95}{88.69}$^{\scalebox{0.5}{$\pm$} \scalebox{0.75}{3.85}}$ & \scalebox{0.95}{64.00}$^{\scalebox{0.5}{$\pm$} \scalebox{0.75}{4.92}}$ & \scalebox{0.95}{62.37}$^{\scalebox{0.5}{$\pm$} \scalebox{0.75}{1.93}}$ & \scalebox{0.95}{83.58}$^{\scalebox{0.5}{$\pm$} \scalebox{0.75}{2.31}}$ & \scalebox{0.95}{83.62}$^{\scalebox{0.5}{$\pm$} \scalebox{0.75}{1.18}}$ & \scalebox{0.95}{66.77}$^{\scalebox{0.5}{$\pm$} \scalebox{0.75}{2.83}}$ \\[0.1em]

& \scalebox{0.92}{LoCoOp}$_{\scalebox{0.55}{MCM}}$ & \scalebox{0.95}{96.51}$^{\scalebox{0.5}{$\pm$} \scalebox{0.75}{3.33}}$ & \scalebox{0.95}{53.49}$^{\scalebox{0.5}{$\pm$} \scalebox{0.75}{8.06}}$ & \scalebox{0.95}{93.52}$^{\scalebox{0.5}{$\pm$} \scalebox{0.75}{5.68}}$ & \scalebox{0.95}{55.59}$^{\scalebox{0.5}{$\pm$} \scalebox{0.75}{9.19}}$ & \scalebox{0.95}{92.88}$^{\scalebox{0.5}{$\pm$} \scalebox{0.75}{4.05}}$ & \scalebox{0.95}{57.89}$^{\scalebox{0.5}{$\pm$} \scalebox{0.75}{5.19}}$ & \scalebox{0.95}{76.45}$^{\scalebox{0.5}{$\pm$} \scalebox{0.75}{7.48}}$ & \scalebox{0.95}{78.59}$^{\scalebox{0.5}{$\pm$} \scalebox{0.75}{2.66}}$ & \scalebox{0.95}{89.84}$^{\scalebox{0.5}{$\pm$} \scalebox{0.75}{4.29}}$ & \scalebox{0.95}{61.39}$^{\scalebox{0.5}{$\pm$} \scalebox{0.75}{3.90}}$ \\

EuroSAT & \scalebox{0.92}{LoCoOp}$_{\scalebox{0.55}{GL}}$ & \scalebox{0.95}{86.13}$^{\scalebox{0.5}{$\pm$} \scalebox{0.75}{12.2}}$ & \scalebox{0.95}{72.68}$^{\scalebox{0.5}{$\pm$} \scalebox{0.75}{7.97}}$ & \scalebox{0.95}{87.81}$^{\scalebox{0.5}{$\pm$} \scalebox{0.75}{6.37}}$ & \scalebox{0.95}{68.58}$^{\scalebox{0.5}{$\pm$} \scalebox{0.75}{6.40}}$ & \scalebox{0.95}{85.93}$^{\scalebox{0.5}{$\pm$} \scalebox{0.75}{4.15}}$ & \scalebox{0.95}{69.90}$^{\scalebox{0.5}{$\pm$} \scalebox{0.75}{3.40}}$ & \scalebox{0.95}{54.89}$^{\scalebox{0.5}{$\pm$} \scalebox{0.75}{4.67}}$ & \scalebox{0.95}{87.00}$^{\scalebox{0.5}{$\pm$} \scalebox{0.75}{1.01}}$ & \scalebox{0.95}{78.69}$^{\scalebox{0.5}{$\pm$} \scalebox{0.75}{4.65}}$ & \scalebox{0.95}{74.54}$^{\scalebox{0.5}{$\pm$} \scalebox{0.75}{3.59}}$ \\[0.1em]

& IDLike & \scalebox{0.95}{98.02}$^{\scalebox{0.5}{$\pm$} \scalebox{0.75}{2.13}}$ & \scalebox{0.95}{73.32}$^{\scalebox{0.5}{$\pm$} \scalebox{0.75}{5.83}}$ & \scalebox{0.95}{95.74}$^{\scalebox{0.5}{$\pm$} \scalebox{0.75}{2.84}}$ & \scalebox{0.95}{65.19}$^{\scalebox{0.5}{$\pm$} \scalebox{0.75}{6.86}}$ & \scalebox{0.95}{94.79}$^{\scalebox{0.5}{$\pm$} \scalebox{0.75}{3.67}}$ & \scalebox{0.95}{69.44}$^{\scalebox{0.5}{$\pm$} \scalebox{0.75}{5.06}}$ & \scalebox{0.95}{63.34}$^{\scalebox{0.5}{$\pm$} \scalebox{0.75}{9.10}}$ & \scalebox{0.95}{86.68}$^{\scalebox{0.5}{$\pm$} \scalebox{0.75}{2.98}}$ & \scalebox{0.95}{87.97}$^{\scalebox{0.5}{$\pm$} \scalebox{0.75}{4.21}}$ & \scalebox{0.95}{73.66}$^{\scalebox{0.5}{$\pm$} \scalebox{0.75}{5.04}}$ \\[0.1em]

& SCT$_{\scalebox{0.7}{MCM}}$ & 
\scalebox{0.95}{99.79}$^{\scalebox{0.5}{$\pm$} \scalebox{0.75}{0.22}}$ & \scalebox{0.95}{32.90}$^{\scalebox{0.5}{$\pm$} \scalebox{0.75}{6.73}}$ & \scalebox{0.95}{98.36}$^{\scalebox{0.5}{$\pm$} \scalebox{0.75}{0.84}}$ & \scalebox{0.95}{44.30}$^{\scalebox{0.5}{$\pm$} \scalebox{0.75}{3.41}}$ & \scalebox{0.95}{97.77}$^{\scalebox{0.5}{$\pm$} \scalebox{0.75}{0.86}}$ & \scalebox{0.95}{48.08}$^{\scalebox{0.5}{$\pm$} \scalebox{0.75}{1.64}}$ & \scalebox{0.95}{83.98}$^{\scalebox{0.5}{$\pm$} \scalebox{0.75}{6.92}}$ & \scalebox{0.95}{73.25}$^{\scalebox{0.5}{$\pm$} \scalebox{0.75}{3.58}}$ & \scalebox{0.95}{94.97}$^{\scalebox{0.5}{$\pm$} \scalebox{0.75}{2.19}}$ & \scalebox{0.95}{49.63}$^{\scalebox{0.5}{$\pm$} \scalebox{0.75}{2.39}}$ \\

& SCT$_{\scalebox{0.7}{GL}}$ & 
\scalebox{0.95}{96.57}$^{\scalebox{0.5}{$\pm$} \scalebox{0.75}{5.41}}$ & \scalebox{0.95}{58.21}$^{\scalebox{0.5}{$\pm$} \scalebox{0.75}{5.31}}$ & \scalebox{0.95}{92.41}$^{\scalebox{0.5}{$\pm$} \scalebox{0.75}{4.55}}$ & \scalebox{0.95}{61.54}$^{\scalebox{0.5}{$\pm$} \scalebox{0.75}{3.47}}$ & \scalebox{0.95}{90.49}$^{\scalebox{0.5}{$\pm$} \scalebox{0.75}{5.02}}$ & \scalebox{0.95}{63.33}$^{\scalebox{0.5}{$\pm$} \scalebox{0.75}{2.71}}$ & \scalebox{0.95}{55.29}$^{\scalebox{0.5}{$\pm$} \scalebox{0.75}{14.7}}$ & \scalebox{0.95}{85.48}$^{\scalebox{0.5}{$\pm$} \scalebox{0.75}{3.68}}$ & \scalebox{0.95}{83.69}$^{\scalebox{0.5}{$\pm$} \scalebox{0.75}{7.02}}$ & \scalebox{0.95}{67.14}$^{\scalebox{0.5}{$\pm$} \scalebox{0.75}{2.20}}$ \\[0.1em]

\rowcolor{gray!20} & \scalebox{0.9}{FA}$_{\scalebox{0.55}{MCM}}$\scalebox{0.9}{(Ours)} & \scalebox{0.95}{\underline{85.24}}$^{\scalebox{0.5}{$\pm$} \scalebox{0.75}{6.88}}$ & \scalebox{0.95}{\underline{78.62}}$^{\scalebox{0.5}{$\pm$} \scalebox{0.75}{3.18}}$ & \textbf{\scalebox{0.95}{36.24}}$^{\scalebox{0.5}{$\pm$} \scalebox{0.75}{10.8}}$ & \textbf{\scalebox{0.95}{92.13}}$^{\scalebox{0.5}{$\pm$} \scalebox{0.75}{2.44}}$ & \textbf{\scalebox{0.95}{28.17}}$^{\scalebox{0.5}{$\pm$} \scalebox{0.75}{10.2}}$ & \textbf{\scalebox{0.95}{94.05}}$^{\scalebox{0.5}{$\pm$} \scalebox{0.75}{1.68}}$ & \scalebox{0.95}{\underline{10.36}}$^{\scalebox{0.5}{$\pm$} \scalebox{0.75}{3.97}}$ & \scalebox{0.95}{\underline{97.71}}$^{\scalebox{0.5}{$\pm$} \scalebox{0.75}{0.69}}$ & \textbf{\scalebox{0.95}{39.99}}$^{\scalebox{0.5}{$\pm$} \scalebox{0.75}{7.79}}$ & \textbf{\scalebox{0.95}{90.63}}$^{\scalebox{0.5}{$\pm$} \scalebox{0.75}{1.91}}$ \\

\rowcolor{gray!20} & \scalebox{0.9}{FA}$_{\scalebox{0.6}{GL}}$\scalebox{0.9}{(Ours)} & \textbf{\scalebox{0.95}{77.82}}$^{\scalebox{0.5}{$\pm$} \scalebox{0.75}{7.28}}$ & \textbf{\scalebox{0.95}{86.88}}$^{\scalebox{0.5}{$\pm$} \scalebox{0.75}{2.15}}$ & \scalebox{0.95}{\underline{56.10}}$^{\scalebox{0.5}{$\pm$} \scalebox{0.75}{8.03}}$ & \scalebox{0.95}{\underline{88.43}}$^{\scalebox{0.5}{$\pm$} \scalebox{0.75}{2.26}}$ & \scalebox{0.95}{\underline{49.87}}$^{\scalebox{0.5}{$\pm$} \scalebox{0.75}{7.69}}$ & \scalebox{0.95}{\underline{89.07}}$^{\scalebox{0.5}{$\pm$} \scalebox{0.75}{1.49}}$ & \textbf{\scalebox{0.95}{8.99}}$^{\scalebox{0.5}{$\pm$} \scalebox{0.75}{2.51}}$ & \textbf{\scalebox{0.95}{97.74}}$^{\scalebox{0.5}{$\pm$} \scalebox{0.75}{0.36}}$ & \scalebox{0.95}{\underline{48.19}}$^{\scalebox{0.5}{$\pm$} \scalebox{0.75}{6.23}}$ & \scalebox{0.95}{\underline{90.53}}$^{\scalebox{0.5}{$\pm$} \scalebox{0.75}{1.42}}$ \\

\bottomrule
\end{tabularx}

\vspace{-0.5em}
\caption{Representative OOD detection results with various ID datasets in the 16-shot setting. We use the same notation as \cref{tab1}.}
\label{tab-otherID}

\end{table*}

\noindent
\textbf{Setup.}\; Following previous studies~\cite{locoop}, we use the ViT-B/16~\cite{vit} as the backbone model. For our model, when using ImageNet-1k as the ID dataset, the epochs are respectively set to 30 and 50 for the 1-shot and 16-shot settings, while for other ID datasets, we set the epoch to 200 following CoOp~\cite{coop}. The value of forced coefficient $K$ is uniformly set to 3, which will be further discussed in the sensitivity study. Other hyperparameters are as follows: learning rate\,=\,2e-3, batch size\,=\,160, SGD (momentum\,=\,0.9, weight decay\,=\,5e-4) as the optimizer with a cosine scheduler, $\tau=1$ in \cref{eq:force_prompt} and \cref{eq:force_prompt_K}. All experiments on our model can be conducted on a single Nvidia A30 GPU. The average experiment results (including our reproductions) over four runs are reported for comparison.

% \noindent
% \textbf{Comparison Methods}\; To validate the effectiveness of our FA, we mainly compare it with CLIP-based OOD detection methods in four directions. For previous post-hoc methods, including MSP~\cite{msp}, ODIN~\cite{ODIN}, Energy~\cite{Energy}, ReAct~\cite{react}, and MaxLogit~\cite{MaxLogit}, we adapt these methods with the CLIP image encoder as the CLIP-based post-hoc methods. For zero-shot methods that only use ID labels, we select MCM~\cite{mcm} and GL-MCM~\cite{glmcm} as baselines. For prompt tuning based methods, we adopt CoOp~\cite{coop}, LoCoOp~\cite{locoop}, IDLike~\cite{idlike}, and SCT~\cite{sct} as baselines. And we adopt NegLabel~\cite{neglabel2}, CSP~\cite{CSP} and CLIPN~\cite{clipn} as the methods that use real outliers (\eg OOD images or labels).

\noindent
\textbf{Comparison Methods.}\; To validate the effectiveness of our FA fairly, we mainly compare it with CLIP-based OOD detection methods that do not use real outliers (\eg OOD labels). 
% in three directions. 
For previous post-hoc methods, including MSP~\cite{msp}, ODIN~\cite{ODIN}, Energy~\cite{Energy}, ReAct~\cite{react}, and MaxLogit~\cite{MaxLogit}, we adapt these methods with the CLIP image encoder as the CLIP-based post-hoc methods. For zero-shot methods, we select MCM~\cite{mcm}, GL-MCM~\cite{glmcm} and CLIPN~\cite{clipn} as baselines. For prompt learning methods, we adopt CoOp~\cite{coop}, LoCoOp~\cite{locoop}, IDLike~\cite{idlike}, and SCT~\cite{sct} as baselines.

% To validate the effectiveness of our FA, we mainly compare it with CLIP-based OOD detection methods in two directions, such as post-hoc methods and prompt tuning methods. For post-hoc methods, we select MCM~\cite{mcm}, GL-MCM~\cite{glmcm}, and CLIPN~\cite{clipn} as zero-shot baselines. For prompt tuning methods, we adopt CoOp~\cite{coop}, LoCoOp~\cite{locoop}, IDLike~\cite{idlike}, and SCT~\cite{sct} as baselines, and the score function follows the original paper.

\noindent
\textbf{Metrics.}\; For evaluation, we adopt the following metrics: (1) The False Positive Rate at 95\% True Positive Rate for in-distribution samples (FPR95); (2) The Area Under the Receiver Operating Characteristic Curve (AUROC); (3) In-distribution data classification Top-1 accuracy (ID ACC).

%------------------------------------------------------------------------
\subsection{Main results}
\label{Comparison}

\noindent
\textbf{Conventional OOD Detection.}\; \Cref{tab1} summarizes our comparison results on the ImageNet-1k benchmarks, which show that our FA achieves state-of-the-art OOD detection performance among other CLIP-based methods under different few-shot settings (More details under the 4-shot setting are provided in the Appendix \ref{appendix-imagenet1k}). Specifically, in the 16-shot setting, FA with GL-MCM score ($\mathrm{FA_{\scriptscriptstyle GL}}$) outperforms the best baseline $\mathrm{SCT_{\scriptscriptstyle GL}}$ (average FPR95 of 25.68$\%$ \vs. 27.27$\%$, and average AUROC of 93.82$\%$ \vs. 93.31$\%$). 
Additionally, in the 1-shot setting, $\mathrm{FA_{\scriptscriptstyle GL}}$ surpasses $\mathrm{SCT_{\scriptscriptstyle GL}}$ by a large margin, showing improvements of at least 1.25$\%$ and 3.81$\%$ on average AUROC and FPR95. More importantly, even the average score of $\mathrm{FA_{\scriptscriptstyle MCM}}$ has exceeded $\mathrm{SCT_{\scriptscriptstyle GL}}$ in the 1-shot setting. 
Moreover, our FA effectively improves the OOD performance without sacrificing the ID classification capability of the model, as shown in \cref{tab-imagenet1k-acc}.

In particular, we can also observe that our method performs not well on the SUN dataset, as shown in \cref{tab1}.
This is related to the presence of a significant number of samples belonging to ID category in these datasets, as demonstrated by recent studies~\cite{idlike}.
In other words, the SUN dataset in the popular OOD benchmark requires more thorough cleaning to be effectively used as an OOD dataset.

% \textcolor{red}{Additionally, as shown in Table \ref{tab_ours+}, our FA performs better when real outliers, such as OOD labels, are available.
% Specially, due to the lack of OOD labels specific to the near-OOD dataset (\ie NINCO~\cite{NINCO}), these methods underperform compared to our FA (AUROC: \textbf{NegLabel/76.13$\%$, CSP/76.83$\%$, FA/79.01$\%$}).}

% \textcolor{red}{Furthermore, we also provide a visualization on real data in the Appendix \ref{appendix-Visualization} to further illustrate our motivation.}

\begin{figure}
    \centering
    \includegraphics[width=0.91\linewidth,  ]{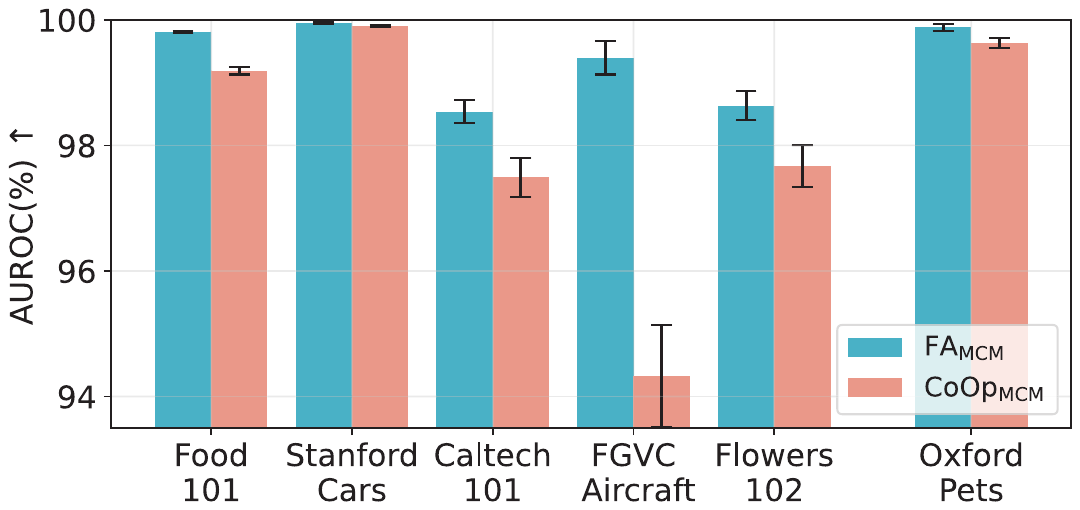}
    \vspace{-0.8em}
    \caption{A portion of OOD detection average performance using other ID datasets under the 16-shot setting. We report the mean and standard deviation of the AUROC for our FA (blue) and CoOp (orange) using the MCM score. More details can be found in the Appendix \ref{appendix-otherID}. }
    \label{fig-otherID}
    \vspace{-1.2em}
\end{figure}

\noindent
\textbf{Challenging OOD Detection.}\; Recent studies~\cite{NINCO,idlike} have shown that some of the used OOD datasets contain more or less images of objects that belong to ID classes in ImageNet-1k~\cite{imagenet1k}. Hence, probing the true performance of OOD detectors for ImageNet-1k demands OOD datasets that are both challenging and genuinely OOD. As shown in \cref{tab-challenge}, to fully substantiate the effectiveness of our FA, we leverage 3 additional cleaner and more challenging OOD datasets, including OpenImage-O~\cite{OpenImage-O}, NINCO~\cite{NINCO}, and ImageNet-O~\cite{ImageNet-O}, inspired by \cite{NINCO}. Notably, both our $\mathrm{FA_{\scriptscriptstyle MCM}}$ and $\mathrm{FA_{\scriptscriptstyle GL}}$ consistently outperform current SOTA methods in terms of average FPR95 and AUROC under different few-shot settings (More details under the 1-shot and 4-shot settings are provided in the Appendix \ref{appendix-imagenet1k}).

\noindent
\textbf{Various ID Datasets.}\; Considering factors like dataset variety, image resolution, and the number of classes~\cite{mcm}, we also leverage 8 additional widely used datasets for few-shot setting as ID datasets for comparison~\cite{coop,mcm,lsn}. Overall, despite our model is simple, it achieves superior average OOD performance on these ID datasets compared to other methods. We selected representative results using  UCF101~\cite{UCF101} and EuroSAT~\cite{EuroSAT} as ID datasets in the 16-shot setting, as shown in \cref{tab-otherID}. Since using the other datasets (\ie Food101~\cite{Food101}, StandfordCars~\cite{StanfordCars}, Caltech101~\cite{Caltech101}, FGVCAircraft~\cite{FGVCAircraft}, Flowers102~\cite{Flowers102}, and OxfordPets~\cite{OxfordPets}) as ID datasets for OOD detection presents relatively lower difficulty as shown in \cref{fig-otherID}, we leave more experimental details in Appendix \ref{appendix-otherID}.

\begin{figure}
    \centering
    \includegraphics[width=0.91\linewidth,  ]{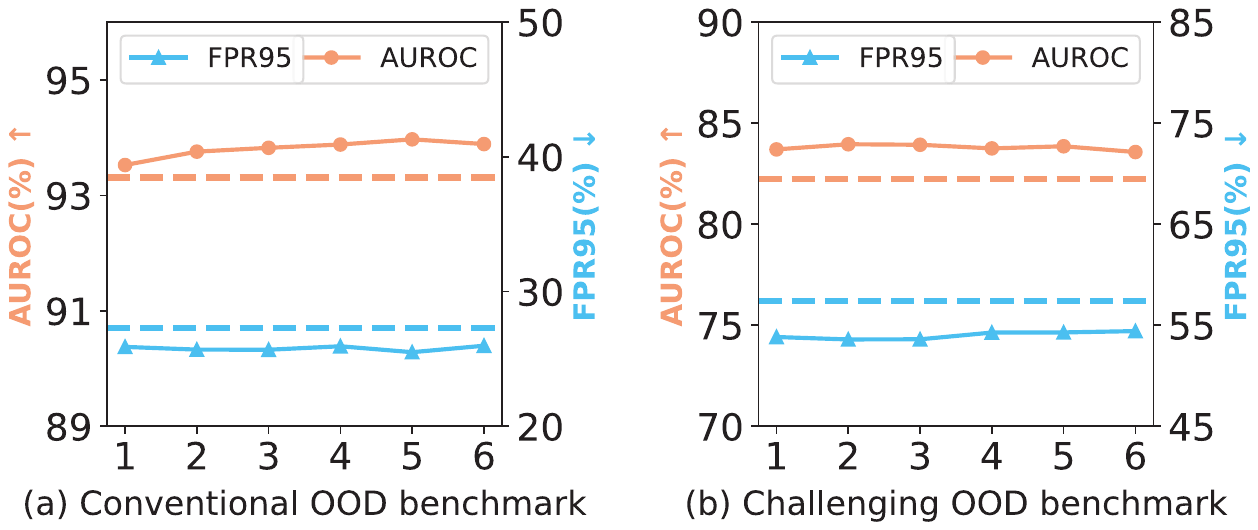}
    \vspace{-0.5em}
    \caption{Sensitivity study of hyperparameter $K$. We report the average FPR95 (Blue) and AUROC (Orange) of our $\mathrm{FA_{\scriptscriptstyle GL}}$ on different OOD benchmarks using ImageNet-1k as the ID dataset under 16-shot setting. Dashed lines represent the performance of the best baseline (\ie \,$\mathrm{SCT_{\scriptscriptstyle GL}}$).  }
    \label{sensitivity}
    \vspace{-1.5em}
\end{figure}

In particular, for the ID dataset EuroSAT, $\mathrm{FA_{\scriptscriptstyle MCM}}$ surpasses the best-competing method $\mathrm{LoCoOp_{\scriptscriptstyle GL}}$ by 16.09$\%$ in average AUROC and 38.7$\%$ in average FPR95, underscoring its effectiveness and versatility even when the ID dataset is of low resolution (64x64 pixels).

\subsection{Ablation study}
\label{ablation}

\noindent
\textbf{Influence of the $\mathcal{L}_{FCE-K}$.}\; As shown in \cref{tab-fce}, we trained the model FA$_{CE}$ and FA$_{FCE-K}$ using the standard cross-entropy loss $\mathcal{L}_{CE}$ and our $\mathcal{L}_{FCE-K}$, respectively. 
Our FA$_{FCE-K}$ significantly outperforms FA$_{CE}$ under the same inference setting, which demonstrate the effectiveness of our $\mathcal{L}_{FCE-K}$.

\noindent
\textbf{Initialization of the forced and original prompt.}\; As shown in \cref{tab-init}, we perform experiments comparing different initialization scenarios for both prompts. Manual initialization uses the embeddings of ``a photo of a [\textit{class-c}]" to initialize the prompt's embeddings, while the random initialization setting is based on CoOp~\cite{coop}. The results show that employing consistent manual initialization for both prompts enables the forced prompt to effectively learn more specific descriptions of the ID classes.

\noindent
\textbf{Influence of the shared learnable vector.}\; We also explore the influence of the shared learnable vector and the independent learnable vector for the forced prompts. As shown in \cref{tab-csc}, we can observe that using the shared learnable vector performs well in the few-shot setting since the independent learnable vector for each class has more parameters and requires more training data~\cite{coop}.

\subsection{Sensitivity study}
\label{Sensitivity}

As shown in \cref{sensitivity}, we explore the influence of the hyperparameter $K$. Overall, with $K$ ranging from 1 to 6, the results indicate that our proposed framework is insensitive to the choice of $K$.
Concretely,  our method is stable and outperforms the best baseline $\mathrm{SCT_{\scriptscriptstyle GL}}$ in average AUROC and FPR95 on both conventional and challenging OOD detection benchmarks. Specially, as $K$ increases, the average AUROC shows an upward trend and then tends to be stable (see \cref{sensitivity} (a)), showing a bottleneck in capturing more comprehensive semantic representations of the ID classes.

\begin{table}[ht]
\centering

\footnotesize

\begin{tabularx}{8cm}{@{}>{\centering\arraybackslash}p{0.8cm} *{4}{>{\centering\arraybackslash}X}}
\toprule

\multirow{2}{*}{\makecell[c]{Method} } &
\multicolumn{2}{c}{MCM} & \multicolumn{2}{c}{GL-MCM} \\[0.3em]

& FPR95$\downarrow$ & AUROC$\uparrow$ & FPR95$\downarrow$ & AUROC$\uparrow$   \\
\midrule

FA$_{CE}$ & 41.43$^{\scalebox{0.6}{$\pm$} \scalebox{0.75}{{3.11}}}$ & 91.01$^{\scalebox{0.6}{$\pm$} \scalebox{0.75}{{0.67}}}$ & 34.29$^{\scalebox{0.6}{$\pm$} \scalebox{0.75}{{1.94}}}$ & 90.99$^{\scalebox{0.6}{$\pm$} \scalebox{0.75}{{0.56}}}$ \\

FA$_{\scalebox{0.65}{$\mathit{FCE-K}$}}$ & 29.07$^{\scalebox{0.6}{$\pm$} {\scalebox{0.75}{1.11}}}$ & 93.77$^{\scalebox{0.6}{$\pm$} {\scalebox{0.75}{0.19}}}$ & \textbf{25.68}$^{\scalebox{0.6}{$\pm$} {\scalebox{0.75}{0.58}}}$ & \textbf{93.82}$^{\scalebox{0.6}{$\pm$} {\scalebox{0.75}{0.11}}}$ \\

\bottomrule
\end{tabularx}

\vspace{-0.5em}
\caption{Influence of the $\mathcal{L}_{FCE-K}$. MCM and GL-MCM refer to the score functions used during model inference.
}
\label{tab-fce}

\vspace{-1.5em}
\end{table}

\begin{table}[ht]
\centering

\footnotesize

\begin{tabularx}{8cm}{@{}>{\centering\arraybackslash}p{0.8cm}>  {\centering\arraybackslash}p{0.8cm} *{4}{>{\centering\arraybackslash}X}}
\toprule

\multirow{2}{*}{\makecell[c]{Forced-\\M-Init} } &
\multirow{2}{*}{\makecell[c]{\scalebox{0.95}{Original}-\\M-Init} } &
\multicolumn{2}{c}{FA$_{\text{MCM}}$} & \multicolumn{2}{c}{FA$_{\text{GL}}$} \\[0.3em]

&& FPR95$\downarrow$ & AUROC$\uparrow$ & FPR95$\downarrow$ & AUROC$\uparrow$   \\
\midrule

\textbf{-} & \textbf{-} & \scalebox{0.95}{40.71}$^{\scalebox{0.6}{$\pm$} {\scalebox{0.75}{4.00}}}$ & \scalebox{0.95}{91.28}$^{\scalebox{0.6}{$\pm$} {\scalebox{0.75}{0.82}}}$ & \scalebox{0.95}{34.88}$^{\scalebox{0.6}{$\pm$} {\scalebox{0.75}{2.92}}}$ & \scalebox{0.95}{92.12}$^{\scalebox{0.6}{$\pm$} {\scalebox{0.75}{0.76}}}$ \\[0.2em]
\checkmark & \textbf{-} & \scalebox{0.95}{40.81}$^{\scalebox{0.6}{$\pm$} {\scalebox{0.75}{2.25}}}$ & \scalebox{0.95}{91.25}$^{\scalebox{0.6}{$\pm$} {\scalebox{0.75}{0.64}}}$ & \scalebox{0.95}{36.03}$^{\scalebox{0.6}{$\pm$} {\scalebox{0.75}{2.84}}}$ & \scalebox{0.95}{91.62}$^{\scalebox{0.6}{$\pm$} {\scalebox{0.75}{1.01}}}$ \\[0.2em]
\textbf{-} & \checkmark & \scalebox{0.95}{31.35}$^{\scalebox{0.6}{$\pm$} {\scalebox{0.75}{1.34}}}$ & \scalebox{0.95}{93.35}$^{\scalebox{0.6}{$\pm$} {\scalebox{0.75}{0.34}}}$ & \scalebox{0.95}{26.39}$^{\scalebox{0.6}{$\pm$} {\scalebox{0.75}{1.02}}}$ & \scalebox{0.95}{93.64}$^{\scalebox{0.6}{$\pm$} {\scalebox{0.75}{0.19}}}$ \\[0.2em]
\checkmark & \checkmark & \scalebox{0.95}{29.07}$^{\scalebox{0.6}{$\pm$} {\scalebox{0.75}{1.11}}}$ & \scalebox{0.95}{93.77}$^{\scalebox{0.6}{$\pm$} {\scalebox{0.75}{0.19}}}$ & \textbf{\scalebox{0.95}{25.68}}$^{\scalebox{0.6}{$\pm$} {\scalebox{0.75}{0.58}}}$ & \textbf{\scalebox{0.95}{93.82}}$^{\scalebox{0.6}{$\pm$} {\scalebox{0.75}{0.11}}}$ \\

\bottomrule
\end{tabularx}

\vspace{-0.5em}
\caption{Ablation study of different initialization scenarios for the forced and original prompts. ``M-Init" represents manual initialization. \checkmark \;represents using manual initialization; \textbf{-} represents using random initialization.}
\label{tab-init}
\vspace{-1.5em}
\end{table}

\begin{table}[ht]
\centering

\footnotesize

\begin{tabularx}{8cm}{@{}>{\centering\arraybackslash}p{0.8cm}>  {\centering\arraybackslash}p{0.8cm} *{4}{>{\centering\arraybackslash}X}}
\toprule

\multirow{2}{*}{\makecell[c]{Shared-\\Vector} } &
\multirow{2}{*}{\makecell[c]{M-Init} } &
\multicolumn{2}{c}{FA$_{\text{MCM}}$} & \multicolumn{2}{c}{FA$_{\text{GL}}$} \\[0.3em]

&& FPR95$\downarrow$ & AUROC$\uparrow$ & FPR95$\downarrow$ & AUROC$\uparrow$   \\
\midrule

\textbf{-} & \textbf{-} & 44.02$^{\scalebox{0.6}{$\pm$} \scalebox{0.75}{{2.24}}}$ & 90.17$^{\scalebox{0.6}{$\pm$} \scalebox{0.75}{{0.42}}}$ & 46.29$^{\scalebox{0.6}{$\pm$} \scalebox{0.75}{{1.33}}}$ & 88.24$^{\scalebox{0.6}{$\pm$} \scalebox{0.75}{{0.41}}}$ \\[0.2em]
\checkmark & \textbf{-} & 40.71$^{\scalebox{0.6}{$\pm$} \scalebox{0.75}{{4.00}}}$ & 91.28$^{\scalebox{0.6}{$\pm$} \scalebox{0.75}{{0.82}}}$ & 34.88$^{\scalebox{0.6}{$\pm$} \scalebox{0.75}{{2.92}}}$ & 92.12$^{\scalebox{0.6}{$\pm$} \scalebox{0.75}{{0.76}}}$ \\[0.2em]
\textbf{-}& \checkmark & 34.69$^{\scalebox{0.6}{$\pm$} \scalebox{0.75}{{0.62}}}$ & 92.57$^{\scalebox{0.6}{$\pm$} \scalebox{0.75}{{0.12}}}$ & 29.27$^{\scalebox{0.6}{$\pm$} \scalebox{0.75}{{0.55}}}$ & 92.82$^{\scalebox{0.6}{$\pm$} \scalebox{0.75}{{0.80}}}$ \\[0.2em]

\checkmark & \checkmark & 29.07$^{\scalebox{0.6}{$\pm$} {\scalebox{0.75}{1.11}}}$ & 93.77$^{\scalebox{0.6}{$\pm$} {\scalebox{0.75}{0.19}}}$ & \textbf{25.68}$^{\scalebox{0.6}{$\pm$} {\scalebox{0.75}{0.58}}}$ & \textbf{93.82}$^{\scalebox{0.6}{$\pm$} {\scalebox{0.75}{0.11}}}$ \\

\bottomrule
\end{tabularx}

\vspace{-0.5em}
\caption{Ablation study of the shared learnable vector. \,\checkmark \,\textbf{/} \textbf{-} for ``Shared-vector" represents using the shared \textbf{/} independent learnable vector for the forced prompt; \,\checkmark \,\textbf{/} \textbf{-} for ``M-Init" represents using manual \textbf{/} random initialization for the forced prompt.
}
\label{tab-csc}

\vspace{-1.5em}
\end{table}

\section{Conclusion}

% In this study, we propose a novel framework for VLMs-based OOD detection called FA, which utilizes the frozen origin branch as a template, forcing the learnable FA branch to acquire knowledge that enables ID data to distinguish classifiers from both branches, while OOD data may not have such capability. 

% In this study, we propose a novel framework for VLMs-based OOD detection called \textbf{F}orced le\textbf{A}rning (FA). FA can use the frozen origin branch with CLIP's prior knowledge as the template, forcing the learnable FA branch to acquire knowledge that enables ID data to effectively distinguish classifiers with similar semantics from both branches, while OOD data may not have such capability. Then, we implement FA via prompt tuning, eliminating the reliance on additional learnable prompts and external datasets. Furthermore, we not only evaluate this simple yet effective approach on conventional benchmarks but also reference various ID datasets as well as challenging OOD datasets, and conduct extensive ablation studies. The results show that our model consistently outperforms current SOTA methods. 

In this study, we propose a novel CLIP-based framework for OOD detection based on \textbf{F}orced prompt le\textbf{A}rning (FA), which focus on fully exploiting the ID knowledge to effectively improve OOD detection without exploring complex OOD-related knowledge. 
We introduce a learnable forced prompt in addition to the frozen original prompt, both of which are using the same manual initialization. 
The forced prompt treats the original prompt as a reference, forcing itself to learn more diversified semantic descriptions of the ID classes rather than being limited to the textual semantics of class labels.
We also introduce a forced coefficient to facilitate the forced prompt in learning more nuanced descriptions of the ID classes.
% Furthermore, we not only evaluate this simple yet effective approach on conventional benchmarks but also reference various ID datasets as well as challenging OOD datasets, and conduct extensive ablation studies. The results show that our model consistently outperforms current SOTA methods. 
Comprehensive experimental evaluations demonstrate that our method consistently surpasses current state-of-the-art methods on diverse benchmarks.
% However,.
% Despite our model achieving superior OOD detection performance, one limitation is the minor fluctuation in ID ACC, which may require combining more prior knowledge from the image branch of CLIP to improve, and we also plan to address this issue in future work. 
We expect that our study can bring new insight on VLMs-based OOD detection methods and inspire more future research. 

\section{Acknowledgments}
This work is supported in part by the National Natural
Science Foundation of China (grant No. 62071502), the Major Key Project of
PCL (grant No. PCL2023A09), and Guangdong Excellent Youth Team Program
(grant No. 2023B1515040025).

{
    \small
    \bibliographystyle{ieeenat_fullname}
    \bibliography{main}
}

\newpage

\appendix

\onecolumn

\begin{center}
    \Large\textbf{FA: Forced Prompt Learning of Vision-Language Models for\\
Out-of-Distribution Detection}  
\end{center}

\begin{center}
    \Large{Appendix / Supplementary Material}  
\end{center}

% \bigskip  

\section{More experimental details of OOD detection with ImageNet-1k.}
\label{appendix-imagenet1k}

Referring to Section \ref{Comparison}, we also conduct experiments under other few-shot settings on both conventional and challenging ImageNet-1k OOD benchmarks. As shown in \cref{appendix-tab-conventional} and \cref{appendix-tab-challenging}, our FA achieves superior average performance.

\begin{table*}[ht]
\centering

\footnotesize

\begin{tabularx}{\textwidth}{p{1.3cm}*{10}{>{\centering\arraybackslash}X}}
% \begin{tabularx}{\textwidth}{>{\raggedright\arraybackslash}m{1.30cm}*{11}{>{\centering\arraybackslash}m{1.1cm}}}
\toprule

\multirow{2}{*}{\makecell[c]{Method} } &

\multicolumn{2}{c}{\scalebox{0.95}{iNaturalist}} & \multicolumn{2}{c}{\scalebox{0.95}{SUN}} & \multicolumn{2}{c}{\scalebox{0.95}{Places}} & \multicolumn{2}{c}{\scalebox{0.95}{Textures}} \vrule  & \multicolumn{2}{c} {\scalebox{0.95}{Average}}     \\

 & \scalebox{0.95}{FPR95}$\downarrow$ & \scalebox{0.95}{AUROC}$\uparrow$ & \scalebox{0.95}{FPR95}$\downarrow$ & \scalebox{0.95}{AUROC}$\uparrow$ & \scalebox{0.95}{FPR95}$\downarrow$ & AUROC$\uparrow$ & FPR95$\downarrow$ & \scalebox{0.95}{AUROC}$\uparrow$ & \scalebox{0.95}{FPR95}$\downarrow$ & \scalebox{0.95}{AUROC}$\uparrow$ \\
\midrule

CoOp$_{\scalebox{0.7}{MCM}}$  & 
\scalebox{0.95}{33.73}$^{\scalebox{0.5}{$\pm$} \scalebox{0.75}{5.73}}$ &
\scalebox{0.95}{92.79}$^{\scalebox{0.5}{$\pm$} \scalebox{0.75}{1.28}}$ & 
\scalebox{0.95}{37.15}$^{\scalebox{0.5}{$\pm$} \scalebox{0.75}{1.52}}$ & 
\scalebox{0.95}{92.17}$^{\scalebox{0.5}{$\pm$} \scalebox{0.75}{0.26}}$ & 
\scalebox{0.95}{44.58}$^{\scalebox{0.5}{$\pm$} \scalebox{0.75}{1.43}}$ & 
\scalebox{0.95}{89.57}$^{\scalebox{0.5}{$\pm$} \scalebox{0.75}{0.32}}$ & 
\scalebox{0.95}{47.27}$^{\scalebox{0.5}{$\pm$} \scalebox{0.75}{0.64}}$ & 
\scalebox{0.95}{89.25}$^{\scalebox{0.5}{$\pm$} \scalebox{0.75}{0.57}}$ & 
\scalebox{0.95}{36.17}$^{\scalebox{0.5}{$\pm$} \scalebox{0.75}{1.43}}$ & 
\scalebox{0.95}{92.18}$^{\scalebox{0.5}{$\pm$} \scalebox{0.75}{0.34}}$  \\

CoOp$_{\scalebox{0.7}{GL}}$  & 
\scalebox{0.95}{18.92}$^{\scalebox{0.5}{$\pm$} \scalebox{0.75}{3.46}}$ & 
\scalebox{0.95}{95.36}$^{\scalebox{0.5}{$\pm$} \scalebox{0.75}{0.94}}$ & 
\scalebox{0.95}{29.59}$^{\scalebox{0.5}{$\pm$} \scalebox{0.75}{1.89}}$ & 
\scalebox{0.95}{92.67}$^{\scalebox{0.5}{$\pm$} \scalebox{0.75}{0.76}}$ & 
\scalebox{0.95}{37.55}$^{\scalebox{0.5}{$\pm$} \scalebox{0.75}{1.19}}$ & 
\scalebox{0.95}{90.11}$^{\scalebox{0.5}{$\pm$} \scalebox{0.75}{0.35}}$ & 
\scalebox{0.95}{51.35}$^{\scalebox{0.5}{$\pm$} \scalebox{0.75}{2.19}}$ & 
\scalebox{0.95}{84.99}$^{\scalebox{0.5}{$\pm$} \scalebox{0.75}{1.50}}$ & 
\scalebox{0.95}{34.35}$^{\scalebox{0.5}{$\pm$} \scalebox{0.75}{1.87}}$ & 
\scalebox{0.95}{90.78}$^{\scalebox{0.5}{$\pm$} \scalebox{0.75}{0.78}}$ \\[0.1em]

\scalebox{0.92}{LoCoOp}$_{\scalebox{0.7}{MCM}}$  & 
\scalebox{0.95}{29.60}$^{\scalebox{0.5}{$\pm$} \scalebox{0.75}{5.72}}$ & 
\scalebox{0.95}{93.86}$^{\scalebox{0.5}{$\pm$} \scalebox{0.75}{1.19}}$ & 
\scalebox{0.95}{32.05}$^{\scalebox{0.5}{$\pm$} \scalebox{0.75}{1.88}}$ & 
\scalebox{0.95}{\underline{93.62}}$^{\scalebox{0.5}{$\pm$} \scalebox{0.75}{0.29}}$ & 
\scalebox{0.95}{40.63}$^{\scalebox{0.5}{$\pm$} \scalebox{0.75}{1.54}}$ & 
\scalebox{0.95}{90.59}$^{\scalebox{0.5}{$\pm$} \scalebox{0.75}{0.28}}$ & 
\scalebox{0.95}{42.41}$^{\scalebox{0.5}{$\pm$} \scalebox{0.75}{1.72}}$ & 
\scalebox{0.95}{90.66}$^{\scalebox{0.5}{$\pm$} \scalebox{0.75}{0.30}}$ & 
\scalebox{0.95}{36.17}$^{\scalebox{0.5}{$\pm$} \scalebox{0.75}{1.43}}$ & 
\scalebox{0.95}{92.18}$^{\scalebox{0.5}{$\pm$} \scalebox{0.75}{0.33}}$ \\

\scalebox{0.92}{LoCoOp}$_{\scalebox{0.7}{GL}}$  & 
\scalebox{0.95}{19.30}$^{\scalebox{0.5}{$\pm$} \scalebox{0.75}{3.01}}$ & 
\scalebox{0.95}{95.98}$^{\scalebox{0.5}{$\pm$} \scalebox{0.75}{0.67}}$ & 
\scalebox{0.95}{\textbf{23.32}}$^{\scalebox{0.5}{$\pm$} \scalebox{0.75}{1.87}}$ & 
\scalebox{0.95}{\textbf{95.05}}$^{\scalebox{0.5}{$\pm$} \scalebox{0.75}{0.53}}$ & 
\scalebox{0.95}{33.25}$^{\scalebox{0.5}{$\pm$} \scalebox{0.75}{1.77}}$ & 
\scalebox{0.95}{91.75}$^{\scalebox{0.5}{$\pm$} \scalebox{0.75}{0.35}}$ & 
\scalebox{0.95}{44.10}$^{\scalebox{0.5}{$\pm$} \scalebox{0.75}{2.52}}$ & 
\scalebox{0.95}{89.25}$^{\scalebox{0.5}{$\pm$} \scalebox{0.75}{0.38}}$ & 
\scalebox{0.95}{29.99}$^{\scalebox{0.5}{$\pm$} \scalebox{0.75}{1.49}}$ & 
\scalebox{0.95}{93.01}$^{\scalebox{0.5}{$\pm$} \scalebox{0.75}{0.38}}$ \\[0.1em]

IDLike  & 
\scalebox{0.95}{23.93}$^{\scalebox{0.5}{$\pm$} \scalebox{0.75}{4.26}}$ & 
\scalebox{0.95}{94.92}$^{\scalebox{0.5}{$\pm$} \scalebox{0.75}{1.77}}$ & 
\scalebox{0.95}{52.70}$^{\scalebox{0.5}{$\pm$} \scalebox{0.75}{8.49}}$ & 
\scalebox{0.95}{88.71}$^{\scalebox{0.5}{$\pm$} \scalebox{0.75}{2.67}}$ & 
\scalebox{0.95}{52.49}$^{\scalebox{0.5}{$\pm$} \scalebox{0.75}{9.67}}$ & 
\scalebox{0.95}{88.49}$^{\scalebox{0.5}{$\pm$} \scalebox{0.75}{3.11}}$ & 
\scalebox{0.95}{\underline{28.31}}$^{\scalebox{0.5}{$\pm$} \scalebox{0.75}{3.52}}$ & 
\scalebox{0.95}{\underline{93.93}}$^{\scalebox{0.5}{$\pm$} \scalebox{0.75}{0.64}}$ & 
\scalebox{0.95}{39.35}$^{\scalebox{0.5}{$\pm$} \scalebox{0.75}{2.60}}$ & 
\scalebox{0.95}{91.51}$^{\scalebox{0.5}{$\pm$} \scalebox{0.75}{1.32}}$ \\[0.1em]

SCT$_{\scalebox{0.7}{MCM}}$ & 
\scalebox{0.95}{34.91}$^{\scalebox{0.5}{$\pm$} \scalebox{0.75}{4.20}}$ & 
\scalebox{0.95}{92.90}$^{\scalebox{0.5}{$\pm$} \scalebox{0.75}{0.73}}$ & 
\scalebox{0.95}{31.84}$^{\scalebox{0.5}{$\pm$} \scalebox{0.75}{3.72}}$ & 
\scalebox{0.95}{93.29}$^{\scalebox{0.5}{$\pm$} \scalebox{0.75}{0.79}}$ & 
\scalebox{0.95}{38.89}$^{\scalebox{0.5}{$\pm$} \scalebox{0.75}{2.97}}$ & 
\scalebox{0.95}{90.49}$^{\scalebox{0.5}{$\pm$} \scalebox{0.75}{0.56}}$ & 
\scalebox{0.95}{42.70}$^{\scalebox{0.5}{$\pm$} \scalebox{0.75}{4.13}}$ & 
\scalebox{0.95}{90.36}$^{\scalebox{0.5}{$\pm$} \scalebox{0.75}{0.61}}$ & 
\scalebox{0.95}{37.08}$^{\scalebox{0.5}{$\pm$} \scalebox{0.75}{2.95}}$ & 
\scalebox{0.95}{91.76}$^{\scalebox{0.5}{$\pm$} \scalebox{0.75}{0.48}}$ \\

SCT$_{\scalebox{0.7}{GL}}$ & 
\scalebox{0.95}{\underline{18.46}}$^{\scalebox{0.5}{$\pm$} \scalebox{0.75}{1.71}}$ & 
\scalebox{0.95}{\underline{96.06}}$^{\scalebox{0.5}{$\pm$} \scalebox{0.75}{0.13}}$ & 
\scalebox{0.95}{\underline{27.22}}$^{\scalebox{0.5}{$\pm$} \scalebox{0.75}{5.49}}$ & 
\scalebox{0.95}{93.57}$^{\scalebox{0.5}{$\pm$} \scalebox{0.75}{1.70}}$ & 
\scalebox{0.95}{\underline{32.46}}$^{\scalebox{0.5}{$\pm$} \scalebox{0.75}{1.26}}$ & 
\scalebox{0.95}{91.39}$^{\scalebox{0.5}{$\pm$} \scalebox{0.75}{0.30}}$ & 
\scalebox{0.95}{43.59}$^{\scalebox{0.5}{$\pm$} \scalebox{0.75}{2.84}}$ & 
\scalebox{0.95}{88.61}$^{\scalebox{0.5}{$\pm$} \scalebox{0.75}{0.51}}$ & 
\scalebox{0.95}{\underline{30.43}}$^{\scalebox{0.5}{$\pm$} \scalebox{0.75}{0.85}}$ & 
\scalebox{0.95}{92.41}$^{\scalebox{0.5}{$\pm$} \scalebox{0.75}{0.41}}$ \\[0.1em]

\rowcolor{gray!20} \scalebox{0.95}{FA}$_{\scalebox{0.6}{MCM}}$(Ours) & 
\scalebox{0.95}{26.91}$^{\scalebox{0.5}{$\pm$} \scalebox{0.75}{3.49}}$ & 
\scalebox{0.95}{94.47}$^{\scalebox{0.5}{$\pm$} \scalebox{0.75}{0.64}}$ & 
\scalebox{0.95}{35.98}$^{\scalebox{0.5}{$\pm$} \scalebox{0.75}{2.71}}$ & 
\scalebox{0.95}{92.65}$^{\scalebox{0.5}{$\pm$} \scalebox{0.75}{0.61}}$ & 
\scalebox{0.95}{33.69}$^{\scalebox{0.5}{$\pm$} \scalebox{0.75}{0.51}}$ & 
\textbf{\scalebox{0.95}{92.78}}$^{\scalebox{0.5}{$\pm$} \scalebox{0.75}{0.07}}$ & 
\textbf{\scalebox{0.95}{25.61}}$^{\scalebox{0.5}{$\pm$} \scalebox{0.75}{1.95}}$ & 
\textbf{\scalebox{0.95}{94.59}}$^{\scalebox{0.5}{$\pm$} \scalebox{0.75}{0.19}}$ & 
\scalebox{0.95}{30.55}$^{\scalebox{0.5}{$\pm$} \scalebox{0.75}{0.83}}$ & 
\scalebox{0.95}{\underline{93.63}}$^{\scalebox{0.5}{$\pm$} \scalebox{0.75}{0.24}}$ \\

\rowcolor{gray!20} \scalebox{0.95}{FA}$_{\scalebox{0.6}{GL}}$(Ours)  & 
\textbf{\scalebox{0.95}{14.01}}$^{\scalebox{0.5}{$\pm$} \scalebox{0.75}{1.62}}$ & 
\textbf{\scalebox{0.95}{96.75}}$^{\scalebox{0.5}{$\pm$} \scalebox{0.75}{0.37}}$ & 
\scalebox{0.95}{29.10}$^{\scalebox{0.5}{$\pm$} \scalebox{0.75}{2.32}}$ & 
\scalebox{0.95}{93.26}$^{\scalebox{0.5}{$\pm$} \scalebox{0.75}{0.57}}$ & 
\scalebox{0.95}{\textbf{30.77}}$^{\scalebox{0.5}{$\pm$} \scalebox{0.75}{0.45}}$ & 
\scalebox{0.95}{\underline{92.59}}$^{\scalebox{0.5}{$\pm$} \scalebox{0.75}{0.19}}$ & 
\scalebox{0.95}{30.98}$^{\scalebox{0.5}{$\pm$} \scalebox{0.75}{2.47}}$ & 
\scalebox{0.95}{92.48}$^{\scalebox{0.5}{$\pm$} \scalebox{0.75}{0.64}}$ & 
\textbf{\scalebox{0.95}{26.21}}$^{\scalebox{0.5}{$\pm$} \scalebox{0.75}{1.09}}$ & 
\textbf{\scalebox{0.95}{93.77}}$^{\scalebox{0.5}{$\pm$} \scalebox{0.75}{0.24}}$ \\

\bottomrule
\end{tabularx}

\vspace{-0.5em} % Adjust the spacing here
\caption{Conventional ImageNet-1k OOD benchmark under the 4-shot setting. We use the same notation as \cref{tab1}}
\label{appendix-tab-conventional}

\end{table*}

\begin{table*}[ht]
\centering

\footnotesize

\begin{tabularx}{\textwidth}{l*{9}{>{\centering\arraybackslash}X}}
% \begin{tabularx}{\textwidth}{>{\raggedright\arraybackslash}m{1.30cm}*{11}{>{\centering\arraybackslash}m{1.1cm}}}
\toprule

\multirow{2}{*}{\textbf{\makecell[c]{Method} }} &

\multicolumn{2}{c}{\textbf{\scalebox{0.925}{OpenImage-O}}} & \multicolumn{2}{c}{\textbf{\scalebox{0.925}{NINCO}}} & \multicolumn{2}{c}{\textbf{\scalebox{0.925}{ImageNet-O}}}  \vrule  & \multicolumn{2}{c} {\textbf{\scalebox{0.925}{Average}}} \\

& \scalebox{0.925}{FPR95}$\downarrow$ & \scalebox{0.925}{AUROC}$\uparrow$ & \scalebox{0.925}{FPR95}$\downarrow$ & \scalebox{0.925}{AUROC}$\uparrow$ & \scalebox{0.925}{FPR95}$\downarrow$ & \scalebox{0.925}{AUROC}$\uparrow$ & \scalebox{0.925}{FPR95}$\downarrow$ & \scalebox{0.925}{AUROC}$\uparrow$   \\
\midrule

\multicolumn{9}{c}{1-shot} \vspace{0.1em}\\
CoOp$_{\scalebox{0.7}{MCM}}$ & 
43.67$^{\scalebox{0.6}{$\pm$} 5.22}$ & 90.91$^{\scalebox{0.6}{$\pm$} 1.01}$ & 
80.32$^{\scalebox{0.6}{$\pm$} 1.69}$ & 71.28$^{\scalebox{0.6}{$\pm$} 1.46}$ & 
73.08$^{\scalebox{0.6}{$\pm$} 2.22}$ & 79.13$^{\scalebox{0.6}{$\pm$} 0.81}$ & 
65.69$^{\scalebox{0.6}{$\pm$} 2.74}$ & 80.44$^{\scalebox{0.6}{$\pm$} 0.94}$   \\

CoOp$_{\scalebox{0.7}{GL}}$ & 
37.36$^{\scalebox{0.6}{$\pm$} 3.79}$ & 91.03$^{\scalebox{0.6}{$\pm$} 0.88}$ & 
74.38$^{\scalebox{0.6}{$\pm$} 1.13}$ & 73.55$^{\scalebox{0.6}{$\pm$} 1.79}$ & 
71.27$^{\scalebox{0.6}{$\pm$} 1.66}$ & 76.16$^{\scalebox{0.6}{$\pm$} 1.41}$ & 
61.01$^{\scalebox{0.6}{$\pm$} 2.03}$ & 80.25$^{\scalebox{0.6}{$\pm$} 1.23}$  \\

LoCoOp$_{\scalebox{0.7}{MCM}}$ & 
44.57$^{\scalebox{0.6}{$\pm$} 1.58}$ & 90.52$^{\scalebox{0.6}{$\pm$} 0.44}$ & 
81.19$^{\scalebox{0.6}{$\pm$} 1.64}$ & 70.18$^{\scalebox{0.6}{$\pm$} 1.96}$ & 
75.60$^{\scalebox{0.6}{$\pm$} 1.96}$ & 78.87$^{\scalebox{0.6}{$\pm$} 0.92}$ & 
67.12$^{\scalebox{0.6}{$\pm$} 1.65}$ & 79.86$^{\scalebox{0.6}{$\pm$} 0.99}$  \\

LoCoOp$_{\scalebox{0.7}{GL}}$ & 
38.46$^{\scalebox{0.6}{$\pm$} 1.54}$ & 91.15$^{\scalebox{0.6}{$\pm$} 0.25}$ & 
76.95$^{\scalebox{0.6}{$\pm$} 0.25}$ & 70.62$^{\scalebox{0.6}{$\pm$} 1.63}$ & 
72.26$^{\scalebox{0.6}{$\pm$} 1.23}$ & 78.80$^{\scalebox{0.6}{$\pm$} 0.91}$ & 
62.56$^{\scalebox{0.6}{$\pm$} 1.63}$ & 80.19$^{\scalebox{0.6}{$\pm$} 0.82}$ \\

IDLike & 
47.48$^{\scalebox{0.6}{$\pm$} 7.36}$ & 89.54$^{\scalebox{0.6}{$\pm$} 2.00}$ & 
71.61$^{\scalebox{0.6}{$\pm$} 5.34}$ & 76.02$^{\scalebox{0.6}{$\pm$} 2.66}$ & 
79.45$^{\scalebox{0.6}{$\pm$} 4.03}$ & 71.59$^{\scalebox{0.6}{$\pm$} 0.86}$ & 
66.18$^{\scalebox{0.6}{$\pm$} 5.58}$ & 79.05$^{\scalebox{0.6}{$\pm$} 1.84}$  \\

SCT$_{\scalebox{0.7}{MCM}}$ & 
43.32$^{\scalebox{0.6}{$\pm$} 7.06}$ & 91.04$^{\scalebox{0.6}{$\pm$} 1.47}$ & 
81.23$^{\scalebox{0.6}{$\pm$} 1.78}$ & 67.81$^{\scalebox{0.6}{$\pm$} 1.78}$ & 
75.01$^{\scalebox{0.6}{$\pm$} 2.24}$ & \underline{80.57}$^{\scalebox{0.6}{$\pm$} 1.03}$ & 
66.52$^{\scalebox{0.6}{$\pm$} 3.53}$ & 79.81$^{\scalebox{0.6}{$\pm$} 1.05}$  \\

SCT$_{\scalebox{0.7}{GL}}$ & 
\underline{35.21}$^{\scalebox{0.6}{$\pm$} 5.99}$ & \underline{91.93}$^{\scalebox{0.6}{$\pm$} 1.43}$ & 
75.37$^{\scalebox{0.6}{$\pm$} 2.52}$ & 70.33$^{\scalebox{0.6}{$\pm$} 1.19}$ & 
70.57$^{\scalebox{0.6}{$\pm$} 3.00}$ & 79.62$^{\scalebox{0.6}{$\pm$} 1.10}$ & 
60.38$^{\scalebox{0.6}{$\pm$} 3.66}$ & 80.63$^{\scalebox{0.6}{$\pm$} 0.88}$  \\

\rowcolor{gray!20}FA$_{\scalebox{0.7}{MCM}}$(Ours) & 
40.21$^{\scalebox{0.6}{$\pm$} 2.76}$ & 91.47$^{\scalebox{0.6}{$\pm$} 0.27}$ & 
\underline{71.30}$^{\scalebox{0.6}{$\pm$} 0.45}$ & \underline{77.08}$^{\scalebox{0.6}{$\pm$}0.50}$ & 
\underline{66.33}$^{\scalebox{0.6}{$\pm$} 0.28}$ & \textbf{81.27}$^{\scalebox{0.6}{$\pm$} 0.53}$ & 
\underline{59.28}$^{\scalebox{0.6}{$\pm$} 0.78}$ & \textbf{83.27}$^{\scalebox{0.6}{$\pm$} 0.29}$  \\

\rowcolor{gray!20}FA$_{\scalebox{0.7}{GL}}$(Ours) & 
\textbf{34.04}$^{\scalebox{0.6}{$\pm$} 1.45}$ & \textbf{92.17}$^{\scalebox{0.6}{$\pm$} 0.08}$ &
\textbf{66.69}$^{\scalebox{0.6}{$\pm$} 0.29}$ & \textbf{78.24}$^{\scalebox{0.6}{$\pm$} 0.52}$ &
\textbf{65.87}$^{\scalebox{0.6}{$\pm$} 0.38}$ & 79.20$^{\scalebox{0.6}{$\pm$} 0.66}$ &
\textbf{55.53}$^{\scalebox{0.6}{$\pm$} 0.47}$ & \underline{83.20}$^{\scalebox{0.6}{$\pm$} 0.37}$  \\

\midrule

\multicolumn{9}{c}{4-shot} \vspace{0.1em}\\
CoOp$_{\scalebox{0.7}{MCM}}$ & 
40.38$^{\scalebox{0.6}{$\pm$} 3.63}$ & 91.40$^{\scalebox{0.6}{$\pm$} 0.79}$ & 
79.46$^{\scalebox{0.6}{$\pm$} 1.49}$ & 72.18$^{\scalebox{0.6}{$\pm$} 1.13}$ & 
71.85$^{\scalebox{0.6}{$\pm$} 1.93}$ & 80.01$^{\scalebox{0.6}{$\pm$} 0.85}$ & 
63.90$^{\scalebox{0.6}{$\pm$} 2.22}$ & 81.19$^{\scalebox{0.6}{$\pm$} 0.57}$   \\
CoOp$_{\scalebox{0.7}{GL}}$ & 
35.24$^{\scalebox{0.6}{$\pm$} 2.38}$ & 91.27$^{\scalebox{0.6}{$\pm$} 0.90}$ & 
74.09$^{\scalebox{0.6}{$\pm$} 1.31}$ & 74.56$^{\scalebox{0.6}{$\pm$} 1.48}$ & 
70.45$^{\scalebox{0.6}{$\pm$} 1.15}$ & 76.23$^{\scalebox{0.6}{$\pm$} 1.70}$ & 
59.92$^{\scalebox{0.6}{$\pm$} 1.47}$ & 80.69$^{\scalebox{0.6}{$\pm$} 1.08}$  \\

LoCoOp$_{\scalebox{0.7}{MCM}}$ & 
40.10$^{\scalebox{0.6}{$\pm$} 2.06}$ & 91.42$^{\scalebox{0.6}{$\pm$} 0.69}$ & 
79.06$^{\scalebox{0.6}{$\pm$} 0.96}$ & 71.68$^{\scalebox{0.6}{$\pm$} 0.68}$ & 
72.46$^{\scalebox{0.6}{$\pm$} 0.41}$ & \underline{80.95}$^{\scalebox{0.6}{$\pm$} 0.41}$ & 
63.87$^{\scalebox{0.6}{$\pm$} 1.13}$ & 81.35$^{\scalebox{0.6}{$\pm$} 0.54}$  \\

LoCoOp$_{\scalebox{0.7}{GL}}$ & 
35.70$^{\scalebox{0.6}{$\pm$} 1.60}$ & 91.91$^{\scalebox{0.6}{$\pm$} 0.46}$ & 
75.86$^{\scalebox{0.6}{$\pm$} 0.37}$ & 72.12$^{\scalebox{0.6}{$\pm$} 0.58}$ & 
68.93$^{\scalebox{0.6}{$\pm$} 1.78}$ & 80.85$^{\scalebox{0.6}{$\pm$} 0.39}$ & 
60.16$^{\scalebox{0.6}{$\pm$} 1.19}$ & 81.63$^{\scalebox{0.6}{$\pm$} 0.35}$ \\

IDLike & 
55.76$^{\scalebox{0.6}{$\pm$} 6.50}$ & 87.94$^{\scalebox{0.6}{$\pm$} 0.90}$ & 
78.82$^{\scalebox{0.6}{$\pm$} 0.25}$ & 72.41$^{\scalebox{0.6}{$\pm$} 1.74}$ & 
83.46$^{\scalebox{0.6}{$\pm$} 0.64}$ & 68.94$^{\scalebox{0.6}{$\pm$} 1.64}$ & 
72.68$^{\scalebox{0.6}{$\pm$} 2.43}$ & 76.43$^{\scalebox{0.6}{$\pm$} 0.79}$  \\

SCT$_{\scalebox{0.7}{MCM}}$ & 
41.15$^{\scalebox{0.6}{$\pm$} 1.31}$ & 91.20$^{\scalebox{0.6}{$\pm$} 0.17}$ & 
80.75$^{\scalebox{0.6}{$\pm$} 1.36}$ & 69.02$^{\scalebox{0.6}{$\pm$} 1.26}$ & 
73.37$^{\scalebox{0.6}{$\pm$} 1.69}$ & 80.63$^{\scalebox{0.6}{$\pm$} 0.84}$ & 
65.09$^{\scalebox{0.6}{$\pm$} 0.88}$ & 80.28$^{\scalebox{0.6}{$\pm$} 0.47}$  \\

SCT$_{\scalebox{0.7}{GL}}$ & 
\underline{35.09}$^{\scalebox{0.6}{$\pm$} 1.20}$ & \underline{91.98}$^{\scalebox{0.6}{$\pm$} 0.19}$ & 
75.56$^{\scalebox{0.6}{$\pm$} 0.92}$ & 71.44$^{\scalebox{0.6}{$\pm$} 1.16}$ & 
69.33$^{\scalebox{0.6}{$\pm$} 1.39}$ & 80.05$^{\scalebox{0.6}{$\pm$} 0.65}$ & 
59.99$^{\scalebox{0.6}{$\pm$} 0.92}$ & 81.17$^{\scalebox{0.6}{$\pm$} 0.23}$  \\

\rowcolor{gray!20}FA$_{\scalebox{0.7}{MCM}}$(Ours) & 
41.23$^{\scalebox{0.6}{$\pm$} 1.49}$ & 91.18$^{\scalebox{0.6}{$\pm$} 0.31}$ & 
\underline{70.43}$^{\scalebox{0.6}{$\pm$} 1.96}$ & \underline{77.40}$^{\scalebox{0.6}{$\pm$} 1.45}$ & 
\underline{64.93}$^{\scalebox{0.6}{$\pm$} 0.65}$ & \textbf{82.18}$^{\scalebox{0.6}{$\pm$} 0.25}$ & 
\underline{58.86}$^{\scalebox{0.6}{$\pm$} 0.98}$ & \underline{83.59}$^{\scalebox{0.6}{$\pm$} 0.49}$  \\

\rowcolor{gray!20}FA$_{\scalebox{0.7}{GL}}$(Ours) & 
\textbf{33.92}$^{\scalebox{0.6}{$\pm$} 0.59}$ & \textbf{91.99}$^{\scalebox{0.6}{$\pm$} 0.26}$ &
\textbf{65.80}$^{\scalebox{0.6}{$\pm$} 1.93}$ & \textbf{78.84}$^{\scalebox{0.6}{$\pm$} 1.19}$ &
\textbf{64.10}$^{\scalebox{0.6}{$\pm$} 1.06}$ & 80.09$^{\scalebox{0.6}{$\pm$} 0.16}$ & 
\textbf{54.60}$^{\scalebox{0.6}{$\pm$} 0.77}$ & \textbf{83.64}$^{\scalebox{0.6}{$\pm$} 0.38}$  \\

\bottomrule
\end{tabularx}

\vspace{-0.5em}
\caption{Challenging ImageNet-1k OOD benchmark under the 1-shot and 4-shot settings. We use the same notation as \cref{tab1}. }
\label{appendix-tab-challenging}

\end{table*}

\section{More experimental details of OOD detection with other ID datasets}
\label{appendix-otherID}

Referring to Section \ref{Comparison},  we also use Food101~\cite{Food101}, StandfordCars~\cite{StanfordCars}, Caltech101~\cite{Caltech101}, FGVCAircraft~\cite{FGVCAircraft}, Flowers102~\cite{Flowers102}, and OxfordPets~\cite{OxfordPets} as ID datasets under the 16-shot setting. As shown in \cref{tab-appendix-otherID}, using these ID datasets for OOD detection presents relatively lower difficulty, and our FA achieves superior OOD detection performance.

\begin{table*}[ht]
\centering

\footnotesize

\begin{tabularx}{\textwidth}{>{\centering\arraybackslash}p{1.15cm} >{\raggedright\arraybackslash}p{1.15cm} *{10}{>{\centering\arraybackslash}X}}
% \begin{tabularx}{\textwidth}{>{\raggedright\arraybackslash}m{1.30cm}*{11}{>{\centering\arraybackslash}m{1.1cm}}}
\toprule

\multirow{2}{*}{\makecell[c]{ID Dataset} } & \multirow{2}{*}{\makecell[c]{Method} } &
\multicolumn{2}{c}{\scalebox{0.95}{iNaturalist}} & \multicolumn{2}{c}{\scalebox{0.95}{SUN}} & \multicolumn{2}{c}{\scalebox{0.95}{Places}} & \multicolumn{2}{c}{\scalebox{0.95}{Textures}} \vrule  & \multicolumn{2}{c} {\scalebox{0.95}{Average}}  \\

 & & \scalebox{0.95}{FPR95}$\downarrow$ & \scalebox{0.95}{AUROC}$\uparrow$ & \scalebox{0.95}{FPR95}$\downarrow$ & \scalebox{0.95}{AUROC}$\uparrow$ & \scalebox{0.95}{FPR95}$\downarrow$ & \scalebox{0.95}{AUROC}$\uparrow$ & \scalebox{0.95}{FPR95}$\downarrow$ & \scalebox{0.95}{AUROC}$\uparrow$ & \scalebox{0.95}{FPR95}$\downarrow$ & \scalebox{0.95}{AUROC}$\uparrow$  \\
\midrule

& CoOp$_{\scalebox{0.7}{MCM}}$ & \scalebox{0.95}{2.29}$^{\scalebox{0.6}{$\pm$} \scalebox{0.75}{1.20}}$ & \scalebox{0.95}{99.35}$^{\scalebox{0.6}{$\pm$} \scalebox{0.75}{0.18}}$ & \scalebox{0.95}{2.02}$^{\scalebox{0.6}{$\pm$} \scalebox{0.75}{0.47}}$ & \scalebox{0.95}{99.49}$^{\scalebox{0.6}{$\pm$} \scalebox{0.75}{0.12}}$ & \scalebox{0.95}{2.25}$^{\scalebox{0.6}{$\pm$} \scalebox{0.75}{0.79}}$ & \scalebox{0.95}{99.43}$^{\scalebox{0.6}{$\pm$} \scalebox{0.75}{0.17}}$ & \scalebox{0.95}{4.74}$^{\scalebox{0.6}{$\pm$} \scalebox{0.75}{0.53}}$ & \scalebox{0.95}{98.53}$^{\scalebox{0.6}{$\pm$} \scalebox{0.75}{0.19}}$ & \scalebox{0.95}{2.83}$^{\scalebox{0.6}{$\pm$} \scalebox{0.75}{0.46}}$ & \scalebox{0.95}{99.19}$^{\scalebox{0.6}{$\pm$} \scalebox{0.75}{0.06}}$  \\
Food101& CoOp$_{\scalebox{0.7}{GL}}$ & \scalebox{0.95}{\underline{0.43}}$^{\scalebox{0.6}{$\pm$} \scalebox{0.75}{0.13}}$ & \scalebox{0.95}{\underline{99.79}}$^{\scalebox{0.6}{$\pm$} \scalebox{0.75}{0.02}}$ & \scalebox{0.95}{0.81}$^{\scalebox{0.6}{$\pm$} \scalebox{0.75}{0.29}}$ & \scalebox{0.95}{99.77}$^{\scalebox{0.6}{$\pm$} \scalebox{0.75}{0.05}}$ & \scalebox{0.95}{0.75}$^{\scalebox{0.6}{$\pm$} \scalebox{0.75}{0.33}}$ & \scalebox{0.95}{99.76}$^{\scalebox{0.6}{$\pm$} \scalebox{0.75}{0.07}}$ & \scalebox{0.95}{4.44}$^{\scalebox{0.6}{$\pm$} \scalebox{0.75}{0.90}}$ & \scalebox{0.95}{98.55}$^{\scalebox{0.6}{$\pm$} \scalebox{0.75}{0.32}}$ & \scalebox{0.95}{1.61}$^{\scalebox{0.6}{$\pm$} \scalebox{0.75}{0.11}}$ & \scalebox{0.95}{99.47}$^{\scalebox{0.6}{$\pm$} \scalebox{0.75}{0.04}}$ \\[0.1em]

\rowcolor{gray!20} & \scalebox{0.9}{FA}$_{\scalebox{0.55}{MCM}}$\scalebox{0.88}{(Ours)} & \scalebox{0.95}{0.57}$^{\scalebox{0.6}{$\pm$} \scalebox{0.75}{0.06}}$ & \scalebox{0.95}{99.77}$^{\scalebox{0.6}{$\pm$} \scalebox{0.75}{0.13}}$ & \scalebox{0.95}{\underline{0.12}}$^{\scalebox{0.6}{$\pm$} \scalebox{0.75}{0.07}}$ & \scalebox{0.95}{\underline{99.93}}$^{\scalebox{0.6}{$\pm$} \scalebox{0.75}{0.02}}$ & \scalebox{0.95}{\underline{0.14}}$^{\scalebox{0.6}{$\pm$} \scalebox{0.75}{0.11}}$ & \scalebox{0.95}{\underline{99.92}}$^{\scalebox{0.6}{$\pm$} \scalebox{0.75}{0.02}}$ & \textbf{\scalebox{0.95}{1.63}}$^{\scalebox{0.6}{$\pm$} \scalebox{0.75}{0.15}}$ & \textbf{\scalebox{0.95}{99.64}}$^{\scalebox{0.6}{$\pm$} \scalebox{0.75}{0.06}}$ & \scalebox{0.95}{\underline{0.61}}$^{\scalebox{0.6}{$\pm$} \scalebox{0.75}{0.13}}$ & \scalebox{0.95}{\underline{99.81}}$^{\scalebox{0.6}{$\pm$} \scalebox{0.75}{0.02}}$ \\

\rowcolor{gray!20} & \scalebox{0.9}{FA}$_{\scalebox{0.6}{GL}}$\scalebox{0.88}{(Ours)} & \textbf{\scalebox{0.95}{0.10}}$^{\scalebox{0.6}{$\pm$} \scalebox{0.75}{0.07}}$ & \textbf{\scalebox{0.95}{99.90}}$^{\scalebox{0.6}{$\pm$} \scalebox{0.75}{0.03}}$ & \textbf{\scalebox{0.95}{0.03}}$^{\scalebox{0.6}{$\pm$} \scalebox{0.75}{0.02}}$ & \textbf{\scalebox{0.95}{99.95}}$^{\scalebox{0.6}{$\pm$} \scalebox{0.75}{0.01}}$ & \textbf{\scalebox{0.95}{0.06}}$^{\scalebox{0.6}{$\pm$} \scalebox{0.75}{0.04}}$ & \textbf{\scalebox{0.95}{99.95}}$^{\scalebox{0.6}{$\pm$} \scalebox{0.75}{0.01}}$ & \scalebox{0.95}{\underline{1.87}}$^{\scalebox{0.6}{$\pm$} \scalebox{0.75}{0.22}}$ & \scalebox{0.95}{\underline{99.56}}$^{\scalebox{0.6}{$\pm$} \scalebox{0.75}{0.07}}$ & \textbf{\scalebox{0.95}{0.52}}$^{\scalebox{0.6}{$\pm$} \scalebox{0.75}{0.05}}$ & \textbf{\scalebox{0.95}{99.84}}$^{\scalebox{0.6}{$\pm$} \scalebox{0.75}{0.01}}$ \\

\midrule

Stanford-& CoOp$_{\scalebox{0.7}{MCM}}$ & \scalebox{0.95}{0.12}$^{\scalebox{0.6}{$\pm$} \scalebox{0.75}{0.02}}$ & \scalebox{0.95}{99.83}$^{\scalebox{0.6}{$\pm$} \scalebox{0.75}{0.08}}$ & \scalebox{0.95}{\underline{0.02}}$^{\scalebox{0.6}{$\pm$} \scalebox{0.75}{0.01}}$ & \scalebox{0.95}{99.97}$^{\scalebox{0.6}{$\pm$} \scalebox{0.75}{0.01}}$ & \scalebox{0.95}{\textbf{0.16}}$^{\scalebox{0.6}{$\pm$} \scalebox{0.75}{0.03}}$ & \scalebox{0.95}{\underline{99.94}}$^{\scalebox{0.6}{$\pm$} \scalebox{0.75}{0.01}}$ & \textbf{\scalebox{0.95}{0.00}}$^{\scalebox{0.6}{$\pm$} \scalebox{0.75}{0.00}}$ & \scalebox{0.95}{99.97}$^{\scalebox{0.6}{$\pm$} \scalebox{0.75}{0.01}}$ & \scalebox{0.95}{0.08}$^{\scalebox{0.6}{$\pm$} \scalebox{0.75}{0.06}}$ & \scalebox{0.95}{99.93}$^{\scalebox{0.6}{$\pm$} \scalebox{0.75}{0.02}}$  \\
Cars& CoOp$_{\scalebox{0.7}{GL}}$ & \scalebox{0.95}{0.06}$^{\scalebox{0.6}{$\pm$} \scalebox{0.75}{0.02}}$ & \scalebox{0.95}{\underline{99.93}}$^{\scalebox{0.6}{$\pm$} \scalebox{0.75}{0.03}}$ & \scalebox{0.95}{0.04}$^{\scalebox{0.6}{$\pm$} \scalebox{0.75}{0.02}}$ & \scalebox{0.95}{99.98}$^{\scalebox{0.6}{$\pm$} \scalebox{0.75}{0.01}}$ & \scalebox{0.95}{0.20}$^{\scalebox{0.6}{$\pm$} \scalebox{0.75}{0.05}}$ & \scalebox{0.95}{99.94}$^{\scalebox{0.6}{$\pm$} \scalebox{0.75}{0.01}}$ & \scalebox{0.95}{\underline{0.01}}$^{\scalebox{0.6}{$\pm$} \scalebox{0.75}{0.00}}$ & \scalebox{0.95}{99.98}$^{\scalebox{0.6}{$\pm$} \scalebox{0.75}{0.01}}$ & \scalebox{0.95}{0.08}$^{\scalebox{0.6}{$\pm$} \scalebox{0.75}{0.03}}$ & \textbf{\scalebox{0.95}{99.96}}$^{\scalebox{0.6}{$\pm$} \scalebox{0.75}{0.02}}$ \\[0.1em]

\rowcolor{gray!20} & \scalebox{0.9}{FA}$_{\scalebox{0.55}{MCM}}$\scalebox{0.88}{(Ours)} & \scalebox{0.95}{\textbf{0.01}}$^{\scalebox{0.6}{$\pm$} \scalebox{0.75}{0.00}}$ & \scalebox{0.95}{99.89}$^{\scalebox{0.6}{$\pm$} \scalebox{0.75}{0.02}}$ & \textbf{\scalebox{0.95}{0.00}}$^{\scalebox{0.6}{$\pm$} \scalebox{0.75}{0.00}}$ & \textbf{\scalebox{0.95}{99.99}}$^{\scalebox{0.6}{$\pm$} \scalebox{0.75}{0.01}}$ & \scalebox{0.95}{\underline{0.16}}$^{\scalebox{0.6}{$\pm$} \scalebox{0.75}{0.06}}$ & \textbf{\scalebox{0.95}{99.94}}$^{\scalebox{0.6}{$\pm$} \scalebox{0.75}{0.01}}$ & \scalebox{0.95}{0.01}$^{\scalebox{0.6}{$\pm$} \scalebox{0.75}{0.01}}$ & \scalebox{0.95}{\underline{99.98}}$^{\scalebox{0.6}{$\pm$} \scalebox{0.75}{0.01}}$ & \textbf{\scalebox{0.95}{0.05}}$^{\scalebox{0.6}{$\pm$} \scalebox{0.75}{0.02}}$ & \scalebox{0.95}{99.95}$^{\scalebox{0.6}{$\pm$} \scalebox{0.75}{0.01}}$ \\

\rowcolor{gray!20} & \scalebox{0.9}{FA}$_{\scalebox{0.6}{GL}}$\scalebox{0.88}{(Ours)} & \scalebox{0.95}{\underline{0.01}}$^{\scalebox{0.6}{$\pm$} \scalebox{0.75}{0.01}}$ & \textbf{\scalebox{0.95}{99.94}}$^{\scalebox{0.6}{$\pm$} \scalebox{0.75}{0.01}}$ & \scalebox{0.95}{0.03}$^{\scalebox{0.6}{$\pm$} \scalebox{0.75}{0.02}}$ & \scalebox{0.95}{\underline{99.98}}$^{\scalebox{0.6}{$\pm$} \scalebox{0.75}{0.01}}$ & \scalebox{0.95}{0.23}$^{\scalebox{0.6}{$\pm$} \scalebox{0.75}{0.02}}$ & \scalebox{0.95}{99.92}$^{\scalebox{0.6}{$\pm$} \scalebox{0.75}{0.01}}$ & \scalebox{0.95}{0.01}$^{\scalebox{0.6}{$\pm$} \scalebox{0.75}{0.01}}$ & \textbf{\scalebox{0.95}{\underline{99.98}}}$^{\scalebox{0.6}{$\pm$} \scalebox{0.75}{0.00}}$ & \scalebox{0.95}{\underline{0.07}}$^{\scalebox{0.6}{$\pm$} \scalebox{0.75}{0.01}}$ & \scalebox{0.95}{\underline{99.95}}$^{\scalebox{0.6}{$\pm$} \scalebox{0.75}{0.00}}$ \\

\midrule

& CoOp$_{\scalebox{0.7}{MCM}}$ & \scalebox{0.95}{19.09}$^{\scalebox{0.6}{$\pm$} \scalebox{0.75}{2.85}}$ & \scalebox{0.95}{96.27}$^{\scalebox{0.6}{$\pm$} \scalebox{0.75}{0.55}}$ & \scalebox{0.95}{6.27}$^{\scalebox{0.6}{$\pm$} \scalebox{0.75}{1.23}}$ & \scalebox{0.95}{98.33}$^{\scalebox{0.6}{$\pm$} \scalebox{0.75}{0.25}}$ & \scalebox{0.95}{11.18}$^{\scalebox{0.6}{$\pm$} \scalebox{0.75}{1.47}}$ & \scalebox{0.95}{97.05}$^{\scalebox{0.6}{$\pm$} \scalebox{0.75}{0.41}}$ & \scalebox{0.95}{6.33}$^{\scalebox{0.6}{$\pm$} \scalebox{0.75}{0.71}}$ & \scalebox{0.95}{98.31}$^{\scalebox{0.6}{$\pm$} \scalebox{0.75}{0.12}}$ & \scalebox{0.95}{10.72}$^{\scalebox{0.6}{$\pm$} \scalebox{0.75}{1.33}}$ & \scalebox{0.95}{97.49}$^{\scalebox{0.6}{$\pm$} \scalebox{0.75}{0.31}}$  \\
Caltech101& CoOp$_{\scalebox{0.7}{GL}}$ & \scalebox{0.95}{16.93}$^{\scalebox{0.6}{$\pm$} \scalebox{0.75}{8.92}}$ & \scalebox{0.95}{96.60}$^{\scalebox{0.6}{$\pm$} \scalebox{0.75}{1.42}}$ & \scalebox{0.95}{7.13}$^{\scalebox{0.6}{$\pm$} \scalebox{0.75}{2.93}}$ & \scalebox{0.95}{98.28}$^{\scalebox{0.6}{$\pm$} \scalebox{0.75}{0.52}}$ & \scalebox{0.95}{12.83}$^{\scalebox{0.6}{$\pm$} \scalebox{0.75}{4.12}}$ & \scalebox{0.95}{96.89}$^{\scalebox{0.6}{$\pm$} \scalebox{0.75}{0.84}}$ & \scalebox{0.95}{11.35}$^{\scalebox{0.6}{$\pm$} \scalebox{0.75}{1.75}}$ & \scalebox{0.95}{97.52}$^{\scalebox{0.6}{$\pm$} \scalebox{0.75}{0.34}}$ & \scalebox{0.95}{12.06}$^{\scalebox{0.6}{$\pm$} \scalebox{0.75}{4.35}}$ & \scalebox{0.95}{97.33}$^{\scalebox{0.6}{$\pm$} \scalebox{0.75}{0.73}}$ \\[0.1em]

\rowcolor{gray!20} & \scalebox{0.9}{FA}$_{\scalebox{0.55}{MCM}}$\scalebox{0.88}{(Ours)} & \scalebox{0.95}{\underline{10.82}}$^{\scalebox{0.6}{$\pm$} \scalebox{0.75}{3.21}}$ & \scalebox{0.95}{\underline{97.83}}$^{\scalebox{0.6}{$\pm$} \scalebox{0.75}{0.51}}$ & \textbf{\scalebox{0.95}{4.16}}$^{\scalebox{0.6}{$\pm$} \scalebox{0.75}{0.70}}$ & \textbf{\scalebox{0.95}{98.93}}$^{\scalebox{0.6}{$\pm$} \scalebox{0.75}{0.20}}$ & \textbf{\scalebox{0.95}{6.66}}$^{\scalebox{0.6}{$\pm$} \scalebox{0.75}{0.56}}$ & \textbf{\scalebox{0.95}{98.30}}$^{\scalebox{0.6}{$\pm$} \scalebox{0.75}{0.19}}$ & \textbf{\scalebox{0.95}{3.74}}$^{\scalebox{0.6}{$\pm$} \scalebox{0.75}{0.79}}$ & \textbf{\scalebox{0.95}{99.09}}$^{\scalebox{0.6}{$\pm$} \scalebox{0.75}{0.18}}$ & \scalebox{0.95}{\underline{6.35}}$^{\scalebox{0.6}{$\pm$} \scalebox{0.75}{0.97}}$ & \scalebox{0.95}{\underline{98.54}}$^{\scalebox{0.6}{$\pm$} \scalebox{0.75}{0.19}}$ \\

\rowcolor{gray!20} & \scalebox{0.9}{FA}$_{\scalebox{0.6}{GL}}$\scalebox{0.88}{(Ours)} & \textbf{\scalebox{0.95}{6.54}}$^{\scalebox{0.6}{$\pm$} \scalebox{0.75}{1.96}}$ & \textbf{\scalebox{0.95}{98.56}}$^{\scalebox{0.6}{$\pm$} \scalebox{0.75}{0.29}}$ & \scalebox{0.95}{\underline{4.44}}$^{\scalebox{0.6}{$\pm$} \scalebox{0.75}{0.38}}$ & \scalebox{0.95}{\underline{98.89}}$^{\scalebox{0.6}{$\pm$} \scalebox{0.75}{0.11}}$ & \scalebox{0.95}{\underline{7.02}}$^{\scalebox{0.6}{$\pm$} \scalebox{0.75}{0.21}}$ & \scalebox{0.95}{\underline{98.28}}$^{\scalebox{0.6}{$\pm$} \scalebox{0.75}{0.05}}$ & \scalebox{0.95}{\underline{4.49}}$^{\scalebox{0.6}{$\pm$} \scalebox{0.75}{1.00}}$ & \scalebox{0.95}{\underline{98.92}}$^{\scalebox{0.6}{$\pm$} \scalebox{0.75}{0.17}}$ & \textbf{\scalebox{0.95}{5.62}}$^{\scalebox{0.6}{$\pm$} \scalebox{0.75}{0.77}}$ & \textbf{\scalebox{0.95}{98.66}}$^{\scalebox{0.6}{$\pm$} \scalebox{0.75}{0.11}}$ \\

\midrule

FGVC-& CoOp$_{\scalebox{0.7}{MCM}}$ & \scalebox{0.95}{43.96}$^{\scalebox{0.5}{$\pm$} \scalebox{0.75}{6.28}}$ & \scalebox{0.95}{89.57}$^{\scalebox{0.5}{$\pm$} \scalebox{0.75}{1.36}}$ & \scalebox{0.95}{18.31}$^{\scalebox{0.5}{$\pm$} \scalebox{0.75}{5.64}}$ & \scalebox{0.95}{96.40}$^{\scalebox{0.5}{$\pm$} \scalebox{0.75}{0.95}}$ & \scalebox{0.95}{21.97}$^{\scalebox{0.5}{$\pm$} \scalebox{0.75}{7.33}}$ & \scalebox{0.95}{94.99}$^{\scalebox{0.5}{$\pm$} \scalebox{0.75}{1.36}}$ & \scalebox{0.95}{17.41}$^{\scalebox{0.5}{$\pm$} \scalebox{0.75}{4.81}}$ & \scalebox{0.95}{96.34}$^{\scalebox{0.5}{$\pm$} \scalebox{0.75}{0.89}}$ & \scalebox{0.95}{25.41}$^{\scalebox{0.5}{$\pm$} \scalebox{0.75}{4.92}}$ & \scalebox{0.95}{94.33}$^{\scalebox{0.5}{$\pm$} \scalebox{0.75}{0.81}}$  \\

Aircraft& CoOp$_{\scalebox{0.7}{GL}}$ & \scalebox{0.95}{64.31}$^{\scalebox{0.5}{$\pm$} \scalebox{0.75}{8.29}}$ & \scalebox{0.95}{81.60}$^{\scalebox{0.5}{$\pm$} \scalebox{0.75}{2.68}}$ & \scalebox{0.95}{41.94}$^{\scalebox{0.5}{$\pm$} \scalebox{0.75}{8.24}}$ & \scalebox{0.95}{90.56}$^{\scalebox{0.5}{$\pm$} \scalebox{0.75}{2.34}}$ & \scalebox{0.95}{45.08}$^{\scalebox{0.5}{$\pm$} \scalebox{0.75}{8.21}}$ & \scalebox{0.95}{88.59}$^{\scalebox{0.5}{$\pm$} \scalebox{0.75}{2.29}}$ & \scalebox{0.95}{54.41}$^{\scalebox{0.5}{$\pm$} \scalebox{0.75}{9.95}}$ & \scalebox{0.95}{84.72}$^{\scalebox{0.5}{$\pm$} \scalebox{0.75}{4.19}}$ & \scalebox{0.95}{51.44}$^{\scalebox{0.5}{$\pm$} \scalebox{0.75}{7.16}}$ & \scalebox{0.95}{86.37}$^{\scalebox{0.5}{$\pm$} \scalebox{0.75}{2.19}}$ \\

\rowcolor{gray!20} & \scalebox{0.9}{FA}$_{\scalebox{0.55}{MCM}}$\scalebox{0.88}{(Ours)} & \scalebox{0.95}{\textbf{7.07}}$^{\scalebox{0.5}{$\pm$} \scalebox{0.75}{3.78}}$ & \scalebox{0.95}{\textbf{98.54}}$^{\scalebox{0.5}{$\pm$} \scalebox{0.75}{0.65}}$ & \textbf{\scalebox{0.95}{0.49}}$^{\scalebox{0.5}{$\pm$} \scalebox{0.75}{0.49}}$ & \textbf{\scalebox{0.95}{99.75}}$^{\scalebox{0.5}{$\pm$} \scalebox{0.75}{0.16}}$ & \textbf{\scalebox{0.95}{1.70}}$^{\scalebox{0.5}{$\pm$} \scalebox{0.75}{0.55}}$ & \textbf{\scalebox{0.95}{99.52}}$^{\scalebox{0.5}{$\pm$} \scalebox{0.75}{0.18}}$ & \textbf{\scalebox{0.95}{0.58}}$^{\scalebox{0.5}{$\pm$} \scalebox{0.75}{0.51}}$ & \textbf{\scalebox{0.95}{99.81}}$^{\scalebox{0.5}{$\pm$} \scalebox{0.75}{0.12}}$ & \textbf{\scalebox{0.95}{2.46}}$^{\scalebox{0.5}{$\pm$} \scalebox{0.75}{1.32}}$ & \textbf{\scalebox{0.95}{99.40}}$^{\scalebox{0.5}{$\pm$} \scalebox{0.75}{0.27}}$ \\

\rowcolor{gray!20} & \scalebox{0.9}{FA}$_{\scalebox{0.6}{GL}}$\scalebox{0.88}{(Ours)} & \scalebox{0.95}{\underline{20.61}}$^{\scalebox{0.5}{$\pm$} \scalebox{0.75}{6.50}}$ & \scalebox{0.95}{\underline{95.76}}$^{\scalebox{0.5}{$\pm$} \scalebox{0.75}{1.37}}$ & \scalebox{0.95}{\underline{2.83}}$^{\scalebox{0.5}{$\pm$} \scalebox{0.75}{1.29}}$ & \scalebox{0.95}{\underline{99.33}}$^{\scalebox{0.5}{$\pm$} \scalebox{0.75}{0.27}}$ & \scalebox{0.95}{\underline{3.96}}$^{\scalebox{0.5}{$\pm$} \scalebox{0.75}{0.89}}$ & \scalebox{0.95}{\underline{98.87}}$^{\scalebox{0.5}{$\pm$} \scalebox{0.75}{0.26}}$ & \scalebox{0.95}{\underline{3.69}}$^{\scalebox{0.5}{$\pm$} \scalebox{0.75}{1.36}}$ & \scalebox{0.95}{\underline{99.12}}$^{\scalebox{0.5}{$\pm$} \scalebox{0.75}{0.34}}$ & \scalebox{0.95}{\underline{7.77}}$^{\scalebox{0.5}{$\pm$} \scalebox{0.75}{2.45}}$ & \scalebox{0.95}{\underline{98.27}}$^{\scalebox{0.5}{$\pm$} \scalebox{0.75}{0.53}}$ \\

\midrule

& CoOp$_{\scalebox{0.7}{MCM}}$ & \scalebox{0.95}{30.90}$^{\scalebox{0.6}{$\pm$} \scalebox{0.75}{1.80}}$ & \scalebox{0.95}{92.38}$^{\scalebox{0.6}{$\pm$} \scalebox{0.75}{0.66}}$ & \scalebox{0.95}{1.84}$^{\scalebox{0.6}{$\pm$} \scalebox{0.75}{1.58}}$ & \scalebox{0.95}{99.57}$^{\scalebox{0.6}{$\pm$} \scalebox{0.75}{0.33}}$ & \scalebox{0.95}{3.18}$^{\scalebox{0.6}{$\pm$} \scalebox{0.75}{1.68}}$ & \scalebox{0.95}{99.21}$^{\scalebox{0.6}{$\pm$} \scalebox{0.75}{0.36}}$ & \underline{\scalebox{0.95}{2.14}}$^{\scalebox{0.6}{$\pm$} \scalebox{0.75}{0.56}}$ & \underline{\scalebox{0.95}{99.53}}$^{\scalebox{0.6}{$\pm$} \scalebox{0.75}{0.09}}$ & \scalebox{0.95}{9.52}$^{\scalebox{0.6}{$\pm$} \scalebox{0.75}{1.19}}$ & \scalebox{0.95}{97.67}$^{\scalebox{0.6}{$\pm$} \scalebox{0.75}{0.33}}$  \\
Flowers102& CoOp$_{\scalebox{0.7}{GL}}$ & \scalebox{0.95}{33.32}$^{\scalebox{0.6}{$\pm$} \scalebox{0.75}{3.10}}$ & \scalebox{0.95}{92.25}$^{\scalebox{0.6}{$\pm$} \scalebox{0.75}{0.79}}$ & \scalebox{0.95}{2.56}$^{\scalebox{0.6}{$\pm$} \scalebox{0.75}{2.16}}$ & \scalebox{0.95}{99.49}$^{\scalebox{0.6}{$\pm$} \scalebox{0.75}{0.39}}$ & \scalebox{0.95}{3.96}$^{\scalebox{0.6}{$\pm$} \scalebox{0.75}{1.91}}$ & \scalebox{0.95}{99.08}$^{\scalebox{0.6}{$\pm$} \scalebox{0.75}{0.38}}$ & \scalebox{0.95}{4.29}$^{\scalebox{0.6}{$\pm$} \scalebox{0.75}{0.64}}$ & \scalebox{0.95}{99.07}$^{\scalebox{0.6}{$\pm$} \scalebox{0.75}{0.16}}$ & \scalebox{0.95}{11.03}$^{\scalebox{0.6}{$\pm$} \scalebox{0.75}{1.77}}$ & \scalebox{0.95}{97.48}$^{\scalebox{0.6}{$\pm$} \scalebox{0.75}{0.39}}$ \\[0.1em]

\rowcolor{gray!20} & \scalebox{0.9}{FA}$_{\scalebox{0.55}{MCM}}$\scalebox{0.88}{(Ours)} & \textbf{\scalebox{0.95}{20.96}}$^{\scalebox{0.6}{$\pm$} \scalebox{0.75}{4.09}}$ & \textbf{\scalebox{0.95}{95.23}}$^{\scalebox{0.6}{$\pm$} \scalebox{0.75}{0.89}}$ & \textbf{\scalebox{0.95}{0.46}}$^{\scalebox{0.6}{$\pm$} \scalebox{0.75}{0.07}}$ & \textbf{\scalebox{0.95}{99.87}}$^{\scalebox{0.6}{$\pm$} \scalebox{0.75}{0.01}}$ & \textbf{\scalebox{0.95}{1.47}}$^{\scalebox{0.6}{$\pm$} \scalebox{0.75}{0.19}}$ & \textbf{\scalebox{0.95}{99.62}}$^{\scalebox{0.6}{$\pm$} \scalebox{0.75}{0.05}}$ & \textbf{\scalebox{0.95}{0.82}}$^{\scalebox{0.6}{$\pm$} \scalebox{0.75}{0.13}}$ & \textbf{\scalebox{0.95}{99.78}}$^{\scalebox{0.6}{$\pm$} \scalebox{0.75}{0.05}}$ & \textbf{\scalebox{0.95}{5.93}}$^{\scalebox{0.6}{$\pm$} \scalebox{0.75}{1.09}}$ & \textbf{\scalebox{0.95}{98.63}}$^{\scalebox{0.6}{$\pm$} \scalebox{0.75}{0.23}}$ \\
\rowcolor{gray!20} & \scalebox{0.9}{FA}$_{\scalebox{0.6}{GL}}$\scalebox{0.88}{(Ours)} & \underline{\scalebox{0.95}{22.63}}$^{\scalebox{0.6}{$\pm$} \scalebox{0.75}{4.45}}$ & \underline{\scalebox{0.95}{94.93}}$^{\scalebox{0.6}{$\pm$} \scalebox{0.75}{0.86}}$ & \underline{\scalebox{0.95}{1.17}}$^{\scalebox{0.6}{$\pm$} \scalebox{0.75}{0.20}}$ & \underline{\scalebox{0.95}{99.71}}$^{\scalebox{0.6}{$\pm$} \scalebox{0.75}{0.02}}$ & \underline{\scalebox{0.95}{2.88}}$^{\scalebox{0.6}{$\pm$} \scalebox{0.75}{0.25}}$ & \underline{\scalebox{0.95}{99.29}}$^{\scalebox{0.6}{$\pm$} \scalebox{0.75}{0.03}}$ & \scalebox{0.95}{2.95}$^{\scalebox{0.6}{$\pm$} \scalebox{0.75}{0.54}}$ & \scalebox{0.95}{99.27}$^{\scalebox{0.6}{$\pm$} \scalebox{0.75}{0.09}}$ & \underline{\scalebox{0.95}{7.41}}$^{\scalebox{0.6}{$\pm$} \scalebox{0.75}{1.28}}$ & \underline{\scalebox{0.95}{98.29}}$^{\scalebox{0.6}{$\pm$} \scalebox{0.75}{0.23}}$ \\

\midrule

Oxford-& CoOp$_{\scalebox{0.7}{MCM}}$ & \scalebox{0.95}{4.64}$^{\scalebox{0.6}{$\pm$} \scalebox{0.75}{1.58}}$ & \scalebox{0.95}{99.02}$^{\scalebox{0.6}{$\pm$} \scalebox{0.75}{0.23}}$ & \scalebox{0.95}{0.33}$^{\scalebox{0.6}{$\pm$} \scalebox{0.75}{0.31}}$ & \scalebox{0.95}{99.90}$^{\scalebox{0.6}{$\pm$} \scalebox{0.75}{0.06}}$ & \scalebox{0.95}{1.02}$^{\scalebox{0.6}{$\pm$} \scalebox{0.75}{0.32}}$ & \scalebox{0.95}{99.77}$^{\scalebox{0.6}{$\pm$} \scalebox{0.75}{0.05}}$ & \scalebox{0.95}{0.79}$^{\scalebox{0.6}{$\pm$} \scalebox{0.75}{0.23}}$ & \scalebox{0.95}{99.82}$^{\scalebox{0.6}{$\pm$} \scalebox{0.75}{0.04}}$ & \scalebox{0.95}{1.69}$^{\scalebox{0.6}{$\pm$} \scalebox{0.75}{0.55}}$ & \scalebox{0.95}{99.63}$^{\scalebox{0.6}{$\pm$} \scalebox{0.75}{0.08}}$  \\
Pets& CoOp$_{\scalebox{0.7}{GL}}$ & \scalebox{0.95}{2.88}$^{\scalebox{0.6}{$\pm$} \scalebox{0.75}{1.13}}$ & \scalebox{0.95}{99.38}$^{\scalebox{0.6}{$\pm$} \scalebox{0.75}{0.18}}$ & \underline{\scalebox{0.95}{0.31}}$^{\scalebox{0.6}{$\pm$} \scalebox{0.75}{0.08}}$ & \underline{\scalebox{0.95}{99.93}}$^{\scalebox{0.6}{$\pm$} \scalebox{0.75}{0.05}}$ & \scalebox{0.95}{0.98}$^{\scalebox{0.6}{$\pm$} \scalebox{0.75}{0.47}}$ & \scalebox{0.95}{99.79}$^{\scalebox{0.6}{$\pm$} \scalebox{0.75}{0.07}}$ & \scalebox{0.95}{1.80}$^{\scalebox{0.6}{$\pm$} \scalebox{0.75}{0.75}}$ & \scalebox{0.95}{99.59}$^{\scalebox{0.6}{$\pm$} \scalebox{0.75}{0.15}}$ & \scalebox{0.95}{1.49}$^{\scalebox{0.6}{$\pm$} \scalebox{0.75}{0.67}}$ & \scalebox{0.95}{99.67}$^{\scalebox{0.6}{$\pm$} \scalebox{0.75}{0.11}}$ \\[0.1em]

\rowcolor{gray!20} & \scalebox{0.9}{FA}$_{\scalebox{0.55}{MCM}}$\scalebox{0.88}{(Ours)} & \underline{\scalebox{0.95}{0.86}}$^{\scalebox{0.6}{$\pm$} \scalebox{0.75}{0.60}}$ & \underline{\scalebox{0.95}{99.73}}$^{\scalebox{0.6}{$\pm$} \scalebox{0.75}{0.15}}$ & \textbf{\scalebox{0.95}{0.09}}$^{\scalebox{0.6}{$\pm$} \scalebox{0.75}{0.06}}$ & \textbf{\scalebox{0.95}{99.94}}$^{\scalebox{0.6}{$\pm$} \scalebox{0.75}{0.02}}$ & \textbf{\scalebox{0.95}{0.37}}$^{\scalebox{0.6}{$\pm$} \scalebox{0.75}{0.12}}$ & \textbf{\scalebox{0.95}{99.90}}$^{\scalebox{0.6}{$\pm$} \scalebox{0.75}{0.02}}$ & \textbf{\scalebox{0.95}{0.22}}$^{\scalebox{0.6}{$\pm$} \scalebox{0.75}{0.14}}$ & \textbf{\scalebox{0.95}{99.94}}$^{\scalebox{0.6}{$\pm$} \scalebox{0.75}{0.03}}$ & \textbf{\scalebox{0.95}{0.39}}$^{\scalebox{0.6}{$\pm$} \scalebox{0.75}{0.22}}$ & \textbf{\scalebox{0.95}{99.88}}$^{\scalebox{0.6}{$\pm$} \scalebox{0.75}{0.05}}$ \\

\rowcolor{gray!20} & \scalebox{0.9}{FA}$_{\scalebox{0.6}{GL}}$\scalebox{0.88}{(Ours)} & 
\textbf{\scalebox{0.95}{0.63}}$^{\scalebox{0.6}{$\pm$} \scalebox{0.75}{0.42}}$ & 
\textbf{\scalebox{0.95}{99.80}}$^{\scalebox{0.6}{$\pm$} \scalebox{0.75}{0.09}}$ & 
\scalebox{0.95}{0.75}$^{\scalebox{0.6}{$\pm$} \scalebox{0.75}{0.08}}$ & 
\scalebox{0.95}{99.81}$^{\scalebox{0.6}{$\pm$} \scalebox{0.75}{0.02}}$ & 
\underline{\scalebox{0.95}{0.64}}$^{\scalebox{0.6}{$\pm$} \scalebox{0.75}{0.10}}$ & 
\underline{\scalebox{0.95}{99.84}}$^{\scalebox{0.6}{$\pm$} \scalebox{0.75}{0.03}}$ & 
\underline{\scalebox{0.95}{0.54}}$^{\scalebox{0.6}{$\pm$} \scalebox{0.75}{0.28}}$ & 
\underline{\scalebox{0.95}{99.85}}$^{\scalebox{0.6}{$\pm$} \scalebox{0.75}{0.05}}$ & 
\underline{\scalebox{0.95}{0.64}}$^{\scalebox{0.6}{$\pm$} \scalebox{0.75}{0.21}}$ & 
\underline{\scalebox{0.95}{99.82}}$^{\scalebox{0.6}{$\pm$} \scalebox{0.75}{0.05}}$ \\

\bottomrule
\end{tabularx}

\vspace{-1em}
\caption{Full numerical results of OOD detection performance with other ID datasets under the 16-shot setting. We use the same notation as \cref{tab1}.}
\label{tab-appendix-otherID}

\end{table*}

% \section{Visualization }
% \label{appendix-Visualization}
% To better defend the motivation, we have included a T-SNE plot on real data.
% We compare with SCT~\cite{sct}, the SOTA few-shot OOD-dependent method. As shown in Fig.~\ref{fig-rebuttal}, unlike SCT's text features, the forced text features yield higher similarity with ID than OOD image features, leading to higher ID data scores than SCT during testing (refer to Section \ref{sec:fpl} for details). 

% \begin{figure*}[ht]
%   \centering
%   % \fbox{\rule{0pt}{1.1in} \rule{0.9\linewidth}{0pt}}
%   \includegraphics[height=3.7cm]{ICCV2025-Author-Kit-Feb/fig_rebuttal.pdf}
  
%    \vspace{-1.2em}
%    \caption{ \scalebox{0.85}{\textbf{T-SNE plot of our FA and SCT.}  }}
%    \label{fig-rebuttal}
%    \vspace{-2em}
% \end{figure*}

\end{document}